\documentclass[twoside,11pt]{article}
\usepackage{jmlr2e}

\usepackage{cite, graphicx, amsmath, amssymb}
\usepackage{algorithmic}
\usepackage{algorithm}
\usepackage{subcaption}

\firstpageno{1}

\newcommand{\R}{{\mathbb R}}
\newcommand{\x}{{\mathbf x}}

\DeclareMathOperator{\var}{var}
\DeclareMathOperator{\Span}{span}

\begin{document}

\title{Manifold unwrapping using density ridges}

\author{\name Jonas N.\ Myhre \email jonas.n.myhre@uit.no \\
       \addr Machine Learning @ UiT Lab \\ Department of Physics and Technology\\
       University of Troms\o \ (UiT) - The Arctic University of Norway\\
       9037 Troms\o, NORWAY
       \AND
       \name Matineh Shaker \email shaker@ece.neu.edu\\
       \addr Department of Electrical and Computer Engineering\\
       409 Dana Research Center, 360 Huntington Avenue\\
       Northeastern University\\
       Boston, MA 02115, USA
       \AND
       \name M. Devrim Kaba \email defrim@gmail.com \\
       \addr General Electric Global Research Center\\
       1 Research Circle,\\
       Niskayuna, NY 12309, USA
       \AND
       \name Robert Jenssen \email robert.jenssen@uit.no \\
       \addr Machine Learning @ UiT Lab \\ Department of Physics and Technology\\
       University of Troms\o \ (UiT) - The Arctic University of Norway\\
       9037 Troms\o, NORWAY
       \AND
       \name Deniz Erdogmus \email erdogmus@ece.neu.edu \\
       \addr Department of Electrical and Computer Engineering\\
       409 Dana Research Center, 360 Huntington Avenue\\
       Northeastern University\\
       Boston, MA 02115, USA}

\editor{Unknown}

\maketitle

\begin{abstract}
  Research on manifold learning within a density ridge estimation framework has shown great potential in recent work for both estimation and denoising of manifolds, building on the intuitive and well-defined notion of principal curves and surfaces. However, the problem of unwrapping or unfolding manifolds has received relatively little attention within the density ridge approach, despite being an integral part of manifold learning in general. 
  
  This paper proposes two novel algorithms for unwrapping manifolds based on estimated principal curves and surfaces for one- and multi-dimensional manifolds respectively. The methods of unwrapping are founded in the realization that both principal curves and principal surfaces will have inherent local maxima of the probability density function. Following this observation, coordinate systems that follow the shape of the manifold can be computed by following the integral curves of the gradient flow of a kernel density estimate on the manifold.
  
  Furthermore, since integral curves of the gradient flow of a kernel density estimate is inherently local, we propose to stitch together local coordinate systems using parallel transport along the manifold.
  
  We provide numerical experiments on both real and synthetic data that illustrates clear and intuitive unwrapping results comparable to state-of-the-art manifold learning algorithms.





\end{abstract}

\section{Introduction}
\label{sec:Introduction}
Manifold learning is one of the fundamental fields in Machine
Learning~\citep{van2009dimensionality,
tenenbaum2000global,zhang2004principal, sun2012stochastic, weinberger2006introduction, burges2009dimension, sorzano2014survey}. It is motivated by the notion that high
dimensional data sets often exhibit intrinsic structure that is concentrated on
or near (sub)manifolds of lower local dimensionality. In addition to this, intrinsic structure that is nonlinear prevents the use of fast linear machine
learning algorithms. 
Furthermore, recent developments have shown that the ridges\footnote{The terms \emph{density ridge}, \emph{principal curve} and \emph{principal surface} will be used interchangeably depending on context.} of the probability
density function can be used for both estimation and denoising of
manifolds~\citep{ozertem2011locally, Genovese2014,
gerber2013regularization, pulkkinen2015ridge, pulkkinen2014generative,hein2006manifold}.

In practice manifold learning is often posed as either learning coordinates that describe the
intrinsic manifold or simply as unwrapping, stretching or unfolding the manifold
such that it can, to some extent, be treated as a linear subspace of $\R^D$.

Most algorithms in manifold learning rely on the
so-called manifold assumption~\citep{lin2008riemannian}. 
The assumption is that the data cloud in
a vector space lies on or close to a hypersurface of lower intrinsic
dimension. 
Doing inference on or along this hypersurface would enable
the use of simple linear methods which are fast and theoretically
well-defined. 
A much used example to illustrate these concepts is the so-called swiss roll~\citep{tenenbaum2000global}. It consists of a 2-D plane embedded in 3-D and folded into a swiss
roll shape. 
Calculating distances along this shape using standard Euclidean geometry
would fail, and so will simple methods like for example linear regression if we wanted to estimate the data relationships. See e.g. Tenenbaum et al.,~\citep{tenenbaum2000global}, for illustrations and details.

There exists many algorithms and methods that globally or locally
tries to unwrap or unfold nonlinear manifolds, but none of which
use the density ridges for this purpose. 
Principal curves and density ridge algorithms, \citep{hastie1989,ozertem2011locally,Gerber2013}, find curves or surfaces that are smooth estimates of the underlying manifold, but the possible intrinsic nonlinear structure will still be present such that linear operations in the ambient space will fail to represent the manifold.

Expanding on these works, our main contribution in this paper is \emph{the use of density ridges to unwrap manifolds}.
In this paper we use the ridges of a probability density function as
defined by Ozertem and
Erdogmus,~\citep{ozertem2011locally}, based on the gradient and Hessian
of a kernel density estimate. This definition, though computationally
expensive, gives an explicit non-parametric and non-ambiguous description of the
density ridge. Furthermore, by tracing curve lengths along the local
one-dimensional ridges of a mode attraction basin\footnote{The basin of attraction is the the set of points in a probability density where the gradient flow curves converges to the same point -- the local mode.} we can estimate distances
and a local coordinate system \emph{along} the manifold. By doing this,
a local linear representation along the manifold structure can be constructed.
In addition, we propose an algorithm inspired by parallel transport
as defined in differential geometry to combine several local coordinate
representations to form a global representation along a single connected
manifold.
Finally we suggest an extension from using one-dimensional ridges to
unfold the manifold to using $d$-dimensional representations.
This is done by creating locally flat charts of the manifold combined with approximated geodesic
distances to relate the charts together to form a global representation of the manifold.

To summarize, our key contributions are: (1) Introduce a new algorithm for manifold
learning that use the ridges of the probability density function. This
enables a completely geometrically interpretable scheme for learning the
structure of a hidden manifold. Note that this work is an extension of the
scheme presented in,~\citep{shaker2014invertible}, which is only local and limited to manifolds
with intrinsic one-dimensional structure. (2) We present a new algorithm
for combining local coordinate charts along a one-dimensional density
ridge into a global atlas over a single connected manifold. (3) We
present a local chart approximation for $d$-dimensional manifolds and an
algorithm for calculating geodesics on $d$-dimensional manifolds
represented by density ridges. Combining these gives a global unwrapping of a single connected manifold containing multiple charts.




\subsection{Related research}
\label{sub:Related_research}
In this work the goal is to unwrap manifolds estimated by density ridges. This yields two fields of literature to reference, the topics related to principal curves/density ridges and topics related to manifold learning. We start with principal curves and surfaces.

Research related to the manifold-estimation part of our work is usually described by the interrelated
terms \emph{principal curves}, \emph{principal surfaces} and
\emph{principal manifolds}.
Principal curves and related subjects have been studied in a wide array
of settings, often under different names and configurations. The most
common case is \emph{principal curves}, which are smooth one-dimensional
curves embedded in $\mathbb R^D$. They were originally posed as smooth
self-consistent curves passing through the `middle' of the data by Hastie and
Stuetzle~\citep{hastie1989}. Kegl et al.\ proposed to constrain the length of the
principal curves, enabling a more thorough theoretical analysis and simpler
implementation~\citep{kegl2000learning}. Einbeck et al.\ suggested an
algorithm for finding local principal curves based on local linear
projections~\citep{Einbeck2005}. Our work is motivated by the work of
Ozertem and Erdogmus, where \emph{local} principal curves are defined as
one-dimensional \emph{ridges} of the data probability
density~\citep{ozertem2011locally}. This framework also extends naturally to cover higher dimensional principal surfaces by using higher dimensional density ridges. See also the work of
Bas, ~\citep{bas2011extracting}, for further details and applications.

In the last few years several very interesting works related to the
density ridge interpretation have been introduced. Genovese et al.\ have
provided a theoretical analysis of non-parametric density ridge
estimation \citep{Genovese2014} and Chen provides asymptotic theory and
formulates density ridges as so-called solution
manifolds~\citep{Chen2014, chen2014asymptotic}.
Another
closely related setting to the principal curves and derivatives of the
probability density is \emph{filament estimation}, which is in practice
very similar to principal curves~\citep{genovese2009path,
genovese2012geometry}.  For estimation purposes, Pulkkinen and
Pulkkinen et al.\, have proposed several practical improvements for the
density ridge estimation framework of Ozertem and
Erdogmus~\citep{ozertem2011locally}, both in the generative model and
as a method of optimization~\citep{pulkkinen2015ridge,
pulkkinen2014generative}. Finally we mention the recent work of Gerber
and Whitaker, \citep{gerber2013regularization}, where the original
framework of Hastie and Stuetzle~\citep{hastie1989} is reformulated to
avoid regularization both for principal curves and surfaces.

Beyond metastudies of principal curves and density ridges themselves,
several applications can be found. Examples include template based
classification~\citep{chang1998principal}, ice floe
detection~\citep{banfield1992ice}, galaxy
identification~\citep{chen2014generalized}, character
skeletonization~\citep{Krzy2002} and clustering~\citep{stanford2000finding}
to name a few.

As an intermediate summary we recall that the work in this paper expands the idea of a principal curve or surface to actually unfolding/unwrapping the curves or surfaces found by the algorithms mentioned above.

In the field of manifold learning within the machine learning literature,
several works exists that are closely related to ours in terms of manifold unwrapping. The closest in
principles and ideas are the \emph{local tangent space alignment}
algorithm,~\citep{zhang2004principal}, the
LOGMAP algorithm for calculating normal
coordinates of a manifold,~\citep{brun2005fast}, and the manifold chart stitching of Pitelis et al.\ in
\citep{pitelis2013learning}. The manifold parzen algorithm and
contractive autoencoders should also be
mentioned,~\citep{vincent2002manifold, rifai2011learning}, as they are, similar to our work, algorithms that tries to learn representations along non-linear
manifolds via estimates of the probability density. 
\subsection{Motivation: Using density ridges to unwrap manifolds}
\label{sub:Motivation_p1}
A kernel density ridge can be used to completely describe a biased version, called a surrogate, of a manifold embedded in $\mathbb R^D$ given sufficient samples and bounded noise~\citep{Genovese2014}. 
%
%
The main motivation for using density ridges to \emph{locally} unwrap nonlinear manifolds comes from the following.
\begin{itemize}
    \item As the ridge is estimated via a probability density estimate, all ridges of dimensions higher than zero will by construction induce a gradient flow contained along the ridge/manifold surrogate. This is a trivial consequence of the definition of critical set by Ozertem and Erdogmus~\citep{ozertem2011locally}.
    \item In the one-dimensional case the gradient flow completely describes local sections (attraction basins) of the manifold estimate, and can thus be unwrapped directly by representing the manifold in approximate arc-length coordinates. This is analoguous to pulling a curly string taut.
    \item In the $d$-dimensional case the gradient flows that converge to the same point can be approximated and unwrapped by a local linear projection. This is analoguous to flattening a curled up sheet.
\end{itemize}
To further emphasize the first point: Unless the underlying manifold is a straight line or flat in all directions, the kernel
    size of the density estimators used -- presented in equation~\eqref{eq:kde} on page~\pageref{eq:kde} -- has to be bounded. 
    This will thus inherently lead to local maxima along the density ridges, which further leads to a gradient flow contained along the ridge. 

For clarity, we split the next part of our motivation into the one-dimensional case and the general $d$-dimensional case, starting with the one-dimensional case.

\begin{figure}[htbp]
\centering
\begin{subfigure}{.45\linewidth}
    \includegraphics[width=\linewidth]{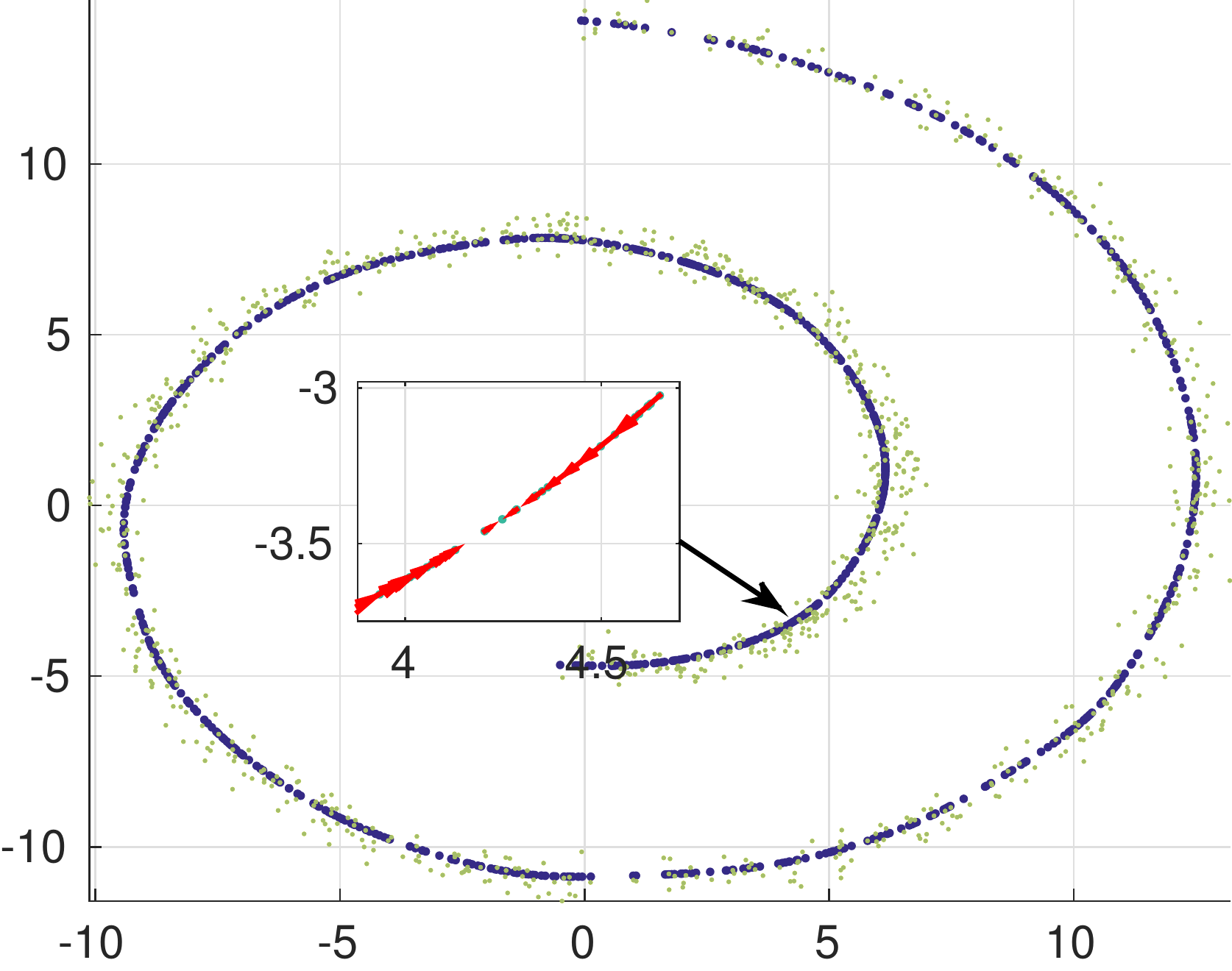}
    \caption{Noisy one-dimensional manifold(light green) embedded in $\mathbb R^2$ and the density ridge estimate(dark blue). The small window shows a selected portion of the density ridge and the corresponding gradient flow. We can see the corresponding local maxima of the gradient flow in the figure on the left.}
    \label{fig:spiral_data_and_curves}
  \end{subfigure} 
  \qquad
\begin{subfigure}{.45\linewidth}
    \includegraphics[width=\linewidth]{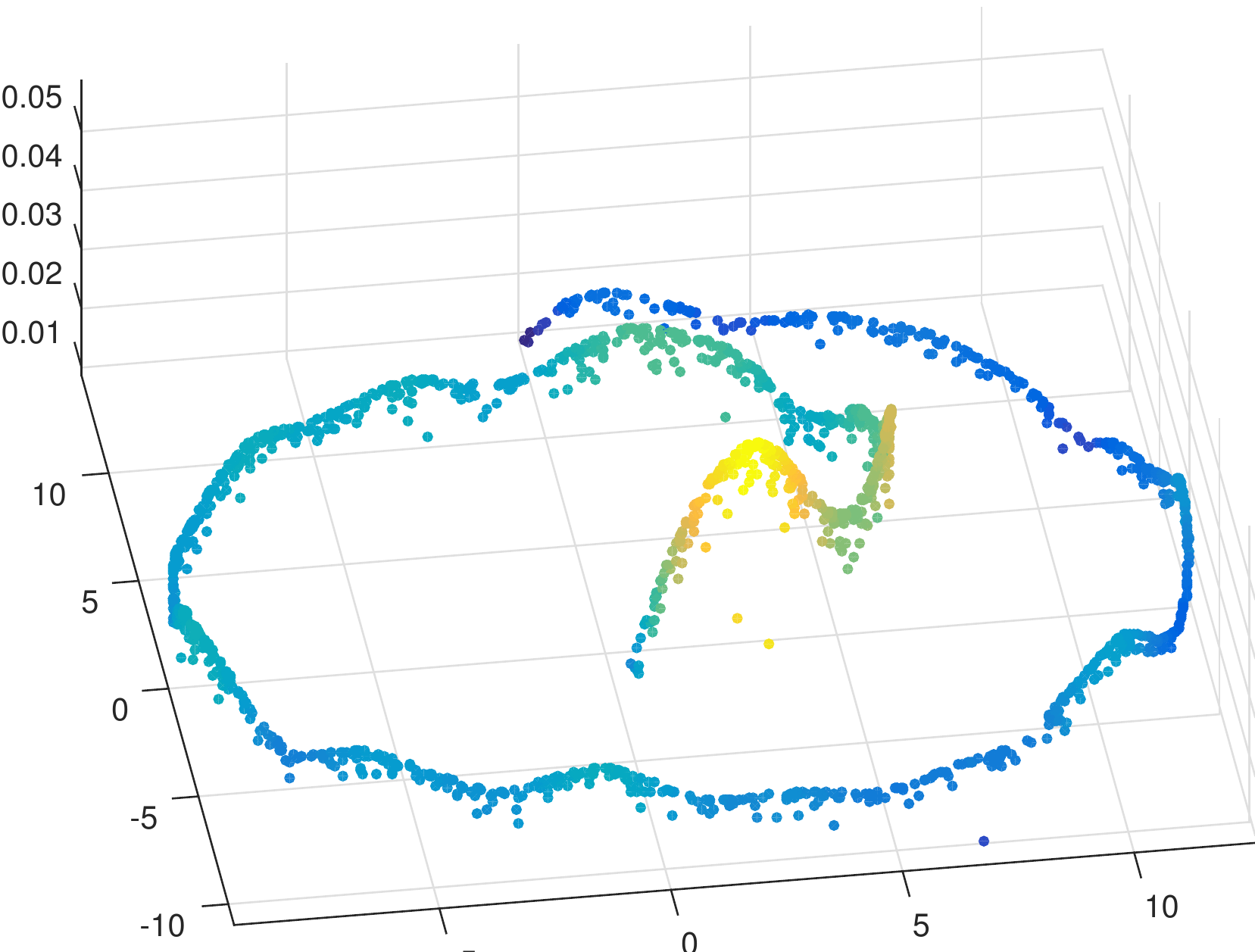}
    \caption{Probability density along the intrinsic one-dimensional ridge from Figure~\ref{fig:local_maxima_ex} shown on vertical axis. Note that the density along the ridge has several local maxima. The points are additionaly marked with color proportional to the density value -- from dark blue (low) to yellow (high) -- to ease the interpretation.}
    \label{fig:local_maxima_ex}
  \end{subfigure} 
    \caption{Gradient flow and probability density along a noisy one-dimensional manifold.}
    \label{fig:one_dim_grad_flow}
\end{figure}

In the one dimensional case all points along the ridge can be described by the piece-wise arc-length of the gradient flow integral curve from the point to the local mode where the gradient vanishes.
This will create a complete description of the coordinate distances from all points on the ridge to their corresponding local maxima collected in local sections of the curve. In
Figure~\ref{fig:spiral_data_and_curves}, we see a 2-D version of the
synthetic `swiss roll'
dataset with added Gaussian noise, \citep{roweis2000nonlinear}. It is clearly a one-dimensional manifold (curve) embedded in $\mathbb R^2$. The density ridge is
shown in dark blue dots,
while the noisy data is shown in light green dots. The ridge was found using
the framework of Ozertem and Erdogmus,~\citep{ozertem2011locally}, and a Runge-Kutta-Fehlberg ODE solver implemented
in MATLAB\footnote{We will go into further details in
section~\ref{sec:principal_curve_unwrapping}}. 
We see that the ridge is a good approximate for the
manifold. To further illustrate the properties of the ridges
we show a 3-D plot with the ridge of the noisy swiss roll on the
horizontal
axes, and the density values along the ridge shown on the vertical axis in
Figure~\ref{fig:local_maxima_ex}.
It is evident that there are local maxima along the single connected ridge. Again, this
observation is key to our proposition; by following the gradient flow of the
density along the ridge we have a way of tracing the underlying manifold
and thus calculating distances \emph{along} the manifold. 

\begin{figure}[htbp]
    \centering
    \begin{subfigure}{.4\linewidth}
        \centering
        \includegraphics[width=\linewidth]{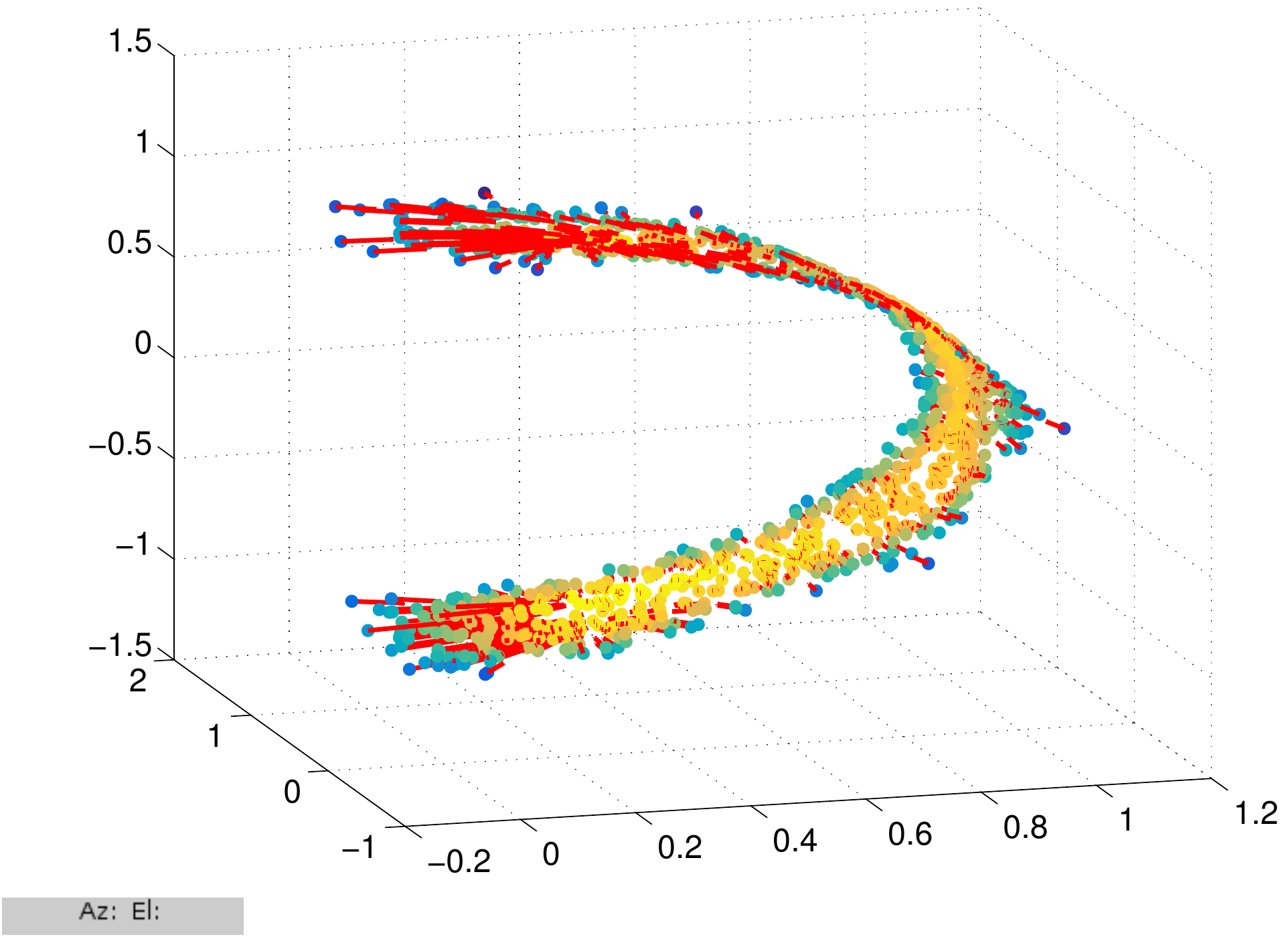}
        \caption{Side view of manifold with gradient flow.}
        \label{fig:d_dim_grad_flow_side}
    \end{subfigure}
    \begin{subfigure}{.4\linewidth}
        \centering
        \includegraphics[width=\linewidth]{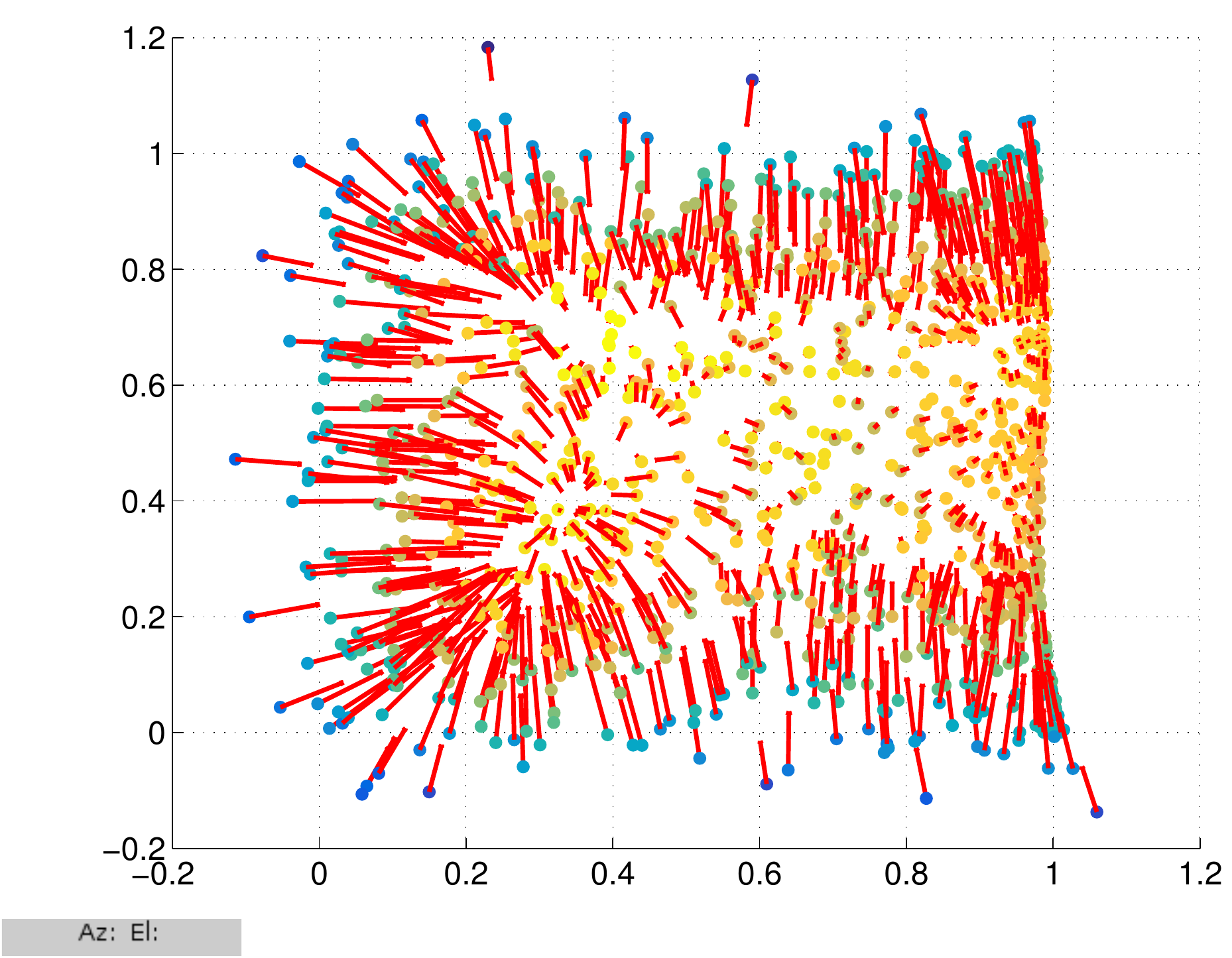}
        \caption{Top view of manifold with gradient flow.}
        \label{fig:d_dim_grad_flow_top}
    \end{subfigure}
    \begin{subfigure}{.4\linewidth}
        \centering
        \includegraphics[width=\linewidth]{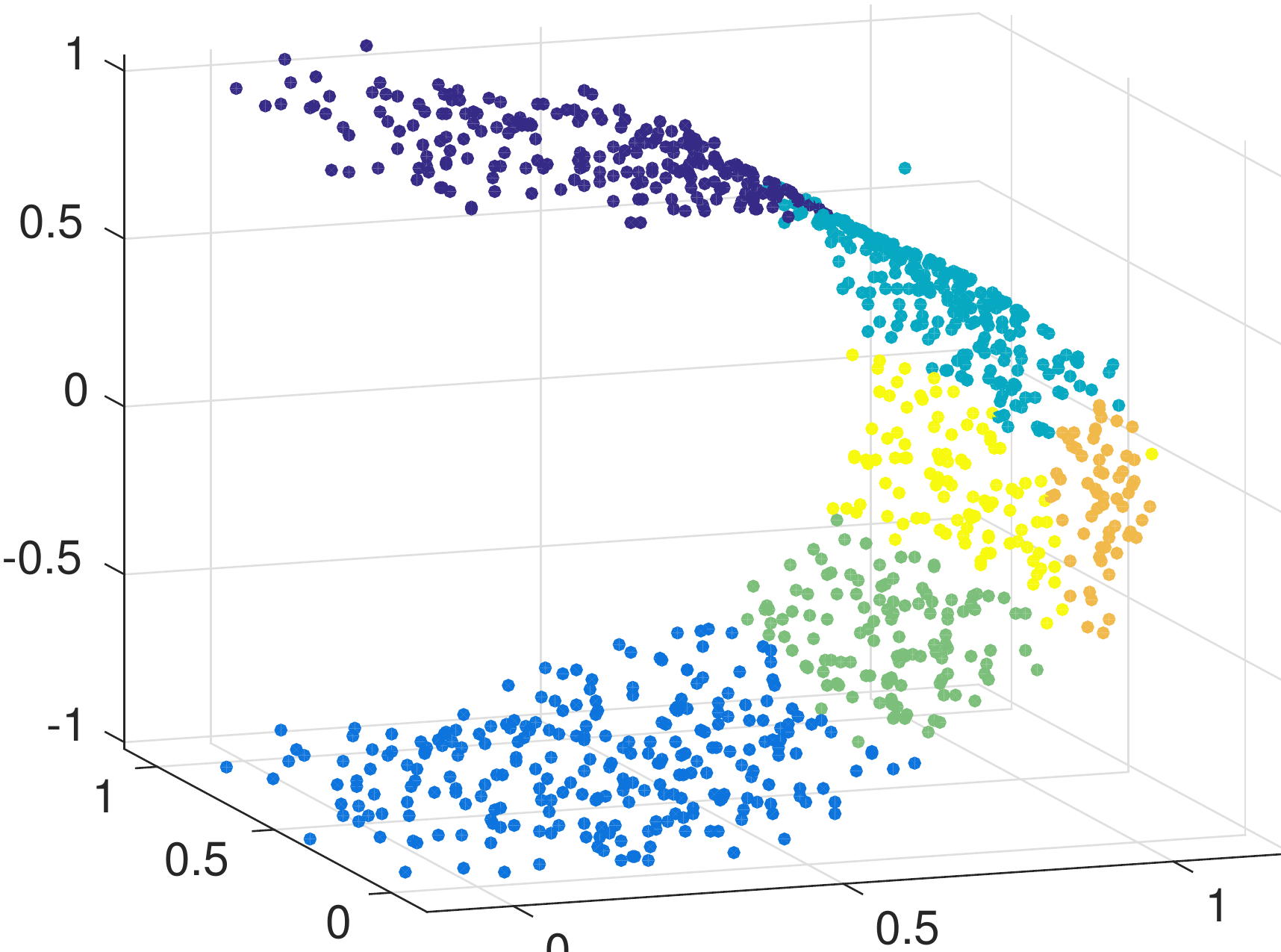}
        \caption{Resulting attraction bases of the gradient flow.}
        \label{fig:mot_d_dim_charts}
    \end{subfigure}
       \begin{subfigure}{.4\linewidth}
        \centering
        \includegraphics[width=\linewidth]{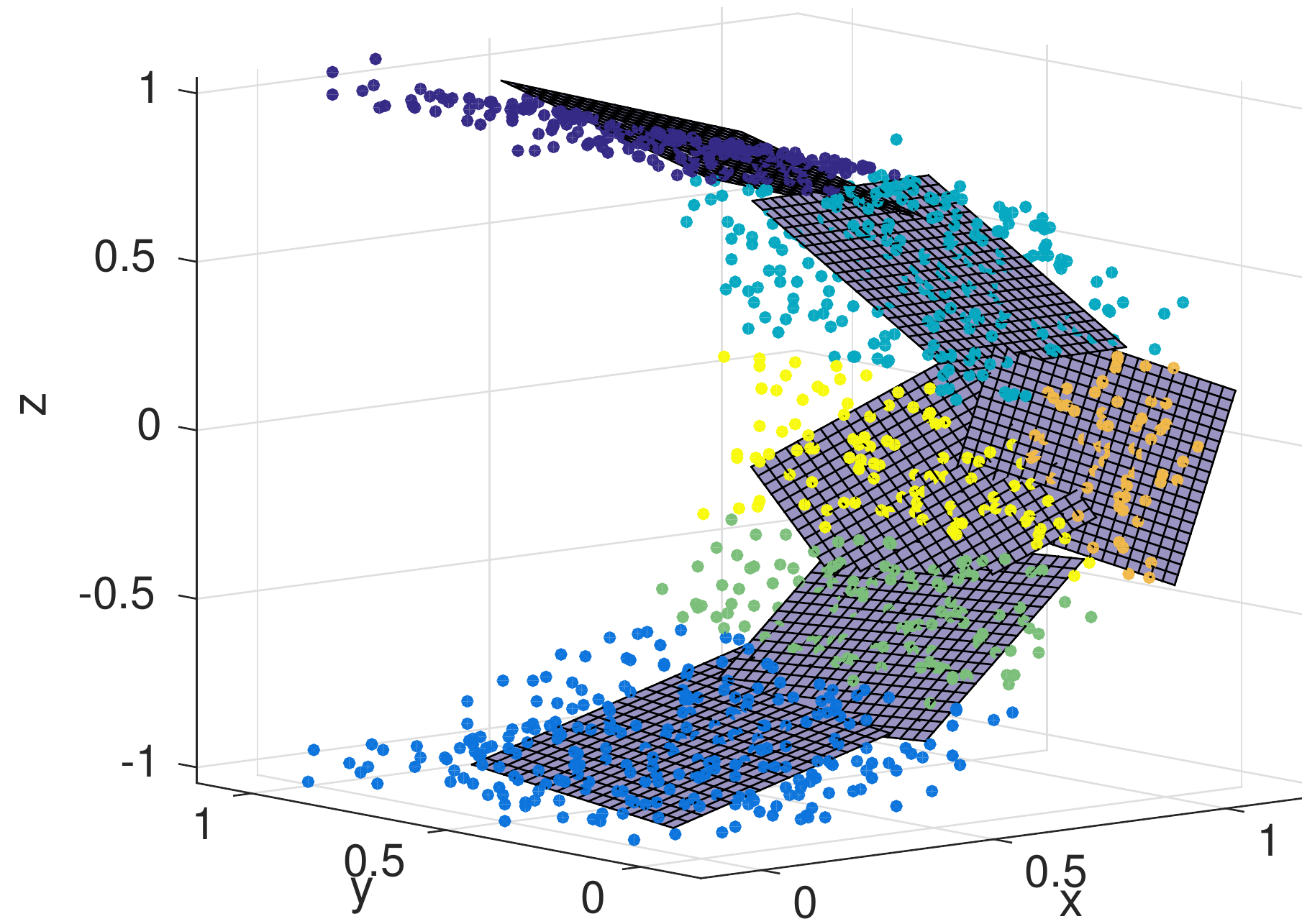}
        \caption{Local linear flattening of attraction basins.}
        \label{fig:d_dim_flat}
    \end{subfigure}
    \caption{Gradient flow on a two-dimensional manifold.}
    \label{fig:d_dim_grad_flow}
\end{figure}

Continuing to the $d$-dimensional case, the gradient flow will generate unique curves from each point on the density towards the local modes. The manifold can thus be inuitively be separated into parts based where the gradient flow converges to the same critical points. Each part can then be mapped by a linear projection approximating the logarithmic map (see for example Figure~\ref{fig:tangentspace} on page~\pageref{fig:tangentspace} or Brun et al.\ \citep{brun2005fast}).
In Figure~\ref{fig:d_dim_grad_flow} we see a two-dimensional manifold embedded in $\mathbb R^3$ estimated by a density ridge and the corresponding gradient flows from different views in Figure~\ref{fig:d_dim_grad_flow_top} and \ref{fig:d_dim_grad_flow_side}. In Figure~\ref{fig:mot_d_dim_charts} we see the sets of points, attraction basins, that have gradient flows that converge to the same point marked with similar colors. 
We see in Figure~\ref{fig:d_dim_flat} that restricting the local flattening to cover only the different attraction basins separately will result in meaningful compact local representations. Flattening the entire manifold would, depending on perspective, most certainly introduce some kind of unwanted folding or collapsing.

%
Continuing from local to global unwrapping, we note that as we are using a kernel density estimate the gradient flow of the probability density is inherently local. Thus piecewise distances -- or any other further operations -- will also
only be local, and to obtain a global unwrapping we need some
strategy to combine local unwrappings. This is also clearly seen in Figure~\ref{fig:d_dim_flat}; the local attraction basins are flattened intuitively, but globally the representation is not meaningful.

To obtain a global representation we take inspiration from the concept of parallel transport from differential geometry~\citep{lee1997riemannian} and introduce a translation along the tangent vectors of
the one-dimensional ridges (translation along geodesics in the $d$-dimensional case) to obtain a complete unwrapped, or simply flat, representation of the ridge.
This is analoguous to parallel transporting vectors along a
geodesic, see for example Freifeld et al.,~\citep{freifeld2014model}, except that the piece-wise local distance along
the geodesic is also added to the transported vector. 

\begin{figure}[htbp]
\centering
\begin{subfigure}{.45\linewidth}
    \includegraphics[width=\linewidth]{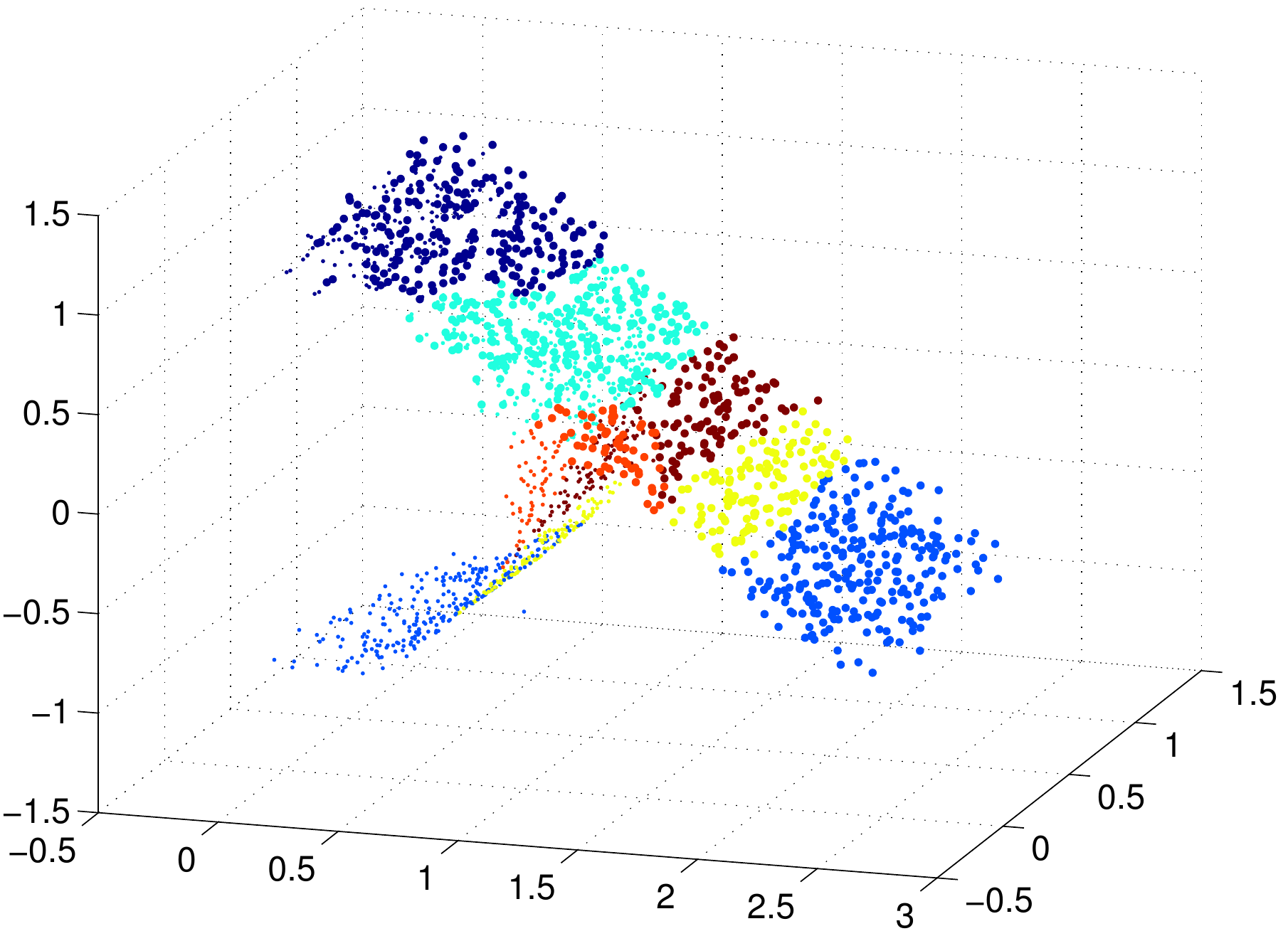}
    \caption{Density ridge and unwrapped version. The color coding corresponds to attraction basins.}
    \label{fig:d_dim_motivation_unwrapped}
  \end{subfigure} 
  \qquad
\begin{subfigure}{.45\linewidth}
    \includegraphics[width=\linewidth]{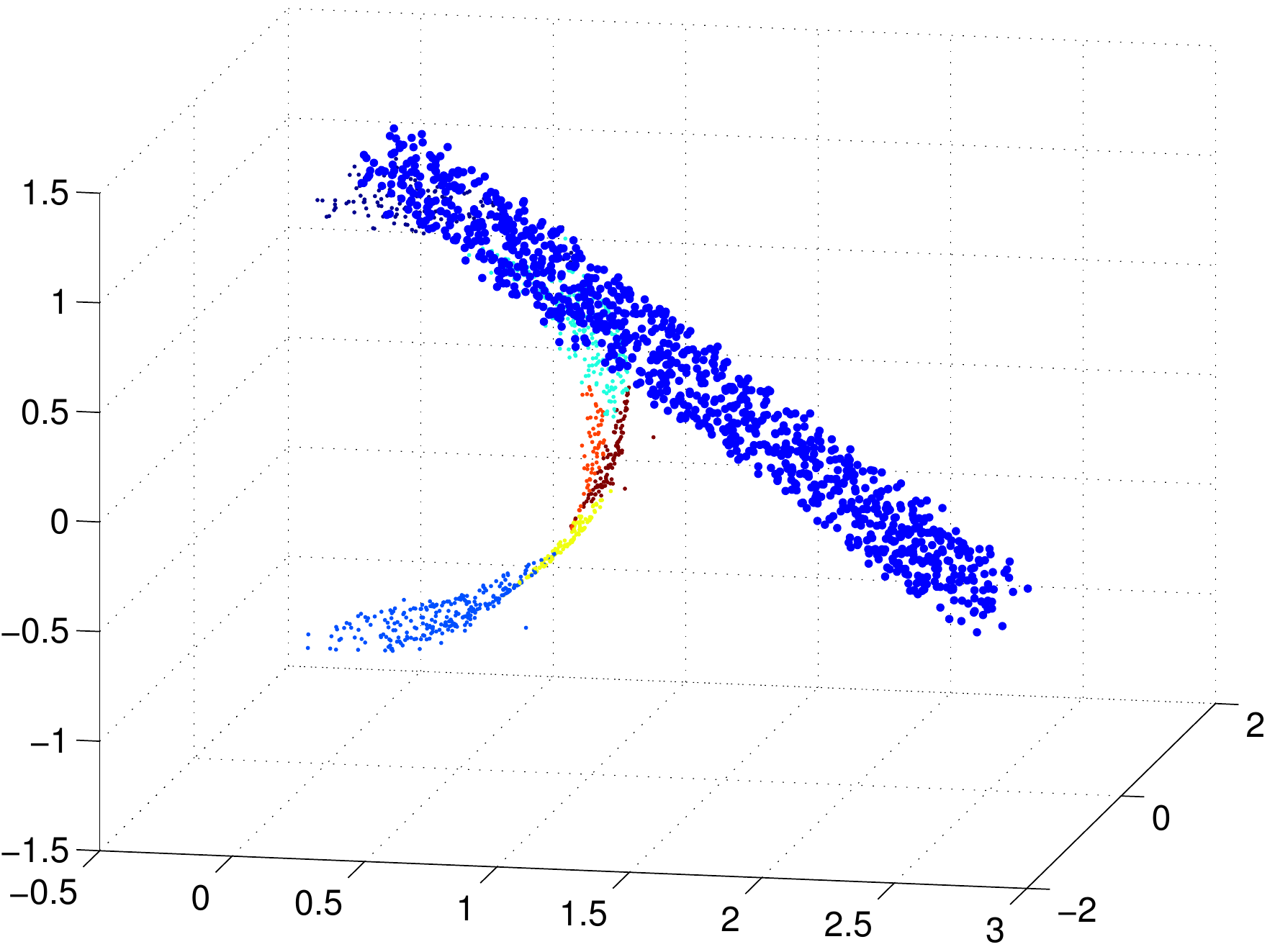}
    \caption{Density ridge and unwrapped version. The color coding of the unwrapped manifold is kept a single color (blue) for easier visualization.}
    \label{fig:d_dim_motivation_unwrapped2}
  \end{subfigure} 
    \caption{Unwrapped version of the two dimensional manifold in Figure~\ref{fig:d_dim_grad_flow}.}
    \label{fig:d_dim_unwrapped}
\end{figure}

The result of global unwrapping\footnote{The result was obtained by using Algorithm~\ref{alg:isometricUnfolding} on page~\pageref{alg:isometricUnfolding} } of the two-dimensional manifold in Figure~\ref{fig:d_dim_grad_flow} is shown in Figure~\ref{fig:d_dim_unwrapped}. 

To summarize, the gradient flow of a kernel density allows unwrapping manifolds using the following two steps:
\begin{enumerate}
    \item Estimating local coordinate patches in the basins of attractions of the pdf.
    \item Stitching together local coordinate charts by approximate parallel tranport/translation along geodesics.
\end{enumerate}
To conclude this section we add two further comments that should be noted:
(1) If the intrinsic manifold turns out to be an embedded \emph{surface} ($S \subset \mathbb R^2$) or
\emph{hypersurface} ($S\subset \mathbb R^d,\;d\geq 3$), we have to consider the case of non-zero curvature,
where an isometric (distance-preserving) mapping or unfolding cannot
be guaranteed. For simplicity, in the rest of this paper we assume that the manifolds we are working with are isometric to $\mathbb R^d$. 
(2) In some cases, the definition of the density ridges allows for
     multiple  orthogonal one-dimensional
     ridges, up to
     $d$ if $f(\x):\mathbb R^d\to \mathbb R$. In such cases local orthogonal coordinate systems can be estimated
     by first estimating the orthogonal ridges and then follow the
     gradient flow along each separate curve. This is useful in
     situations where the underlying one-dimensional manifold is
     corrupted by noise that changes along the manifold.
\subsection{Structure of paper}
The rest of the paper is organized as follows:
We start by formally introducing principal curves and density ridges as
in the framework of Ozertem and Erdogmus~\citep{ozertem2011locally}. We then include some relevant
topics from differential geometry, and show
connections to density ridges as found by kernel density estimation in Section~\ref{sec:diff_geo}.
  In Section~\ref{sec:principal_curve_unwrapping} we present the
principal curve/one-dimensional density ridge unwrapping algorithm
followed by the ridge translation algorithm in
Section~\ref{sec:parallel_transport}. Section~\ref{sec:n_dim_ridges}
proposes extensions to density ridges of higher dimension and finally in
Section~\ref{sec:numerical_experiments} we show numerical examples and
illustrations of the presented methods on both synthetic and real datasets.

\section{Principal curves and density ridges}
\label{sub:Principal_curves}

Principal curves were originally introduced by Hastie and
Stuetzle~\citep{hastie1989}. Several
extensions were made, \citep{kegl2000learning, Einbeck2005,
tibshirani1992principal}, until Ozertem and Erdogmus, \citep{ozertem2011locally}, redefined principal
curves and surfaces as being the \emph{ridges} of a probability density estimate,~\citep{ozertem2011locally}.
  Given a probability density $f(\x)$, its gradient
  $g(\x)=\nabla^Tf(\x)$ and
  Hessian matrix $H(\x)=\nabla \nabla^T f(\x)$, the ridge can be defined in terms of the
  eigendecomposition of the Hessian matrix.
\begin{definition}[Ozertem 2011]
  \label{def:ozertem}
  A point $\x$ is on the $d$-dimensional ridge, $R$ of its probability
  density function, when the gradient $g(\x)$ is orthogonal to at least
  $D-d$ eigenvectors of $H(\x)$ and the corresponding $D-d$ eigenvalues are all negative.
\end{definition}
We express the spectral
decomposition of $H$ as $H(\x) = Q(\x)\Lambda(\x)Q(\x)^T$, where $Q(\x)$ is
the matrix of eigenvectors sorted according to the size of the eigenvalue and
$\Lambda_{ii}(\x)=\lambda_i$, $\lambda_1(\x) > \lambda_2(\x) > \dots$,
is a diagonal matrix of sorted eigenvalues. Furthermore $Q(\x)$ can be
decomposed into $\left[Q_\perp(\x) \; Q_\parallel(\x)  \right]$, where
$Q_\perp$ is the $d$ first eigenvectors of $Q(\x)$, and $Q_\parallel$
are the $D-d$ smallest. The latter is referred to as the
\emph{orthogonal subspace} due to the fact that when at a ridge point,
all eigenvectors in $Q_\perp$ will be orthogonal to $g(\x)$. 

This motivates the following initial value problem for
projecting points onto a density ridge:

\begin{equation}
  \frac{\text{d}\mathbf{y}_t}{\text{d}t} =
  V_tV_t^Tg(\mathbf{y}_t),
  \label{eq:ode_pc}
\end{equation}
where $V_t = Q_\perp(\x(t))$ at $\mathbf y_t = \mathbf{y}(t)$, and
$\mathbf{y}(0)=\mathbf{x}$. We denote the set of $\mathbf{y}$'s that
satisfy equation~\eqref{eq:ode_pc}, calculated via the kernel density estimator
$\hat f(\x)$, as the $d$-dimensional \emph{ridge estimator} $\hat R$.

Of great value is the recent paper by Genovese et al.,
\citep{Genovese2014}, which showed that the Hausdorff distance between a $d$-dimensional manifold
embedded in $D$ dimensions and the $d$-dimensional ridge of the density is bounded under certain restrictions wrt.\ noise
and the closeness of the density estimate to the true density. They also show that the kernel density ridges are consistent estimators of the true underlying ridges and refer to the ridge as a \emph{surrogate} of the underlying
manifold\footnote{See Theorem 7 in \citep{Genovese2014}.}. Thus, $\hat R$ is a point set representing the underlying
manifold with theoretically established bounds under Hausdorff loss.

\subsection{Orthogonal local principal curves}
\label{sub:olpc}
In the special case of a projection onto a principal curve, $d=1$, we note that the
construction of the orthogonal subspace allows for choosing $d$
different orthogonal subspaces to project the gradient to. The most
intuitive setting, which is most commonly used~\citep{ozertem2011locally,
Genovese2014, bas2011extracting} is to use the top eigenvector(s)
corresponding to eigenvalue(s) as the orthogonal subspace. We recall
that the eigenvalues and eigenvectors of the Hessian matrix of a
function points to the directions of largest second order change. In the
case of a density ridge, the change in density \emph{along} the ridge is assumed to
be much lower than the change in density when moving \emph{away} from
the ridges, hence the choice of largest eigenvector pointing away from
the ridge.
  This local ordering, though intuitive from the Hessian point of view, has not been well studied, and could in
some cases actually give non-intuitive results\footnote{See the T-shaped
mixture of Gaussians in \citep{ozertem2011locally}.}. We will not go further into this in this paper.

\begin{figure}[htbp]
\begin{center}
  \includegraphics[width=2.5in]{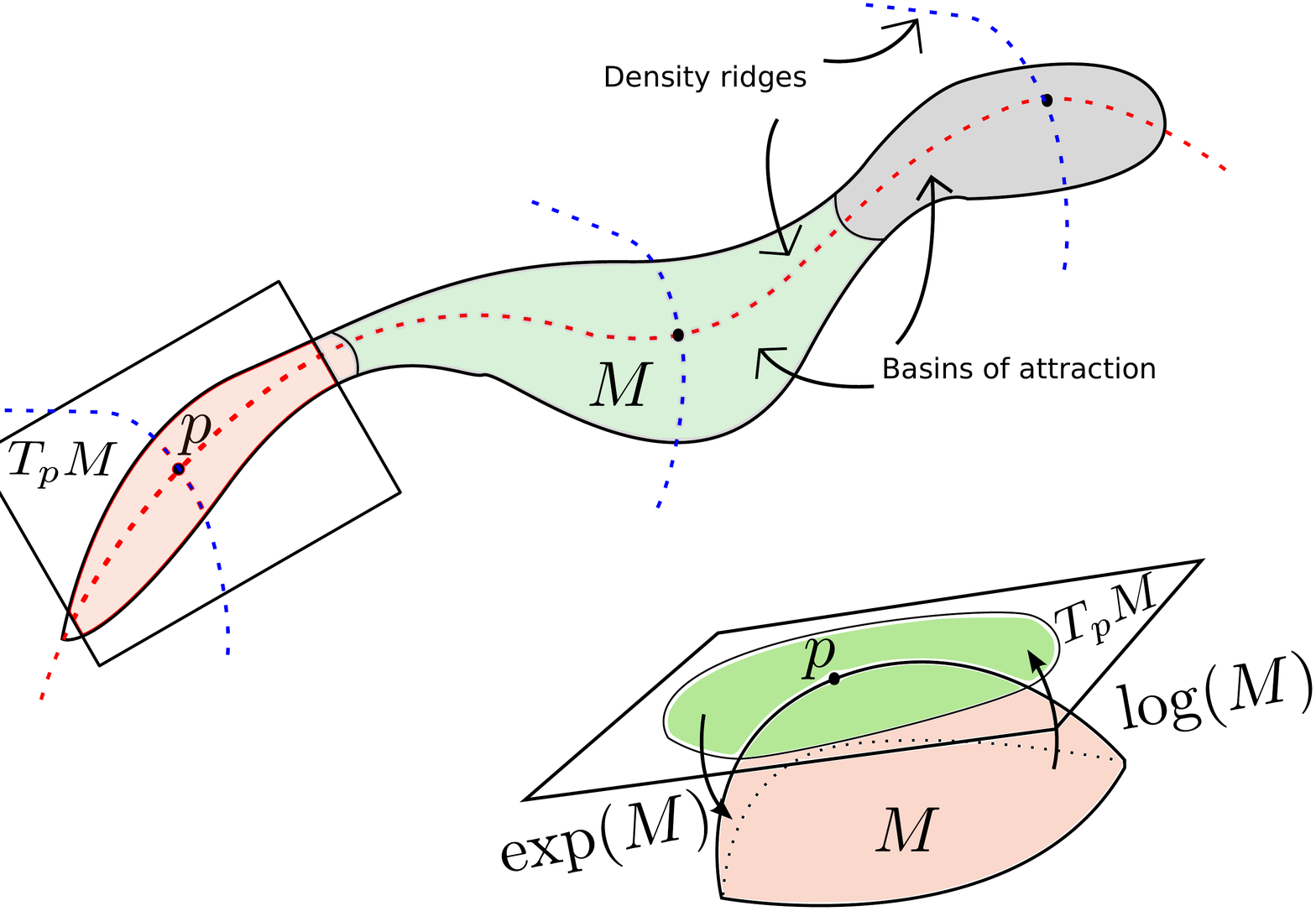}
\end{center}
\caption{A manifold with a one-dimensional ridge/principal curve. The
local basins of attraction can be decomposed using orthogonal principal
curves. $M$ is the intrinsic manifold, $T_pM$ denotes the tangent space
of $M$ at $p$. The green area denoted in the lower right sketch is the
area of $T_pM$ constrained by the \emph{exponential map}.}
\label{fig:tangentspace}
\end{figure}

The importance of this local decomposition is that considering a
$d$-dimensional ridge, the ridge tracing algorithm, presented in
Section~\ref{sec:principal_curve_unwrapping}, can be applied on each of the local
orthogonal subspaces. In the case of e.g.\ different local scales
of noise this could yield a better desciption of the data.
Figure~\ref{fig:tangentspace} shows a conceptual illustration of the
idea, where the overall dominating nonlinearity can be described by a
principal curve/one-dimensional density ridge, but since the orthogonal structure varies along the
ridge local decomposition provides further detail.
\section{Relevant topics from differential geometry}
\label{sec:diff_geo}
\emph{Manifold learning} is a framework
which
is inspired by concepts from differential
geometry~\citep{brun2005fast, tenenbaum2000global, pitelis2013learning,
roweis2000nonlinear, gerber2013regularization,zhang2004principal}. The
main idea is that data sets or data structures seldom fill the vector
space they are represented in. Even in low dimensions, e.g.\
$\mathbb{R}^3$, data often concentrate around clearly bounded subregions
or so-called \emph{manifolds}.

  In some works the problem of learning representations \emph{on} or
  \emph{along} a manifold has been termed manifold unwrapping or
  unfolding~\citep{weinberger2006introduction, sun2012stochastic}. The idea is that learning the structure of a manifold keeps the local structures intact more strongly than global structures, so that the (global) nonlinearities can be unfolded, unrolled or unwrapped. This
  is perhaps the term most related to our work.

Unfolding or unwrapping a manifold can intuitively be performed in two
different ways. Either we use some function that stretches or flattens the
manifold such that we can use linear methods to calculate distances
along the manifold, or we can somehow estimate the structure of the
manifold such that distance measures can be defined directly along the manifold
again allowing linear operations in the resulting coordinate system. 


The ideas and concepts of manifold learning stems from differential geometry, the field of mathematics which studies smooth geometric objects and,  closely related, smooth functions. Its main object of interest is the manifold, which can be roughly regarded as a well behaving smooth (topological) space.

A clear definition of a manifold can be found in either
books of Lee or Tu,
\citep{lee1997riemannian, tu2008introduction}:
\begin{definition} 
A (topological) manifold is a second countable, locally Euclidean, Hausdorff space.
\end{definition}
Local Euclidean structure is analoguous to how humans perceive the
surface of the earth. At smaller scales traversing a path along the surface
will seem like a straight line, but on larger scales
paths along the surface of the earth are clearly curved.
  A Hausdorff space is
a space where two separate points have disjoint
neighboorhoods~\citep{weissteinHausdorff}.  E.g.\ a surface embedded in
$\mathbb R^3$ that intersects with itself will have points that shares
neighborhoods and is thus not Hausdorff.

In simple terms; the intuition behind differential geometry and manifolds is to describe an
object locally at a certain point or a certain homogenuous region using
derivatives. Although often not explicitly stated the kernel density
estimate and its derivatives -- see e.g.~\citep{chacon2013data} -- can be
interpreted in a differential geometry setting. We recall that
Genovese et al.,~\citep{Genovese2014}, has shown that kernel density
derivatives can be used to estimate a smoothed version of an underlying
manifold sampled from data with noise.

Before we go into further discussion of the connection between the
kernel density estimate of data sampled from a manifold and differential
geometry, we introduce a few basic concepts.

  First: throughout this discussion we are talking about \emph{submanifolds}
  of $\mathbb{R}^D$~\citep{lee1997riemannian}. 
  Given a manifold $M$ diffeomorphic to $\mathbb R^d$,
  at each point $p\in M$
  the \emph{tangent space}, $T_pM$, is the Euclidean space of dimension $d$ which is tangent to $M$ at $p$~\citep{lee1997riemannian}.
  See Figure~\ref{fig:tangentspace} for a sketch of the related concepts.
  The term \emph{tangent to}, can intuitively be interpreted as either
  the space of tangent vectors of all possible curves passing through $p$ or the space spanned
by the partial derivatives of the parametrization of $M$ at $p$~\citep{lee1997riemannian}.
  A disjoint union of all tangent spaces
of $M$ is called the \emph{tangent bundle}.
  We denote the coordinate transformation from the tangent
space $T_pM$to the manifold $M$ as the \emph{exponential map}, and the inverse
transformation from the $M$ to the $T_PM$ as the \emph{log
map}~\citep{brun2005fast}.
  Finally we note that vectors in $T_pM$ can be expressed by a local
  basis of differentials $E_i =  \frac{\partial p}{\partial x^i}
$. These are called the \emph{local coordinates} at $p$,
\citep{lee1997riemannian}.
  These local coordinates represents a
Euclidean
subspace of same dimension $d$ as $M$.

As for the statiscal model, we assume the same model as in 
\citep{Genovese2014} where the data points, $X=\left\{ \x_i \right\}_{i=1}^n \in \mathbb{R}^D$, are sampled with noise from a distribution supported on $M$. If we let $P_M$ be the distribution of points on and along $M$ and $\Phi_\sigma$ be a Gaussian distribution with zero mean and $\sigma I$ covariance -- also in $\mathbb R^D$ -- that represents noisy samples that does not lie directly on the manifold, we get the following model\footnote{$*$ denotes convolution.}: $$P = P_M * \Phi_\sigma.$$

\noindent The kernel density estimator is defined as follows:

\begin{equation}
  \hat p(\x) = \frac{1}{n}\sum^{n}_{i=1}
  \frac{1}{\sigma^2}K\left(\frac{||\x - \x_i||}{\sigma^2}\right),
  \label{eq:kde}
\end{equation}
where $K(\cdot)$ is any symmetric and positive semi-definite kernel function. Note that we skip the normalizing constant, as it is simply a matter of scale and we are only interested in the \emph{direction} of the gradient and the Hessian eigenvectors. We also note that in this work we restrict ourselves to use the Gaussian kernel $K(\x_j, \x_i) = \exp(-||x_i-x_j||^2/2\sigma^2)$. 
From the kernel density estimate we can calculate the gradient
as follows(notation adapted from~\citep{Genovese2014}):
\begin{equation}
  g(\x) = - \frac{1}{n} \sum^{n}_{i=1} \frac{\x-\x_i}{\sigma^2}
  K\left(\frac{||\x - \x_i||}{\sigma^2}\right).
  \label{eq:kdde}
\end{equation}
And finally the Hessian matrix:
\begin{equation}
  H(\x) = \frac{1}{n} \sum^{n}_{i=1} \left( \mathbf u_i\mathbf u_i^T
  - \frac{1}{\sigma^2} I
  \right)
  K\left(\frac{||\x - \x_i||}{\sigma^2}\right),
  \label{eq:kde_hess}
\end{equation}
where $\mathbf u_i = (\x-\x_i)$.
From these equations we see that the gradient is simply the average of all
vectors pointing out from $\x$ weighted by the kernel function value of
the norm of the vectors. We also note that depending on the kernel size
$\sigma^2$ the gradient will in practice be the average of the vectors
pointing from $\x$ to all points within the neighborhood covered by the
particular kernel size. If a point and its neighbors within the kernel
size distance lies on or close to the manifold $M$, this clearly
is a good candidate for a local tangent space estimate.

\begin{figure}[htbp]
\begin{center}
  \includegraphics[width=2.5in]{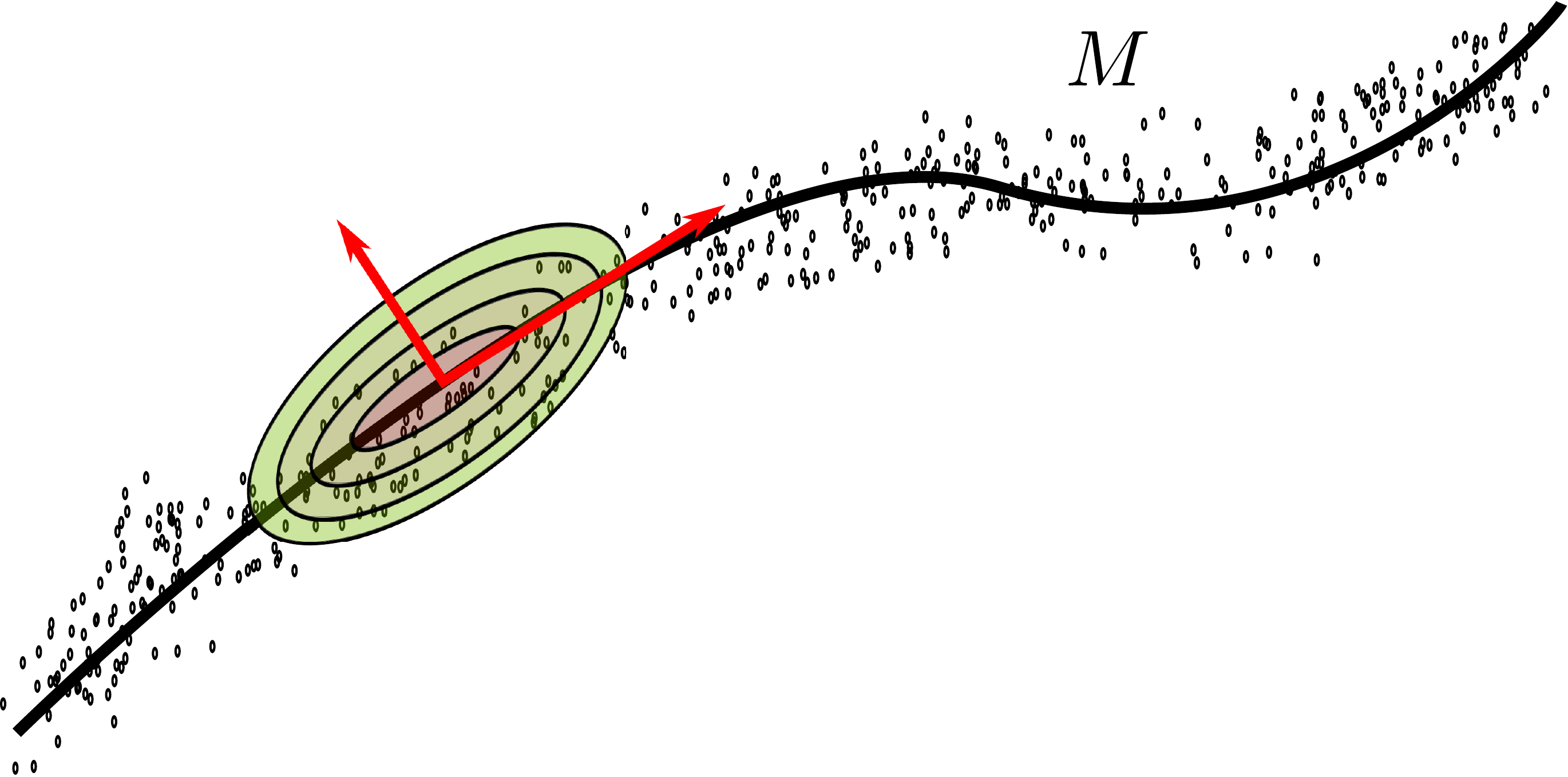}
\end{center}
\caption{Illustration of the eigenvectors (red arrows) of the kernel density Hessian matrix at a point on the manifold $M$ (thick black line). A figurative illustration of the contours of
the kernel centered at the same point is also shown to illustrate the bandwidth of the kernel.}
\label{fig:hess_pca_drawing}
\end{figure}

If we look at the expression for the Hessian matrix, we notice that it
turns out to be proportional to the sample covariance matrix at $x$ with the variance
normalized wrt the kernel size $\sigma^2$. Like the kernel density
gradient, the weight of the points taken into account in calculating the
covariance is governed by the kernel size, leading to a \emph{local}
sample covariance estimate at $\x$. We note that if a very large kernel size, in the order of $\var\{X\}$, is used, the Hessian
matrix will resemble the standard sample
covariance matrix and eigendecomposition would yield standard PCA\@.
  Considering this we can interpret the Hessian matrix as fixing and aligning
a local Gaussian distribution to a local subset of the data. For a one-dimensional curve
with multidimensional additive noise the local Gaussian will clearly
have a strong correlation along the curve wich will be reflected in the
eigenvectors of the local Hessian estimate.
A simple illustration of this is shown in Figure~\ref{fig:hess_pca_drawing}.
  Considering the case of a principal \emph{surface} we can continue our
principal component analogy. Assuming the points have been sampled from
some surface with some level of noise, the kernel density Hessian
estimate at $\x$ will in practice calculate a covariance matrix estimate based on the points within
the range covered by the kernel centered at $\x$. Unless the noisy points
that does not lie on $M$ completely dominate, we expect that
most of the variance is concentrated along the manifold, and thus the
largest eigenvectors according to eigenvalue will span the local tangent space $T_\x M$.

Since by definition the kernel density gradient, $g(\x)$, lies in the span of
$d$ Hessian eigenvectors if $\x$ is on the $d$-dimensional ridge, the corresponding $d$
eigenvectors forms a natural basis for local coordinates
in $T_ \x M$. Consequently the parallel Hessian eigenvectors, $Q_{||}$, can be
calculated for each $\x$ that lies on the density ridge, also for out of
sample points, and will thus form an approximate tangent bundle on the ridge estimate of the manifold, $\hat{R}$.

%
\subsection{Properties of density ridges}
\label{sub:conn_manifold_ridge}
Throughout this section we consider a single
connected submanifold $M$ of Euclidean space without boundary that satisifies the properties given in
\citep{Genovese2014}.

Recall that a principal \emph{curve} can be approximated by a one-dimensional density ridge and a principal
\emph{surface} by a $d$-dimensional density ridge. Some
properties are general, while some apply only to principal curves, or
vice versa. We start by introducing the gradient flow, which we use in both cases later in this paper.

An integral curve of the positive gradient flow is a curve $\lambda(t)$
such that $\lambda'(t)=\nabla f(\lambda(t))$.
For every point $x_0$ in the domain of $f$ there exists an integral
curve that starts at $x_0$. Given a critical point\footnote{A local mode
in the case where the function $f$ is a probability density function.} $\mathbf{m}$,
$\nabla(\mathbf{m})=0$, all points that have integral curves that
converges to the critical point are said to lie in the \emph{basin of
attraction} of $\mathbf{m}$~\citep{arias2013estimation}. This property is
the foundation of mode based clustering algorithms, e.g\
mean shift clustering~\citep{comaniciu2002mean, myhre2015consensus}.

The idea of a basin of attraction can be extended to hold for density
ridges as well. Given a density ridge of dimension $d$ a local mode will
still satisfy the criteria for being a point on the ridge. The
differential equation -- equation~\eqref{eq:ode_pc} on page
\pageref{eq:ode_pc} -- for projecting points towards the density ridge
always follows the gradient, so the density ridge(s) can be divided into
basins of attraction based on the original points before projection.
This naturally divides a manifold estimated by a density ridge into
non-overlapping subsets.
Closely related: a \emph{chart} of a smooth manifold is a diffeomorphism from a
neighborhood $U$ on the manifold to $V\subset\mathbb{R}^d$, $\phi:U \to
V$. The local density ridge unwrapping algorithm, presented in
Section~\ref{sec:principal_curve_unwrapping} can be considered an
algorithm for learning the map $\phi$, and we thus have a chart for each
separate basin of attraction of the manifold. We note that
in classical differential geometry it is common to consider smoothly
overlapping charts~\citep{tu2008introduction}, which is not the case in this work.

We end this section with a short summary of important
properties and observations for principal curves and surfaces separately.
\\
\\
\emph{Principal curves}:
\begin{itemize}
  \item The gradient flow integral curves $\gamma$ are geodesics, allowing for
    parallel transport, this will be introduced in Section~\ref{sec:parallel_transport}.
  \item The gradient flow separates the manifold into non-overlapping
    regions $C_i$ - attraction basins, such that $M = \bigcup\limits_{i=1}^{\#\text{charts}}C_i$
  \item Curves does not have intrinsic curvature, and thus principal curves can always
    be unfolded isometrically.
  \item $T_{\x}M=\Span\left(g(\x)\right)$, $g(\x) = \lambda Q_{\parallel}(\x)$. Thus, the gradient flow curve starting on points on the manifold is completely contained along the one-dimensional manifold, $\gamma \in M$.
\end{itemize}

\emph{Principal surfaces}:
\begin{itemize}
  \item The gradient flow separates the manifold into non-overlapping
    regions - attraction basins. Again $M = \bigcup\limits_{i=1}^{\#\text{charts}}C_i$.
  \item The gradient flow curves $\gamma$ are not necessarily geodesics. If the initial point is 
    \emph{on} the manifold the gradient flow integral curves will be restricted to the manifold. Thus, if $\gamma(0) \in M$, $\gamma(t) \in M \:\; \forall{t}$. 
  \item Curvature must be introduced, isometric unfolding can no longer
    be guaranteed for arbitrary surfaces.
  \item $T_{\x}M=\Span\left(Q_{\parallel}(\x)\right)$. Thus the direction of the gradient determines the gradient flow along the manifold. 
\end{itemize}

\section{Proposed method: Principal curve unwrapping by tracing the ridge}
\label{sec:principal_curve_unwrapping}
In this section we present the first part of our suggested algorithm for unwrapping
manifolds, which deals with data distributed on and along a \emph{one-dimensional manifold}.
We start by recalling that the probability density function along a local
principal curve/one-dimensional density ridge will contain local maxima.
For a point $\x\in \hat R$ the gradient, $g(\x)$, is by definition
orthogonal to all except one of the eigenvectors of the local
Hessian $H(\x)$. Thus any gradient on the ridge which is not exactly zero
will point along the ridge towards a local mode. This restricts
the gradient flow,~\citep{arias2013estimation}, along the ridge to stay
on the ridge, and thus we can formulate a differential equation that
sends the points on the principal curve, along the
curve/ridge towards the local mode. We note that Arias-Castro et al.\ showed that a
gradient ascent scheme for estimating the gradient flow lines converges
uniformly to the true integral curve lines~\citep{arias2013estimation}.
Then, by using either a gradient ascent scheme
-- like in mean shift clustering -- or as in this work a
differential equation solver, we can estimate the gradient flow lines
along a ridge, and by
calculating the piece-wise Euclidean distances between the steps of the gradient flow we can
estimate the distance from a point to a local mode along the ridge. Since this is the only possible path
between the point on the ridge and the local mode it converges to, it is
by construction a geodesic. These geodesic distances along the
principal curve we will take as local coordinates in the basin of
attraction of the local mode, $\mathbf m_i$, with the smallest eigenvector
of $H(\mathbf m_i)$ as basis\footnote{The intuition is that the curvature as described by the Hessian eigenvalues should be lower \emph{along} the ridge, than orthogonally off the ridge. Thus the choice of lowest eigenvector as basis.}.
We state the differential equation for
following the gradient along the ridge as follows:
\begin{equation}
\label{eq:ode_ridge}
\frac{\text d \mathbf z(t)}{\text d t} = g(\mathbf z (t)),
\end{equation}
where $\mathbf z(0)=\mathbf y(T)$, $T$ such that $V_T^Tg(\mathbf
y_T) \approx 0$.

Solving this equations gives coordinate lengths along the ridge and thus yields a complete local coordinate description for every
point in the local basin of attraction of $\mathbf m_i$. 
  %
\begin{algorithm}[tp]
  \caption{Local density ridge unwrapping.}
  \begin{algorithmic}[1]
    \REQUIRE Input points, $\x_i$, projected to a one-dimensional density ridge
    $\hat R$ by solving Equation~\eqref{eq:ode_pc}.
    \STATE Compute the trajectory, $Z_i$, from each point on the density ridge
    to the associated mode by solving Equation~\eqref{eq:ode_ridge}.
    \STATE For each trajectory $Z_i = \begin{bmatrix}\mathbf z_i^1 &\mathbf z_i^2 &\cdots& \mathbf z_i^N \end{bmatrix}$ calculate:
    \begin{equation}
      \label{eq:stepwise_length_gradient}
      \mathbf c_i = \sum^{N-1}_{j=1} \|\mathbf z_i^{j} - \mathbf
      z_i^{j+1}\|_2
    \end{equation}
    \ENSURE Local manifold chart coordinate $\mathbf c_i$.
  \end{algorithmic}
  \label{alg:ridgeTracing}
\end{algorithm}
In practice we solve this using a Runge Kutta fourth and fifth order
pair, also known as the Runge-Kutta-Fehlberg
method(RKF5)~\citep{fehlberg1969low}, implemented
in MATLAB\@. This is an adaptive stepwise solver,
which yields benefits both in terms of speed and accuracy. For density
ridge estimations we use the kernel density estimate from Equation~\eqref{eq:kde}, its
gradient, Equation~\eqref{eq:kdde}, and
the Hessian matrix from Equation~\eqref{eq:kde_hess} as presented in the
previous section. The ridge projection for each data point considered is
independent from other projections, so processing each point in parallel is possible. In this
work we have used the parallel processing toolbox in MATLAB to achieve a local
speedup proportional to the number of cores on the computer. 

We have summarized the steps necessary to perform local principal curve
unwrapping in Algorithm~\ref{alg:ridgeTracing}.

\begin{figure}[htbp]
\begin{center}
  \includegraphics[width=.8\columnwidth]{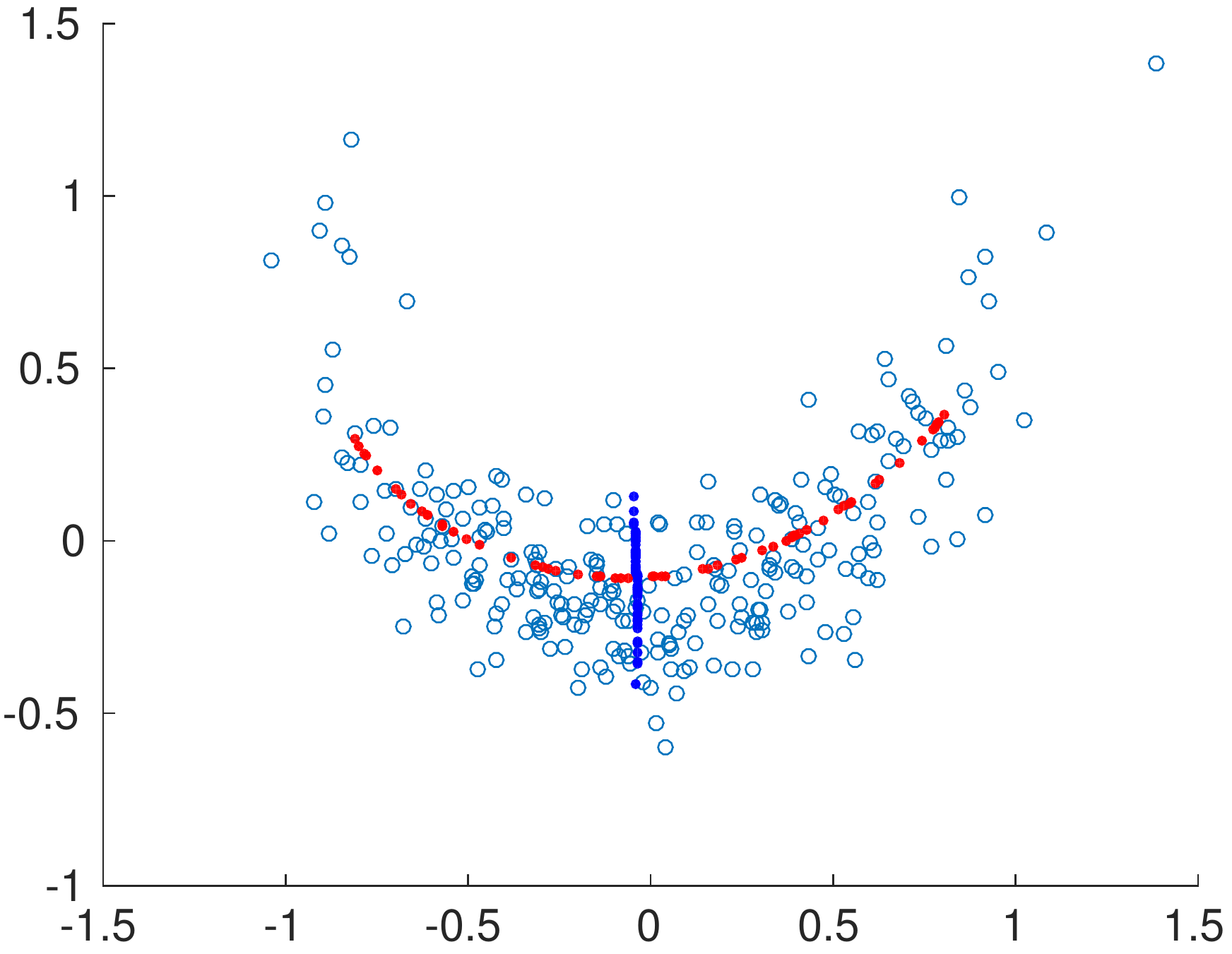}
\end{center}
\caption{Nonlinear and noncovex data set with density ridges shown in
red(top) and blue(second).}
\label{fig:nonlinear_and_nonconvex}
\end{figure}

\begin{figure}[htbp]
\begin{center}
  \includegraphics[width=.6\linewidth]{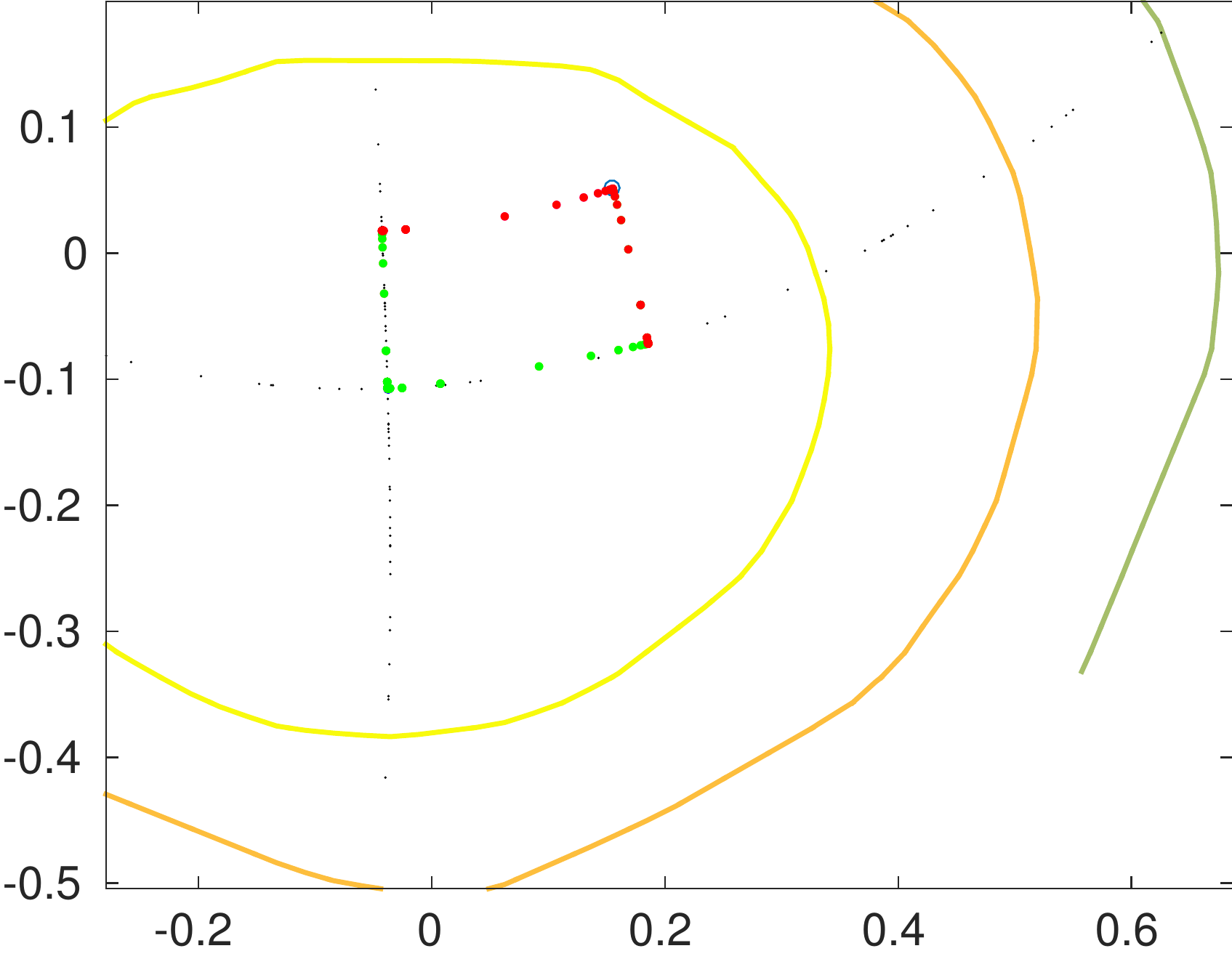}
\end{center}
\caption{A single data point projected to the two one-dimensional density
ridges, trajectories shown in red and then along the gradient to the local mode(green). The
density ridges as shown in Fig~\ref{fig:nonlinear_and_nonconvex} are marked with small
black dots.}
\label{fig:single_point_projection}
\end{figure}

We conclude this section with an illustration of
Algorithm~\ref{alg:ridgeTracing}.
In Figure~\ref{fig:nonlinear_and_nonconvex} we see a two dimensional
sample from a Gaussian distribution superimposed with an underlying
one-dimensional nonlinear structure. The two orthogonal density ridges
found using the RKF45 solver are shown in red and blue dotted lines.
In Figure~\ref{fig:single_point_projection} we see the stages in
Algorithm~\ref{alg:ridgeTracing} for a single data point from
Figure~\ref{fig:nonlinear_and_nonconvex}. First the point is projected
to each of the two orthogonal density ridges using the RKF45 algorithm, shown in
red. Second, the point is projected along the ridge by following the gradient towards the
local mode shown in green. We see that by calculating the piece-wise
Euclidean distances between the gradient ascent steps -- the green
points -- we get approximate coordinate distance along the intrinsic nonlinear
geometry.
%
\section{Parallel transport along density ridges}
\label{sec:parallel_transport}
The local density ridge unwrapping algorithm presented in the previous
section is only valid for local regions of the manifold.
This implies that the local unwrapping algorithm can result in several
charts for a single connected manifold if there are multiple local
maxima along the ridge. So given a manifold $M$ embedded in $\mathbb R^d$ and a ridge approximator
$\hat R$ consisting of a set over non-overlapping attraction bases, we
can, by using Algorithm~\ref{alg:ridgeTracing}, obtain a local coordinate chart for each of
the local attraction basins.

In the current setting where the underlying manifold
describing the
nonlinearity can be described by a one-dimensional ridge, the
trajectories along the manifold can trivially be considered
geodesics\footnote{The shortest path between two points on a one
dimensional curve must be along the curve itself.}.
This allows the use of \emph{parallel
transport} to transport vectors from one mode to another if they are connected by
a principal curve/one-dimensional ridge~\citet{lee1997riemannian}. In this work we will not go
further into the differential geometric framework for parallel transport --
see for example~\citep{lee1997riemannian} -- , but rather develop a
discrete empirical algorithm for sewing together local charts stemming
from Algorithm 1 by translation along geodesics.

We make the following assumptions:
\begin{itemize}
  \item If there is a one-dimensional ridge between two local modes, it
    can by definition be considered a geodesic.
  \item The local neighborhood around a density mode can be described by
    trajectories along local orthogonal density ridges.
\end{itemize}
To form a global representation of a manifold $M$, we need to relate the
coordinate charts of the manifold to each other.

To do this we select a reference mode along the principal curve and
translate the other local coordinate systems connected to the same ridge
towards the reference mode. This is a straight forward operation, but there are a
few issues that needs to be considered. First, each local system needs
a basis and consequently the bases along the curve needs to be checked
in case they are not equally aligned to the ridge.
Second, once a reference mode and basis is chosen, the translation of
other charts towards that mode can give ambiguities in terms of sign, if
two charts are translated towards the same mode from different
directions, a positive and negative direction needs to be defined.

\begin{figure}[htpb]
\centering
\begin{subfigure}{.4\linewidth}
\centering
  \includegraphics[width=\linewidth]{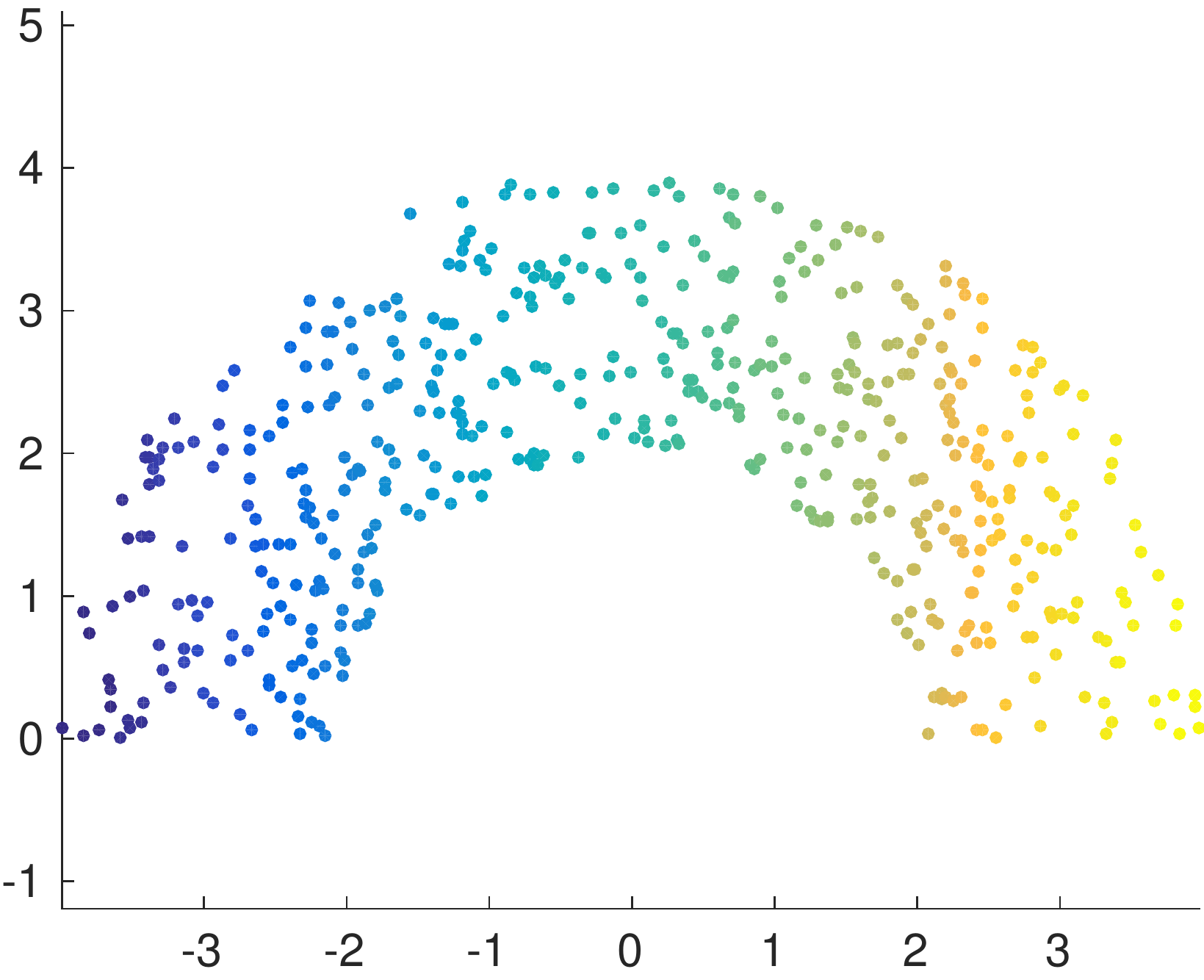}
\caption{Uniform data perturbed with a nonlinear shape.}
\label{fig:one_moon_input}
\end{subfigure}
\qquad
\begin{subfigure}{.4\linewidth}
\centering
  \includegraphics[width=\linewidth]{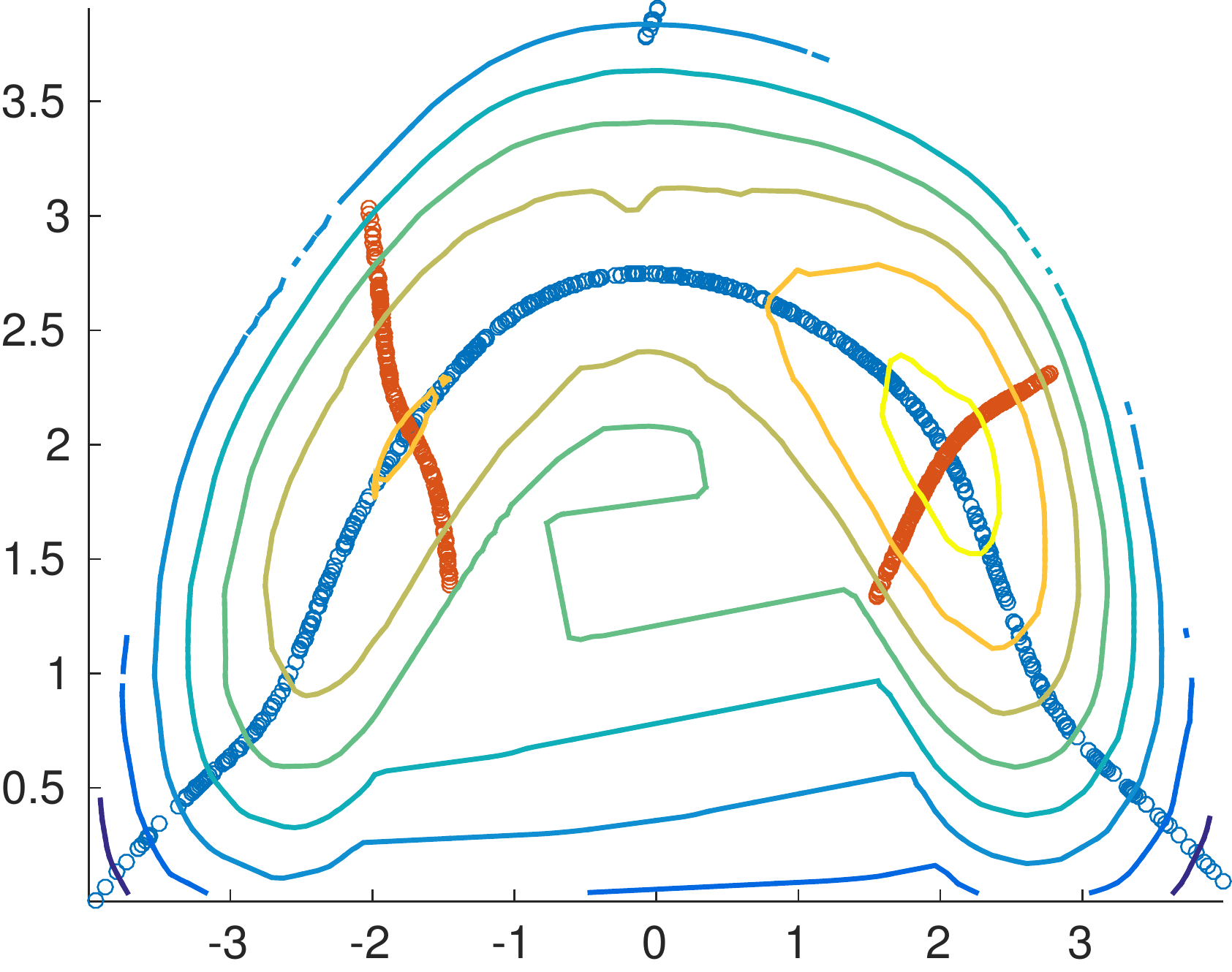}
\caption{Nonlinear uniform data set with kernel density estimate
contours and orthogonal one-dimensional ridges shown.}
\label{fig:one_moon_dens_curves}
\end{subfigure}
\caption{Nonlinear uniform  data set.}
\label{fig:one_moon_all}
\end{figure}

\begin{figure}[htp]
\begin{center}
  \includegraphics[width=2.5in]{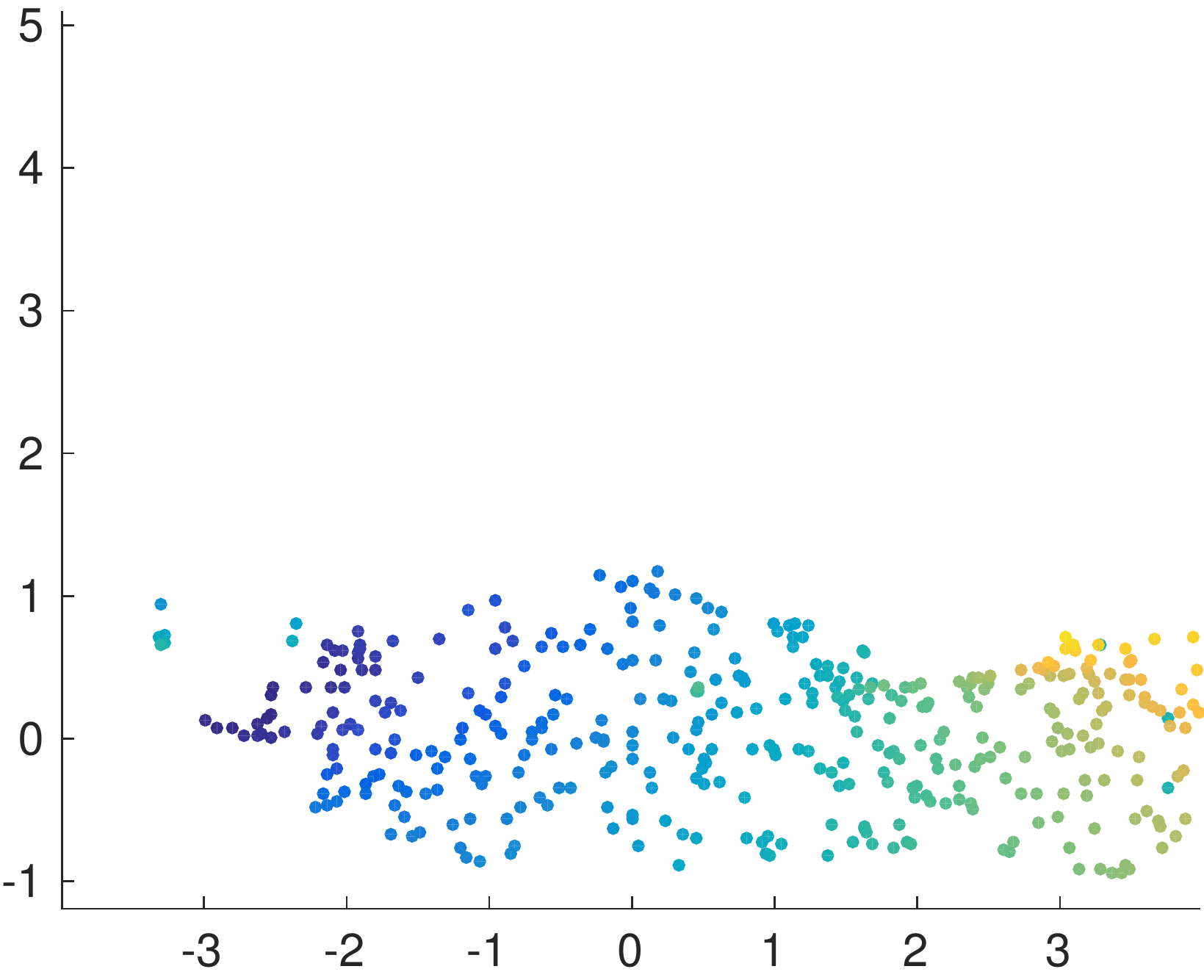}
\end{center}
\caption{Result after applying the density ridge translation algorithm
on dataset shown in Figure~\ref{fig:one_moon_input}. Color coding
follows from input data set.}
\label{fig:one_moon_stitched_charts}
\end{figure}

Given a manifold $M$ and its ridge estimator $\hat R$, the set of local
modes, $\left\{ \mathbf m_i \right\}_{i=1}^{m}$, along the ridge, the gradient and
Hessian at each point in both $\hat R$ and $\left\{ \mathbf m_i
\right\}_{i=1}^{m}$, we define the following:
\begin{definition}
  At a given a reference mode $\mathbf m_r$, we define the $d$ eigenvectors
  of the Hessian of the kernel density estimate at $\mathbf m_r$ as the basis of the
  unwrapped coordinates.
\end{definition}
Using this we can check the sign of the determinant of the other
bases along $\hat R$ and change signs if the bases have a different
orientations. This ensures a consistens orientation, unless the manifold
is not orientable (e.g.\ a M\"obius strip)~\citep{spivak1965calculus}.
\begin{definition}
  Given a reference mode $\mathbf m_r$, we define the direction towards
  the \emph{closest} local mode along the ridge $\hat R$ as the
  \emph{positive} direction.
\end{definition}
As the ridge connecting two local modes is one-dimensional,
tangent vectors by finite differences from mode to reference mode can be
used to identify the relevant directions.
Even though the ridge estimate $\hat{R}$ and local modes are known, we do not know if two
local modes are connected and the order of the points contained in a
local attraction base. As preliminary solutions to this, we use
Dijkstra's algorithm for identifying the path along the ridge from mode
to mode, and use that path to calculate finite difference tangent
vectors~\citep{dijkstra1959note}. To check if two modes are connected we use the \emph{flood
fill} algorithm~\citep{heckbert1990seed}.
  Both algorithms operate on a
nearest neighbor graph which in this case consists of the every input
point projected to the density ridge estimator as well as the set of
local modes found by following the gradients along the ridge for all
points.
  The flood fill algorithm simply checks
the number of connected components in a graph. Thus if there are several
components along a one-dimensional density ridge, either the
neighborhood graph must be expanded or the ridge is disconnected.
  Dijkstras algorithm calculates the shortest path along nodes in a
sparse graph. In ridges of dimension higher than one, this would lead to
poor estimates, but since all points along the one-dimensional ridges
lie on a curve this is not a problem. In practice, the Dijkstra
algorithm ends up simply sorting the points along the density ridge.
Thus we have all steps needed to translate a local chart along the connecting density ridge and obtain a global unwrapping.
  The density ridge translation algorithm is summarized in in
Algorithm~\ref{alg:parallelTransport}.
\begin{algorithm}[t!]
  \caption{Density ridge translation algorithm}
  \begin{algorithmic}[1]
    \REQUIRE Ridge estimator $\hat R$, local modes $\mathbf m_i
    $ and unwrapped coordinates $\mathbf c_i$ corresponding to
    each local mode.
    \STATE Create a nearest neigborhood graph consisting of $\left\{
    \mathbf m_i \right\}_{i=1}^m$ and $\hat R$.
    \STATE Select a reference mode, $\mathbf m_r$, and calculate the
    geodesic path, $\gamma_{i,r}(t)$to all other modes $\mathbf m_i$ along the ridge using Dijkstras
    algorithm.
    \STATE Chose the mode closest to $\mathbf m_r$ and define the direction along
    $\hat R$ towards that mode as the positive direction.
    \FORALL{Local coordinate charts, $\mathbf c_i$}
    \STATE {Calculate fininite difference tangent vectors along the
    geodesic path, $t_l = \gamma_{i, r}(t_l) - \gamma(t_{l-1})$.}
    \STATE {Select the coordinate direction of the translation. E.g.\
    the first coordinate if translation is along the first principal
    curve/density ridge.}
    \STATE {Translate the local coordinates along the geodesic by adding
    the tangent vector length to the chosen coordinate
    direction.\begin{equation}
      \mathbf c_{translated} = \sum_l\mathbf c_i^j + ||t_l||
    \end{equation}}
  \ENDFOR
  \STATE Concatenate all translated charts into a global representation $\mathbf Z$.
    \ENSURE Global representation, $\mathbf Z$ of concatenated and connected charts.
  \end{algorithmic}
  \label{alg:parallelTransport}
\end{algorithm}
We end this section with an example. In
Figure~\ref{fig:one_moon_input} and Figure~\ref{fig:one_moon_dens_curves}
we see a two-dimensional uniform sample superimposed with a nonlinear
structure, its density and one-dimensional orthogonal density ridges.
The color coding of Figure~\ref{fig:one_moon_input} represents the order of the data points in the
horizontal direction.
Figure~\ref{fig:one_moon_stitched_charts}
shows the results of the density ridge translation algorithm. We see
that the underlying uniform distribution is revealed. There are some
artifacts at the boundary of the manifold, this is probably due to the
smoothness of the kernel density estimate and the fact that the
kernel density estimate is close to a mixture of two Gaussians. There will always be a tradeoff
between bias in the principal curves and non-smooth, but well fitting
ridges. In other words, the smoother the curve, the larger the bias and the closer the density
is to a mixture of a low number of Gaussians~\footnote{Recall that a
kernel density estimate in the extreme case of very small or very large bandwidth is just
a mixture of either $N$ Gaussians or a single Gaussian respectively.}.
We also have to note that the origin of the data structure has been
moved to $\left[ 0, 0 \right]^T$. This is due to the arbitrary choice of
reference mode. One could keep the original reference mode as origin of
the unwrapped representation,
but this would not affect the result other than a translation.






\section{Expansion to $d$-dimensional ridges}
\label{sec:n_dim_ridges}
In the case where the underlying manifold cannot be
described by a principal curve, a density ridge of higher dimension, a
principal \emph{surface}, must be
considered. 
This causes a problem with our previous approach: the gradient, $g(\x)$, of a point on a 
$d$-dimensional ridge lies in the span of $d$ eigenvectors of $H(\x)$ such that following the gradient flow will not lead to unique coordinates on the manifold. 
So for example applying Equation~\eqref{eq:ode_ridge} on a $d$-dimensional ridge would lead to a ridge-constrained mean shift, i.e.\ gradient flow trajectories
that lies completely on the ridge. 
Depending on how the data is
distributed along the manifold as well as geometric properties such as
curvature, a $d$-dimensional ridge cannot in general be decomposed into
$d$ orthogonal one-dimensional ridges.

Also, in general, the curvature of the manifold has to be considered in higher dimensions -- we can no longer be sure that an
isometric unfolding exists. Furthermore, comparing with the previous
section, a path between two points is also not necessarily a geodesic
even if the path lies completely on the manifold. Thus the parallel
transport/translation analogy cannot be directly used.

Based on the aforementioned, we propose an alternative unwrapping scheme for the density ridge
framework in the case of $d$-dimensional nonlinear structures. 
We recall that in the one dimensional case we unfold local charts of the density ridge, and if there are several charts, transport them along the ridge to get a global representation. 
In the $d$-dimensional case we cannot directly unfold the local charts by following the gradient flow, so we insted do a local linear approximation by projecting the data points in the local basin of attraction to the tangent space of the local mode of the chart, presented in Section~\ref{sub:local_approximation}.
Once we have linear approximations of all local charts of the manifold we calculate approximate geodesics and use parallel transport to send all charts to a reference mode. Finally, the unwrapping is done by sending the charts back along the geodesic by parallel transport, while at the same time constraining the piece-wise steps of the geodesic to lie in the tangent space of the reference mode, presented in Section~\ref{sub:Combining geodesics and local approximations}. 
To approximate the geodesic we use an algorithm proposed by Doll{\'a}r et al.~\citep{dollar2007non}, presented in Section~\ref{sub:approximating_geodesics}.

\subsection{Local linear approximation}
\label{sub:local_approximation}
As calculating distances along gradient lines that cover the
$d$-dimensional manifold will not give consistent orthogonal coordinates
we will expand our approach and use local linear approximations. The
approximations will consist of a local flattening centered at the local modes of
the $d$-dimensional density ridge. 
By flattening at the local modes we get charts centered at the points of highest probability and the points within the chart are points where the gradient flow converges to the same mode. This ensures that the charts represents points that are close in density and that the charts are limited in extension since the gradient flow lines are restricted to locally smooth areas of the manifold. 
Following this, the flattening process consists of projecting the data points corresponding to a
local mode to the space spanned by the Hessian eigenvectors that span
the gradient (the local tangent space):
\begin{equation}
  \mathbf c_{i} = Q_{||}(\mathbf m_i)Q_{||}(\mathbf m_i)^T\mathbf x_{i},
\end{equation}
where $\mathbf c_i \in Q_{||}$ is a local coordinate for a point
$\mathbf x_i$ that has converged to mode \#$i$, $\mathbf m_i$. This is
equivalent to an approximate inverse exponential map on the manifold at
$\mathbf{m}_i$ -- the log map. In this manner we get a local
linear approximation for each local mode in the data set. 

Also, comparing with the previously noted similarities between the Hessian of the KDE estimate and the local sample covariance matrix, we can view this local flattening directly as local PCA where the neighborhood is determined by the coverage of the kernel bandwidth.
\subsection{Approximating geodesics}
\label{sub:approximating_geodesics}
To obtain a global coordinate representation on a connected manifold
that has a set of approximated local coordinate charts we proposed
parallel transport of tangent space coordinates towards a reference
mode. In the one-dimensional case a geodesic was directly available in
the principal curve. In the $d$-dimensional case we need a numerical
scheme to approximate the geodesic. In this paper we reframe an idea
presented by Doll{\'a}r et al. \citep{dollar2007non}, to approximate
geodesics on the manifold. The idea is to minimize the distance between two points, while at the same time keeping the starting points and endpoints fixed and making sure that all points on the path lies approximately on the manifold.
  Formally, given a sequence of points $\left\{\gamma_i\right\}_{i=1}^n \in M$ between
two points on the manifold $\x$ and $\mathbf y$ we can formulate the problem of finding a shortest path constrained to the manifold as follows:
  \begin{equation}
\begin{aligned}
& \underset{\gamma}{\text{minimize}}
& & \sum^{n}_{l=2} || \gamma_{l} - \gamma_{l-1}||^2 \\
& \text{subject to}
& & \gamma_1 = \x,\; \gamma_n = \mathbf y, \; \gamma \in M.
\label{eq:geod_opt}
\end{aligned}
\end{equation}
To optimize \eqref{eq:geod_opt} the path $\gamma$ is initialized using
Dijkstra's algorithm and further discretized with linear interpolation between the $n$ given points. Then minimization is performed by alternating between gradient descent to shorten distance and
density ridge projection, equation \eqref{eq:ode_pc} to ensure that points stay on the manifold. In addition we use another idea from the work of Doll{\'a}r et al.~\citep{dollar2007non} for fast out of sample projections. After a selection of points have been projected to the density ridge, by equation \eqref{eq:ode_pc}, the tangent space of the ridge estimator is at each point $\x$ spanned by the parallel Hessian eigenvectors $Q_\parallel(\x)$. To project a new out-of-sample point $\x_o$ orthogonally to the ridge, $||\x_o - \x_r||^2$, where $\x_r$ is a point on the ridge, needs to be minimized. This can be solved by setting $\x_r$ to the closest point of $\x_o$ on the manifold and then performing gradient descent as follows:
\begin{equation}
    \x_r \leftarrow \x_r + \alpha Q_\parallel(\x_r)Q_\parallel(\x_r)^T \left(\x_o -\x_r\right)
    \label{eq:out_of_sample}
\end{equation}

\begin{figure}[htbp]
  \begin{center}
    \includegraphics[width=3.5in]{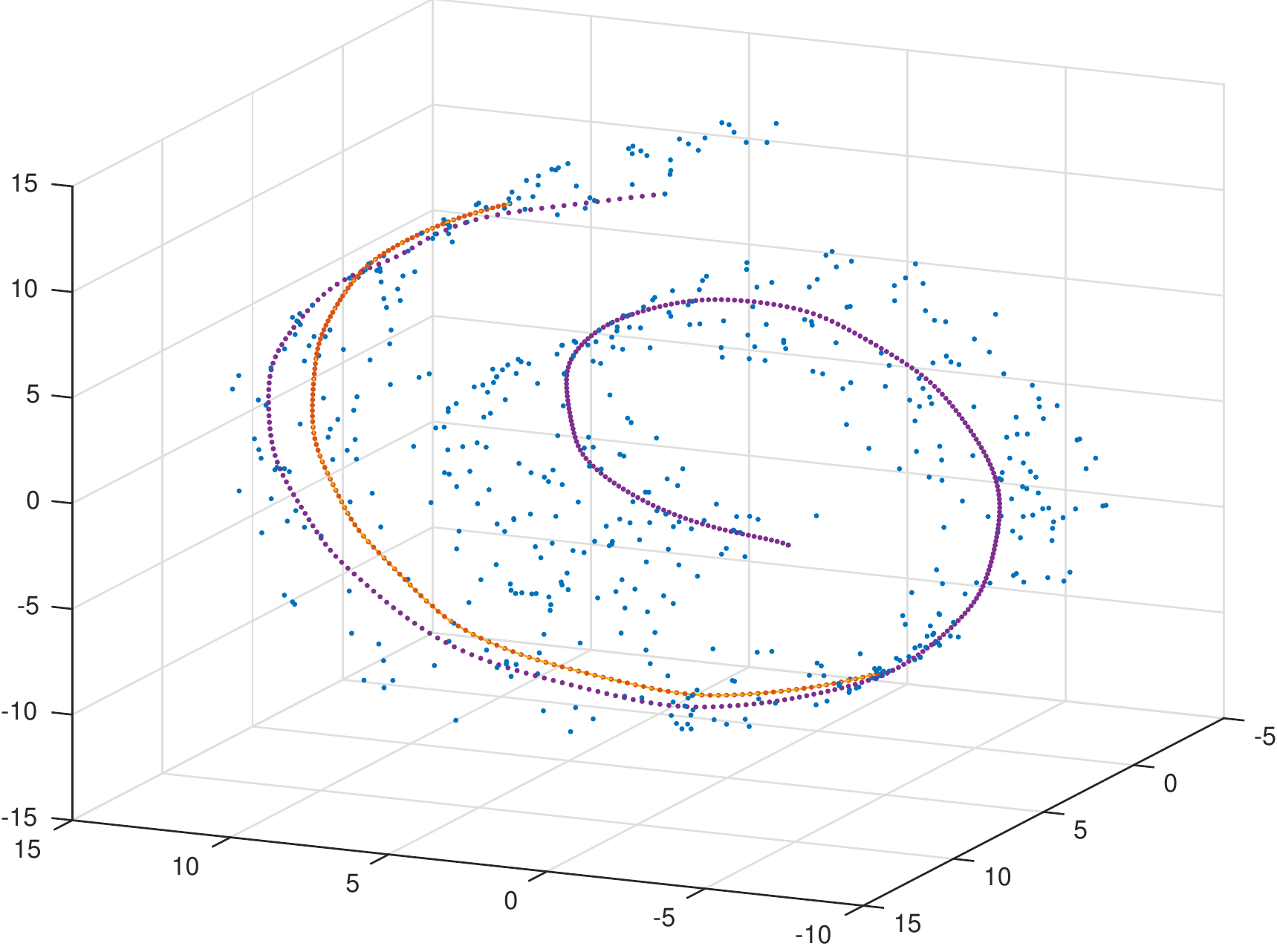}
  \end{center}
  \caption{Swissroll dataset showing two different geodesics
  approximated by the alternating least squares scheme in equation \eqref{eq:geod_opt}.}
  \label{fig:nDimRoll}
\end{figure}

An illustration of the geodesics found by the alternating optimization of equations \ref{eq:geod_opt} and \ref{eq:out_of_sample} on the `swiss roll' dataset,
\citep{tenenbaum2000global}, is shown in Figure~\ref{fig:nDimRoll}. 

\subsection{Combining geodesics and local approximations}
\label{sub:Combining geodesics and local approximations}
Once we have local approximations in place we can in principle directly
unfold manifolds that are isometric. We recall that if a manifold is
isometric it has no intrinsic curvature and can be covered by a
global chart, in other words: the diffeomorphism $\phi:\mathbb R^d \to M$ preserves distances. Still, if we use the algorithms presented previously in this paper we cannot in general find a single global chart. Combining the two previous sections we have the following: an estimate of the underlying density ridge $\hat{R}$ which is close to the true manifold, a set of local linear approximations $C$ of each local basin of attraction and the possibility of estimating geodesics between points in $\hat{R}$ by Equation~\eqref{eq:geod_opt}.

Using these tools we can construct an algorithm for stitching together the local
approximations $C$ to obtain a global unwrapping. Below we present a short outline of the algorithm, see the appendix for a complete description of the algorithm:
\begin{enumerate}
    \item For each mode, project the points in the local attraction basin to the space spanned by the parallel Hessian eigenvectors.
    \item Among all modes, select a reference mode and calculate the piece-wise geodesic distance from all other modes to the reference mode
    \item Transport the local approximations towards the reference mode by
    iteratively translating and projecting to the local tangent space along
    the geodesics. Algorithm~\ref{alg:dDimParallelTransport} is presented in Appendix~\ref{app:parallelTransport}.
    \item Once all coordinates are centered on the reference mode send them
    along the tangent vector of the reversed geodesic projected to the local
    tangent space of the reference mode. This is presented in Algorithm~\ref{alg:isometricUnfolding} in Appendix~\ref{app:unfolding}.
\end{enumerate}
We conclude this section with a short example. Again we turn to the swiss roll example, where a plane is deformed into a swiss roll shape embedded in $\mathbb R^3$. See Figure~\ref{fig:swiss_unrolled} for the data and unwrapped result.

\section{Numerical experiments}
\label{sec:numerical_experiments}
We now show a selection of experiments to illustrate the presented
frameworks. Both real and synthetic data is used. In the real data sets we preprocess the data with pca and reduce the dimension to two or three to enable visual evaluation.

We start by illustrating the density ridge projections using rkf45 and also how the estimates are influenced by the choice of kernel size. Then we illustrate density ridge unwrapping and translation on synthetic data sets and images of faces and the mnist handwritten digits, ~\citep{lecun1998gradient}\footnote{http://www.cs.nyu.edu/~roweis/data.html}.

In all experiments we assume that the dimension of the underlying manifold is
known.
\subsection{Density ridge estimation}
\label{sub:dre_experiments}
We start by illustrating the density ridge estimation algorithm using
the rkf45 scheme implemented in matlab. This is an adaptive
solver for initial value problems that reduces the parameter choices to
only the kernel size of the kernel density estimate. To emphasize the
algorithm we focus on examples in three dimensions so they easily can be visualized.

\begin{figure}[htbp]
\begin{center}
  \includegraphics[width=2.5in]{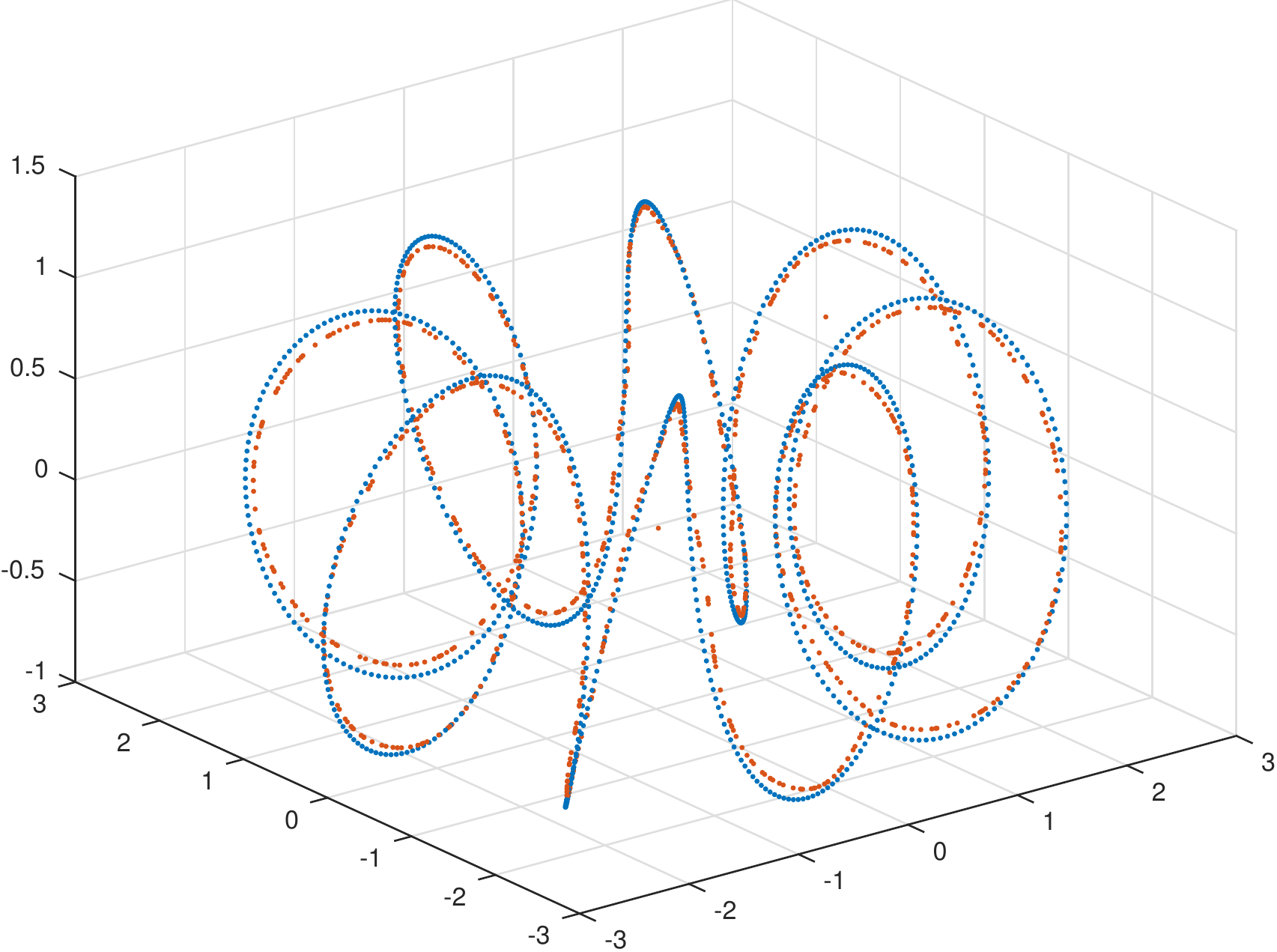}
\end{center}
\caption{Helix data set and density ridge projection. True data set
without noise shown in blue and density ridge projection with lowest
mean square error in red.}
\label{fig:helix_and_ridge}
\end{figure}

We start with the `helix' data set, available in the \emph{drtoolbox} of
van der maaten~\citep{van2009dimensionality}. It is a one-dimensional
manifold embedded in $\mathbb R^3$, sampled uniformly with gaussian
noise. We run the rkf45 scheme over a range of kernel sizes and select
the one with the lowest mean square error. The result is shown in
figure~\ref{fig:helix_and_ridge}.

\begin{figure}[htbp]
\centering
\begin{subfigure}{.4\linewidth}
  \includegraphics[width=\linewidth]{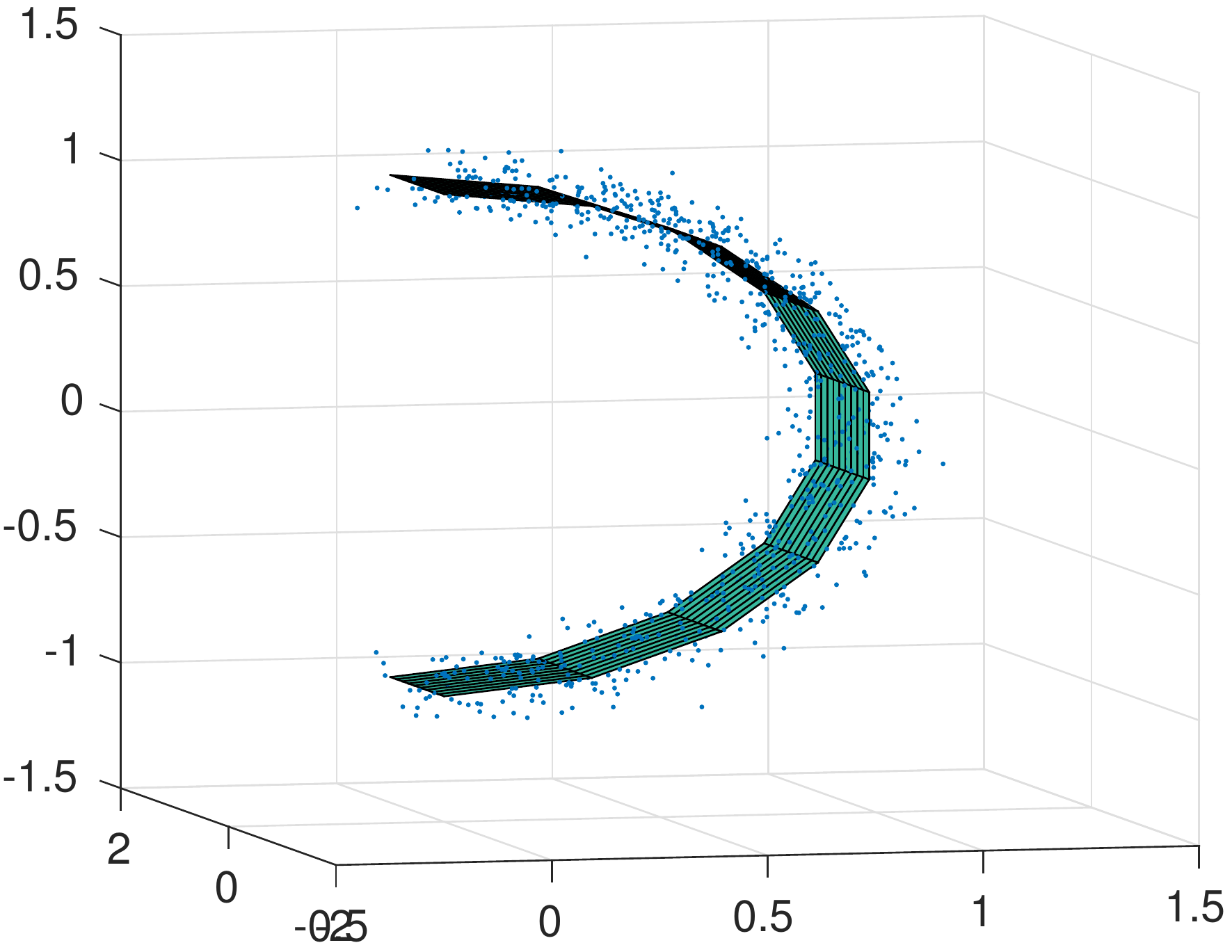}
\caption{Half cylinder manifold(solid) sampled with noise(dots).}
\label{fig:bent_manifold_data}
\end{subfigure}
\qquad
\begin{subfigure}{.4\linewidth}
  \includegraphics[width=\linewidth]{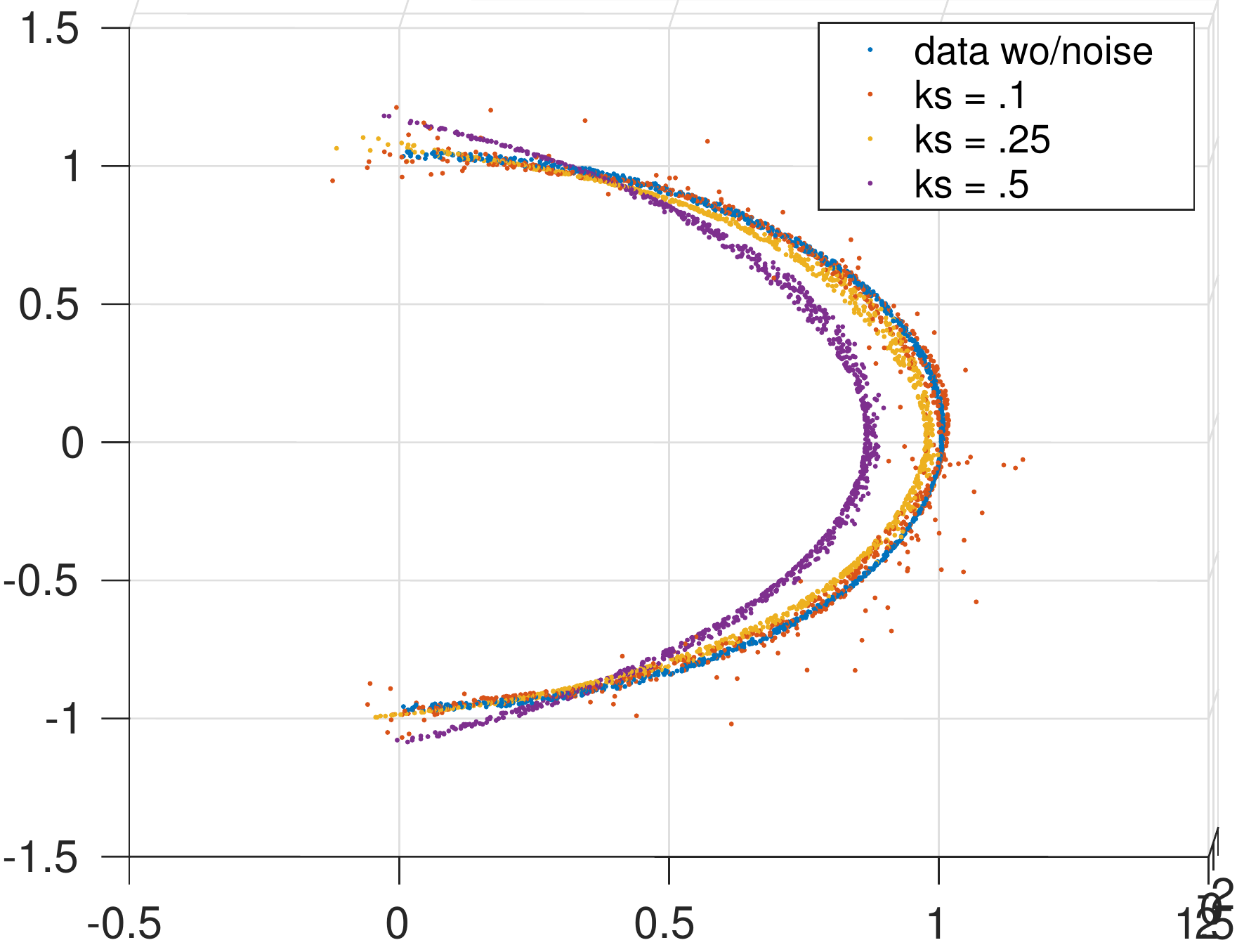}
\caption{Estimated density ridge of half cylinder manifold. The ridge is
shown for a range of kernel sizes to illustrate the bias introduced. The true manifold is shown in blue dots.}
\label{fig:bent_manifold_bias}
\end{subfigure}
\caption{Noisy half-cylinder manifold estimated at different kernel sizes.}
\label{fig:bent_manifold}
\end{figure}

The next example shows a two dimensional plane bent into a half-cylinder
shape. A uniform sample is drawn from the surface and zero mean gaussian
noise is added, see figure~\ref{fig:bent_manifold_data}. We show the result using different kernel sizes in figure~\ref{fig:bent_manifold_bias}, to visually confirm that the bias of the density ridge estimate increases as the kernel size increases. the largest kernel size gives the most smooth result, but the curvature of the half-cylinder is reduced. for the smaller kernel sizes we see that the ridge estimator is closer to the true manifold, but that many of the noisy points become local maxima and is not projected to the density ridge. 

\subsection{influence of kernel size}
\label{sub:influence_of_ks}
The choice of kernel bandwidth is very important in estimating density
ridges, and can have great influence on the results.

To illustrate the influence of kernel size we sample uniformly from a sphere,
$\mathbb{s}^2$, and add different levels of $n(0,\varepsilon_i)$
noise. In table~\ref{tab:ks} the mean squared error between the true
manifold and the ridge for the different kernel bandwidths and levels of
noise is shown.
\begin{table}[h]
  \centering
  \caption{Mean squared error of ridge estimates for different levels of
  noise($\varepsilon$) and kernel sizes($\sigma^2$).}
  \begin{tabular}{ccc}
    $\sigma^2$ \textbackslash $\varepsilon$ & .05 & .1 \\
    \hline
    .1    &0.0103    & 0.0973 \\
    .25   &0.0079    & 0.0909 \\
    .5    &0.0140    & 0.0969 \\
    .7    &0.0347    & 0.1267 \\
    1     &0.1256    & 0.1901 \\
    2     &0.2046    & 0.2738 \\
  \end{tabular}
  \label{tab:ks}
\end{table}
In this concrete example we see that a kernel size of approximately
$0.25$ gives the density ridge that is closest to the underlying
manifold. We also see that even in the low noise case, the mean squared
error of the estimates deteriorates quickly due to bias in the
estimations.

In the rest of the paper -- unless otherwise noted -- we have used a heuristic choice of the average distance to the 12th nearest neighbor as kernel size for the gaussian kernel~\citep{myhre2012mixture}.


\subsection{Manifold unwrapping}
\label{sub:manifold_unwrapping}
\begin{figure}[htbp]
\centering
\begin{subfigure}{.4\linewidth}
    \includegraphics[width=\linewidth]{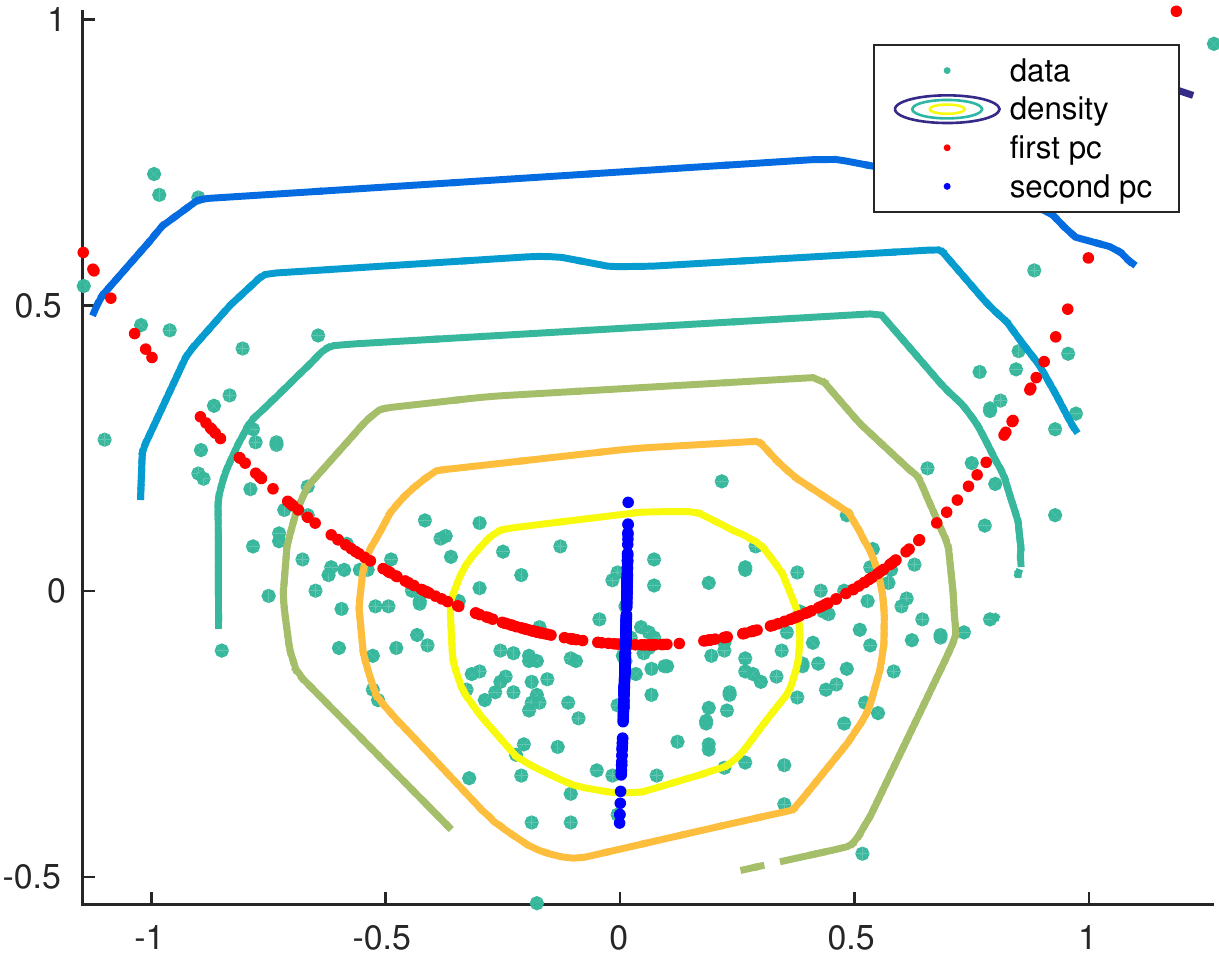}
  \caption{Nonlinear unimodal data set with density ridges shown in red
  and blue and the equiprobable contour lines of the kernel density
  estimate shown.
  }
  \label{fig:nonlin_unimodal}
  \end{subfigure}
  \qquad
  \begin{subfigure}{.4\linewidth}
  \includegraphics[width=\linewidth]{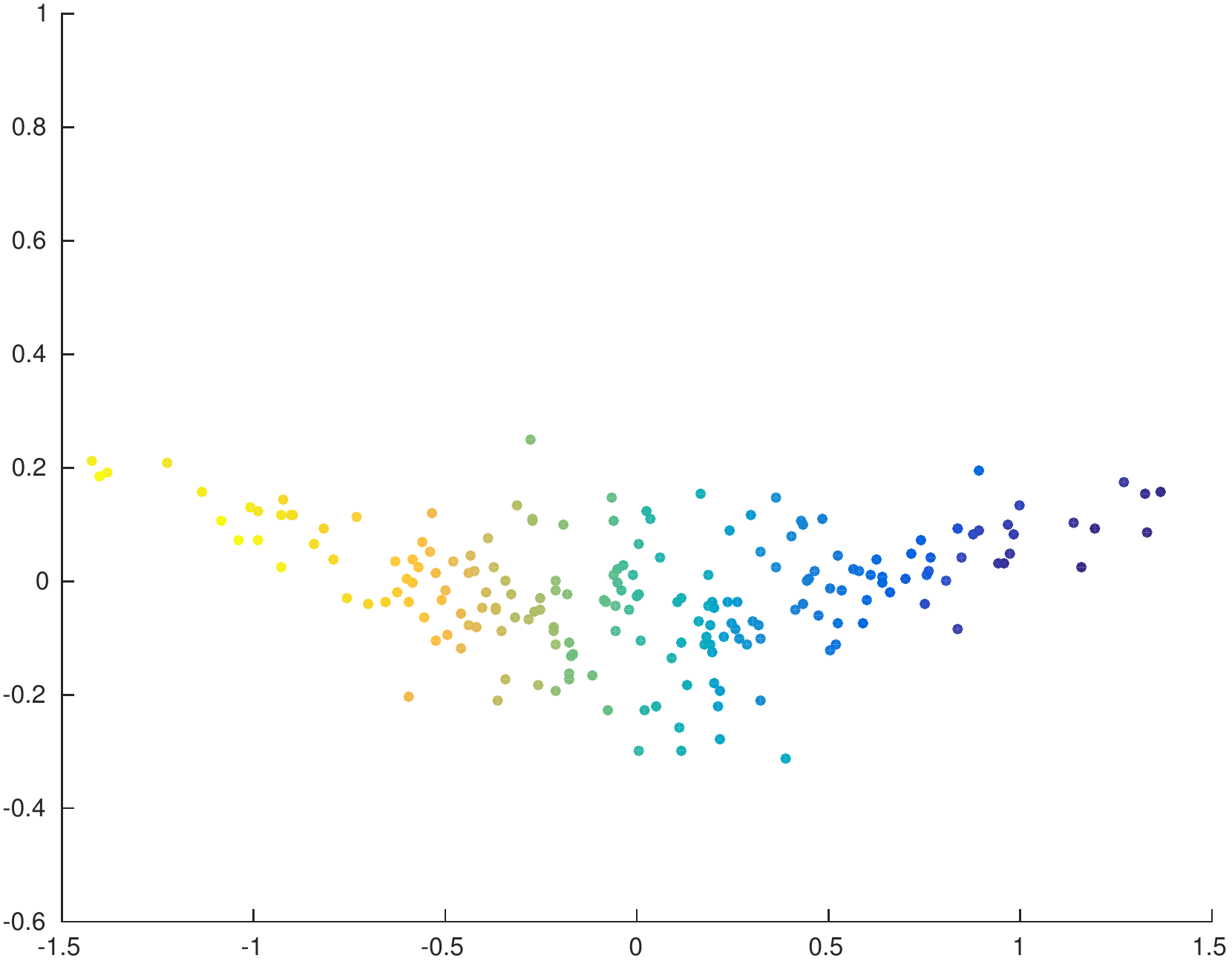}
\caption{Unwrapped version of the nonlinear distribution.}
\label{fig:one_moon_data}
  \end{subfigure}
  \caption{One crescent moon data set. Original and unwrapped version.}
  \label{fig:one_moon}
\end{figure}

In this section we show results on unwrapping one-dimensional manifolds
with noisy embeddings in higher dimensions -- $\mathbb R^d$, $d\geq 1$.
We start with the full unimodal example from
section~\ref{sec:principal_curve_unwrapping}.
In figure~\ref{fig:nonlin_unimodal} we see the data set, its two
orthogonal density ridges and the contours of the density.
The same data set can be seen in figure~\ref{fig:nonlin_unimodal}. After
applying the local density ridge unwrapping algorithm, we get the
results shown in figure~\ref{fig:one_moon_data}.
%
%
We see that by calculating distances along the underlying structure
we reveal the distribution along the intrinsic geometry. In this case a
gaussian kernel was used with a manually chosen kernel bandwidth of
$0.5$. If we inspect closer we see that there is a slight bias in
the first ridge, show in red, in figure~\ref{fig:nonlin_unimodal}. This
bias can also be seen in the unwrapped results, as they show a very
downscaled version of the original nonlinearity. This could be avoided
by either choosing a smaller kernel size, but risking more variance in
the ridge and thus less accurate unwrapped coordinates or by using a
varibable kernel density
estimator~\citep{wand1994kernel,bas2011extracting}.
%
%

\subsection{Density ridge translation}
\label{sub:density_ridge_tranlation}
\begin{figure}[htbp]
\begin{center}
  \includegraphics[width=2.5in]{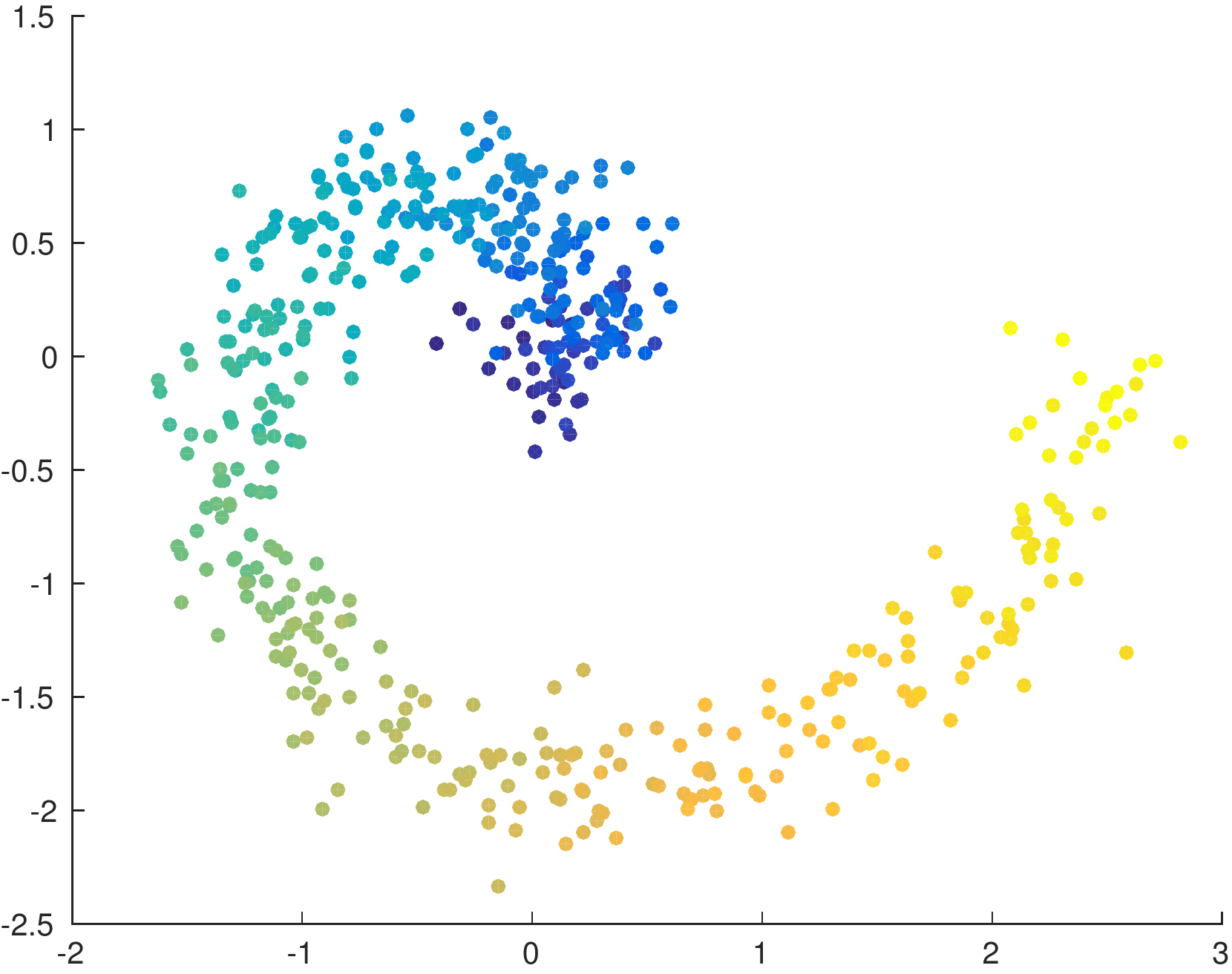}
\end{center}
\caption{Spiral shaped data set with gaussian noise. Color coding
indicates data point order in parameterization.}
\label{fig:spiral_data}
\end{figure}

\begin{figure}[htbp]
\centering
\begin{subfigure}{.4\linewidth}
  \includegraphics[width=\linewidth]{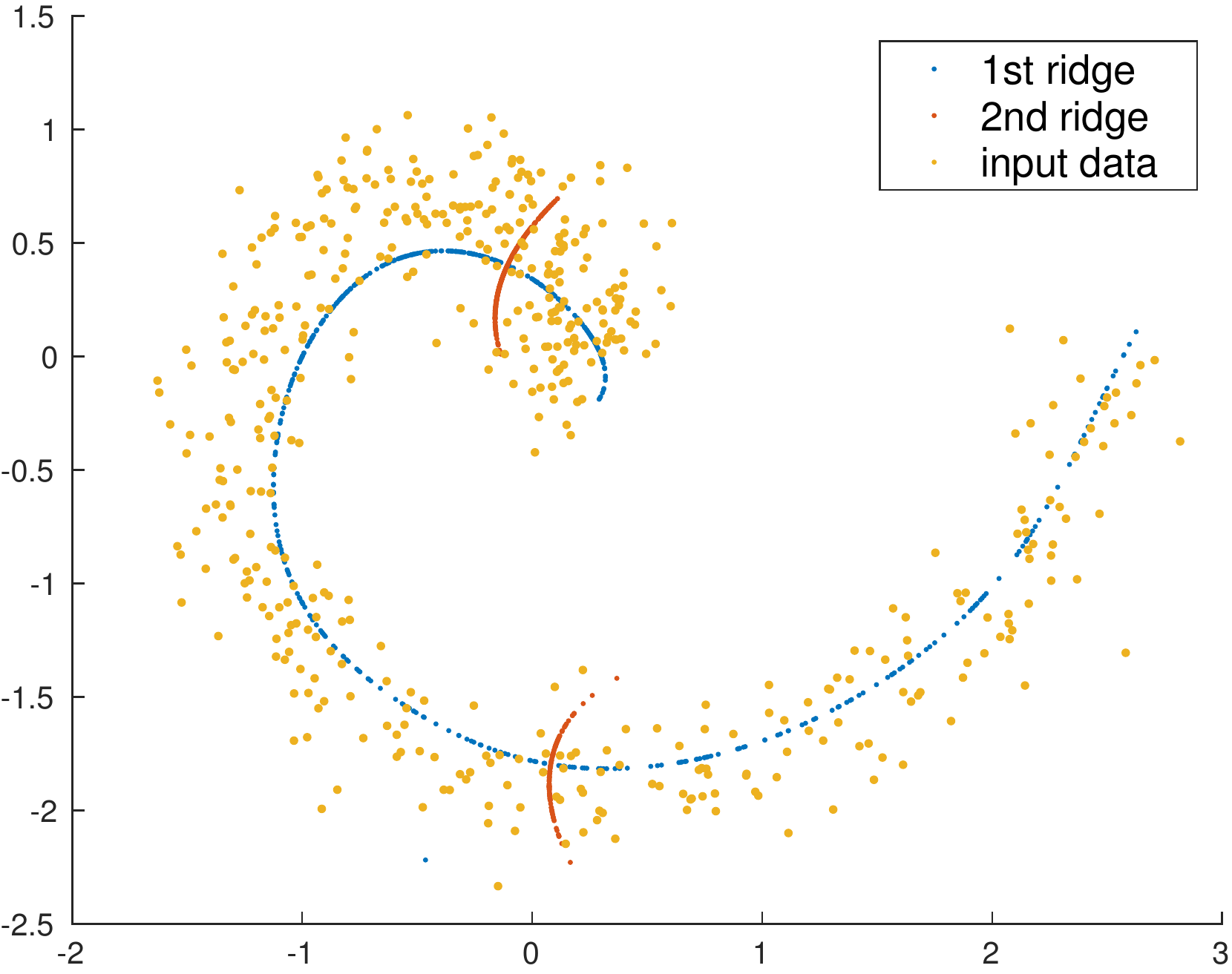}
\caption{Spiral shaped dataset with two orthogonal density ridges.}
\label{fig:spiral_data_with_curves}
\end{subfigure}
\qquad
\begin{subfigure}{.4\linewidth}
  \includegraphics[width=\linewidth]{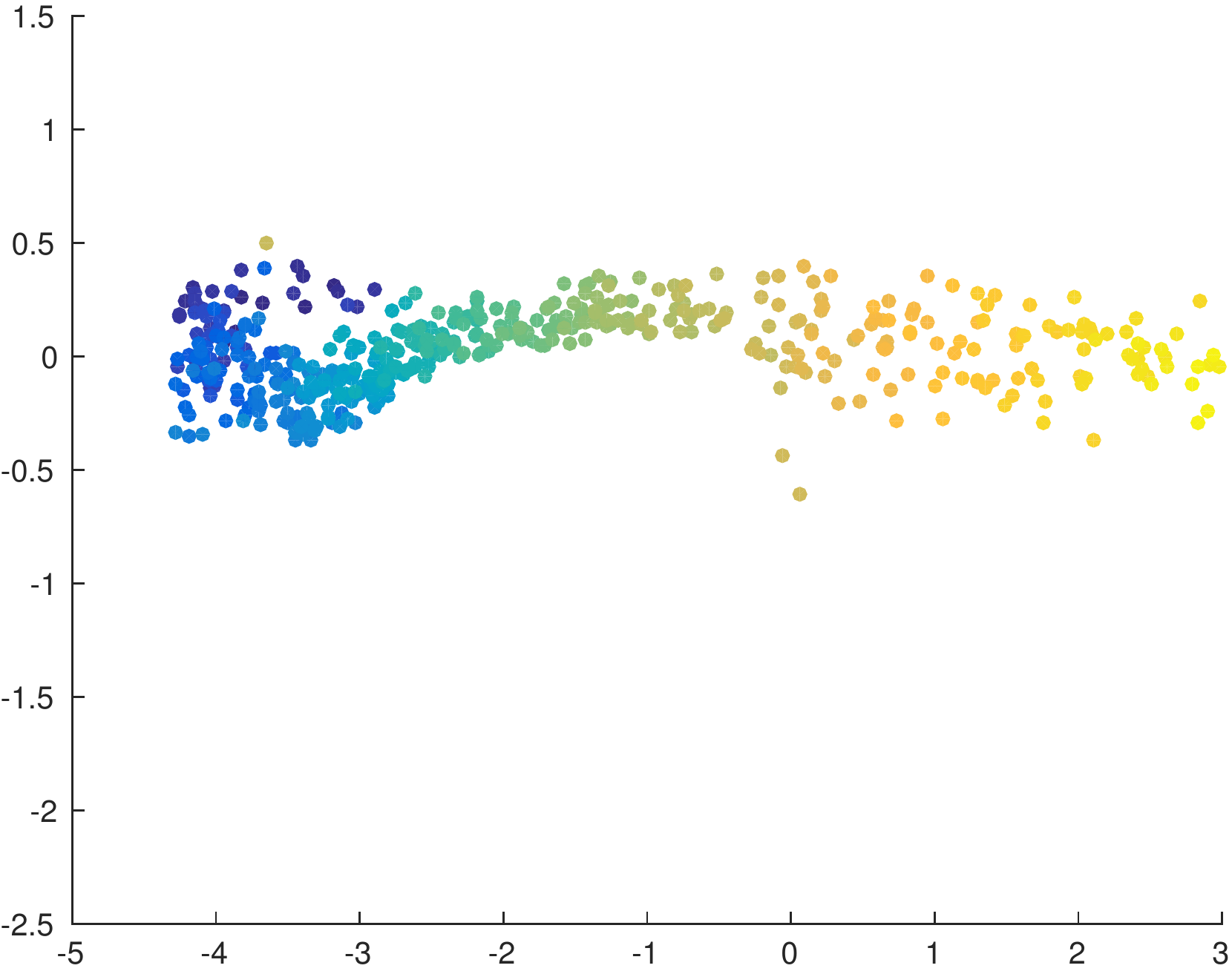}
\caption{Spiral shaped data set after the density ridge translation
algorithm.}
\label{fig:spiral_data_stitched}
\end{subfigure}
\caption{Spiral data set with density ridges and unwrapped version.}
\label{fig:spiral_data_stitched_and_unwrapped}
\end{figure}

To illustrate the density ridge translation algorithm we use a spiral shaped data set with
gaussian noise as shown in figure~\ref{fig:spiral_data}. The underlying manifold is
parameterized as $\bigl(\begin{smallmatrix}
  \mathbf x_1\\
  \mathbf x_2\\
\end{smallmatrix} \bigr) = \bigl(\begin{smallmatrix}
r\cos\theta\\
r\sin\theta
\end{smallmatrix} \bigr)$
, with $\theta = \left[ 0,2\pi \right]$ and $r=\left[ 0, 2 \right]$. The
$\theta$ parameter is indicated by the color
coding from dark blue to light yellow. The two local orthogonal density ridges are
shown in figure~\ref{fig:spiral_data_with_curves} and we note that there
are two separate local maxima. This
will give rise to two different local charts by algorithm~\ref{alg:ridgeTracing}. In
figure~\ref{fig:spiral_data_stitched} we see the results after applying
the density ridge translation algorithm on the charts. 
It is clear that the new unwrapped representation of the data follows the color coding directly, and the total horizontal variation from $-4$ to $3$ fits nicely
with a centered version -- $\left[-\pi, \pi\right]$ -- of the original parameterization from $0$ to $2\pi$. In the top left
of the unwrapped data structure we see some distortion with respect to
the parameterization/color coding of the original data. This is most
probably again due to bias in the density ridge estimation. The
nonlinearity of the innermost part of the spiral is most likely on a
smaller scale than the kernel bandwidth is able to capture, resulting in
a density ridge that is straighter than it should be and thus the projections will be less accurate.
\subsection{pca preprocessing and local density ridge unwrapping}
\label{sub:pca_pre}
In this section we present results on real data sets preprocessed by principal component analysis~\citep{jolliffe2002principal}.
\subsubsection{Subset of mnist handwritten digits}
\label{sub:n_dim_mnist}
In this experiment we use a subset of the mnist handwritten images. We
preprocess the data by projecting the data to the first three
principal components. We use images of
the number one -- $1$ -- in this experiment, since it is very susceptible
to geometric variations like rotations, translations and also more nonlinear deformations due to its simple shape. 

\begin{figure}[htpb]
\centering
\begin{subfigure}{.4\linewidth}
\centering
  \includegraphics[width=\linewidth]{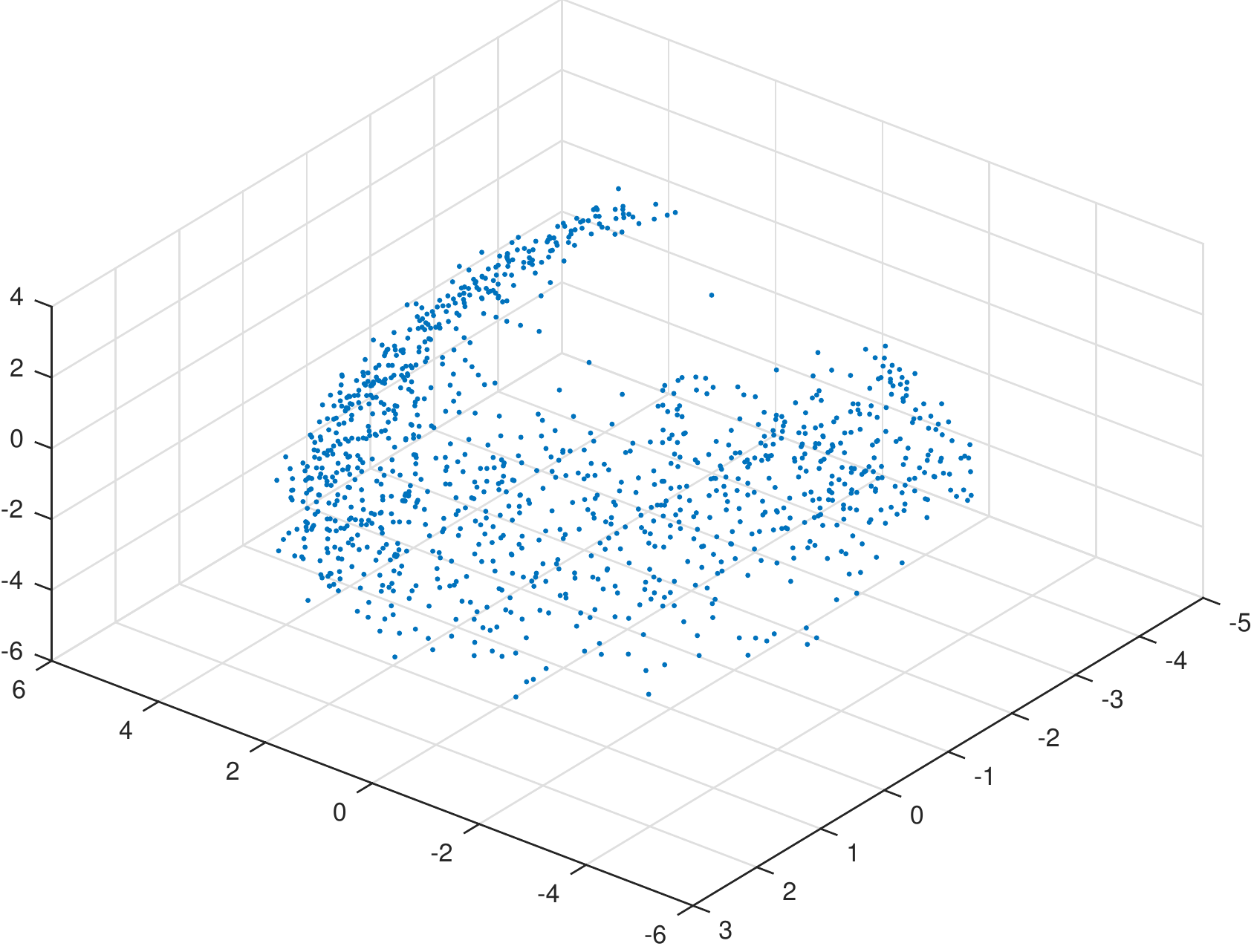}
\caption{The ones of the mnist data set projected to top three principal
curves.}
\label{fig:mnist_projected}
\end{subfigure}
\qquad
\begin{subfigure}{.4\linewidth}
\centering
  \includegraphics[width=\linewidth]{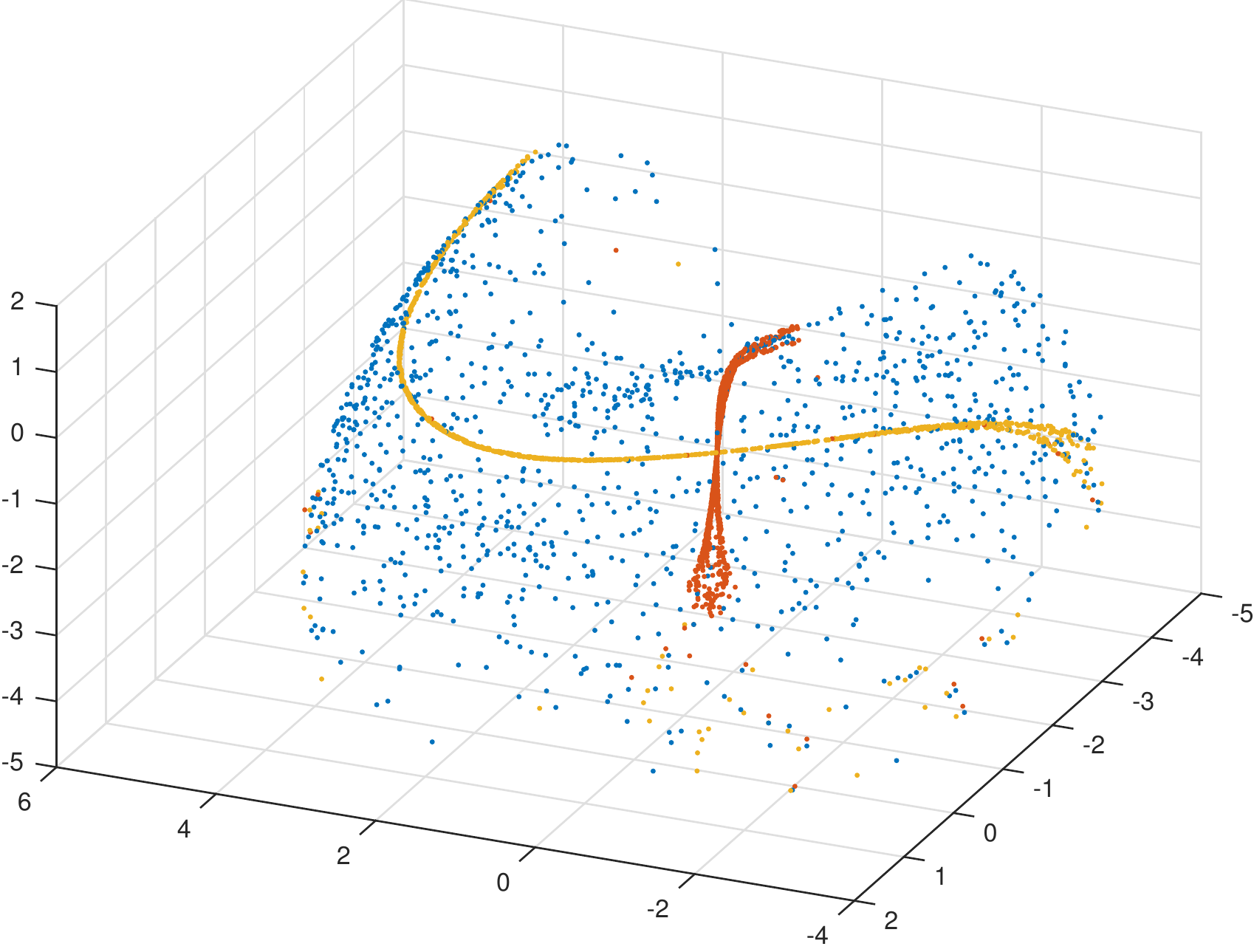}
\caption{Mnist handwritten digits image data projected to the top three principal components. The red and yellow curves shows the two one-dimensional density ridges of the projected dataset.}
\label{fig:mnist_ridge_curves}
\end{subfigure}
\caption{mnist-ones projected to three dimensions.}
\label{fig:mnist_proj}
\end{figure}

\begin{figure}[htbp]
\begin{center}
  \includegraphics[width=\linewidth]{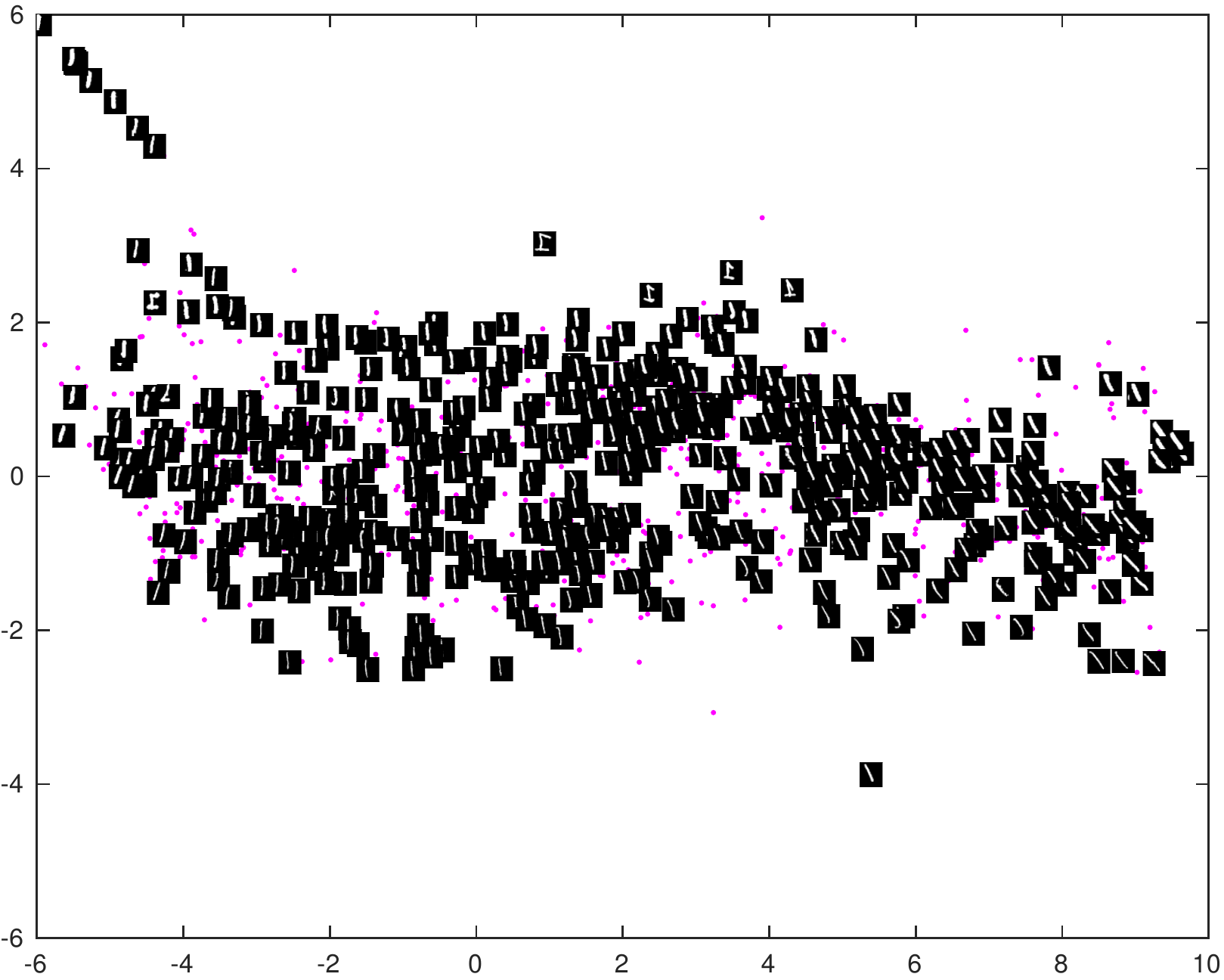}
\end{center}
\caption{mnist data set -- as shown in figure~\ref{fig:mnist_ridge_curves} -- unwrapped by using algorithm~\ref{alg:ridgeTracing}.
}
\label{fig:mnist_unwrapped}
\end{figure}

The data projected to the top three principal components
is shown in 
figure~\ref{fig:mnist_projected}.
We clearly see that the data follows a nonlinear submanifold and that the
variation of the images varies along a low dimensional surface. To test
our framework, we project the data to the two-dimensional density ridge,
and find the local modes by following the gradient.
If we use a large kernel bandwidth we introduce some bias in the density
ridge estimate seen e.g.\ in the splitting of the curves near the boundary of the manifold, but at the same time we get two global principal curves
that cover the entire projected data set, shown in figure~\ref{fig:mnist_ridge_curves} such that there is no need to translate charts together and we can directly get a global unwrapping.
In figure~\ref{fig:mnist_unwrapped} we see the results of applying the local density
ridge unwrapping algorithm (algorithm~\ref{alg:ridgeTracing}).


In figure~\ref{fig:mnist_unwrapped} we see the unwrapped mnist digits with a selection of random images of the actual digits on top.
We see that the coordinates represents a
clear structure in the data -- a straight up version of the digit around the origin and different, but symmetric variations along the two axes. The horizontal axis represents the orientation of the digits, while the vertical axis represesents thickness of the digits. in section~\ref{sec:comparison} we compare our results with other known manifold learning algorithms. 
  as a final remark in this setting we note that the use of pca preprocessing is a
\emph{linear} projection such that the geometric properties of the data
should not be influenced by the transformation. Ideally
pca would remove all ambient space except $\mathbb R^{d+1}$ such that the codimension\footnote{if $m \in \mathbb R^d$ is embedded in $\mathbb R^d$ the \emph{codimension} is $d-d$.} Is
one.

\subsubsection{Frey faces}
\begin{figure}[htpb]
    \centering
        \begin{subfigure}{.4\linewidth}
            \centering
            \includegraphics[width=\linewidth]{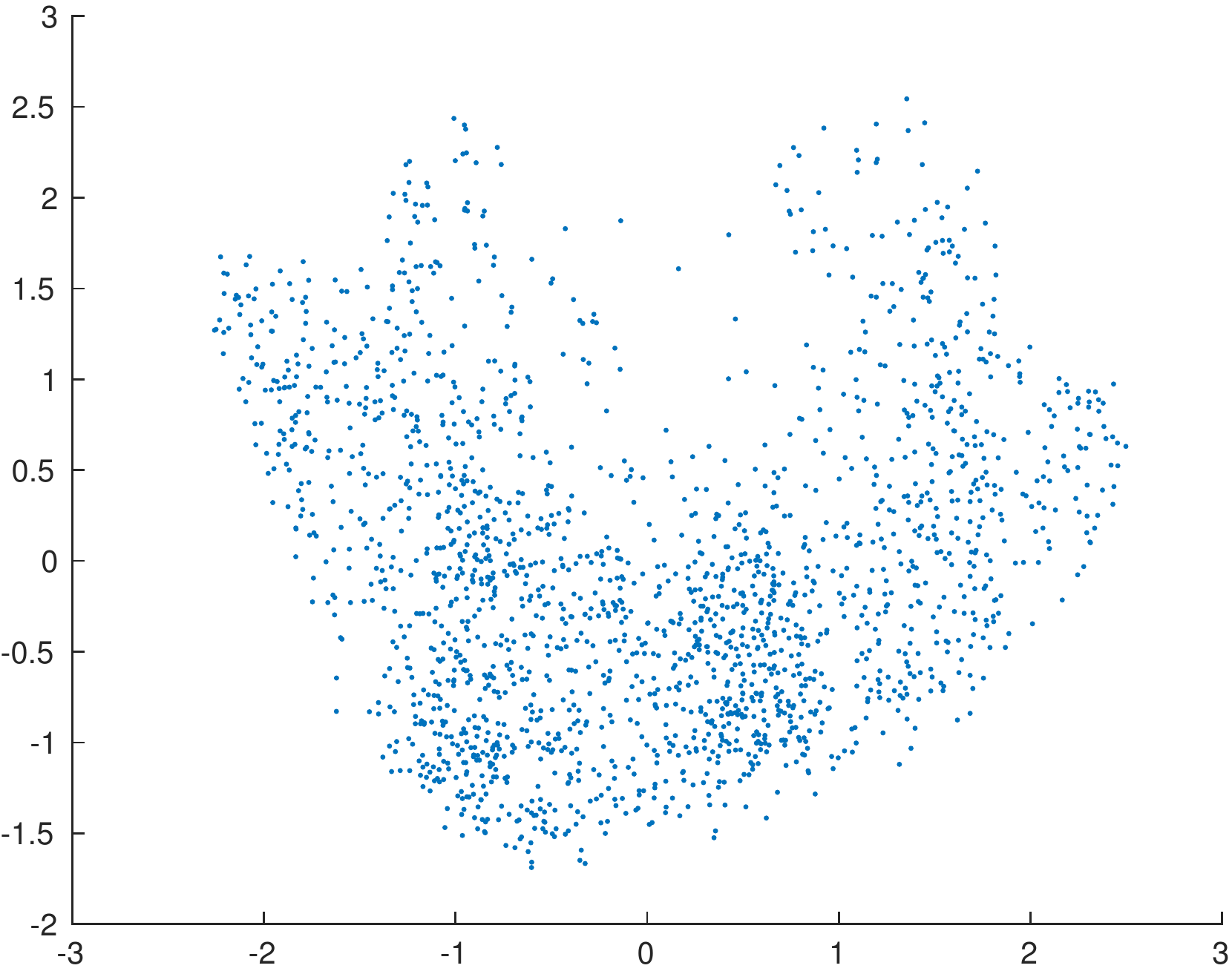} 
            \caption{Frey face images after reduction to two dimensions by principal
component analysis.}
            \label{subfig:faces_reduced}
        \end{subfigure}
        \qquad
        \begin{subfigure} {.4\linewidth}
            \centering
            \includegraphics[width=\linewidth]{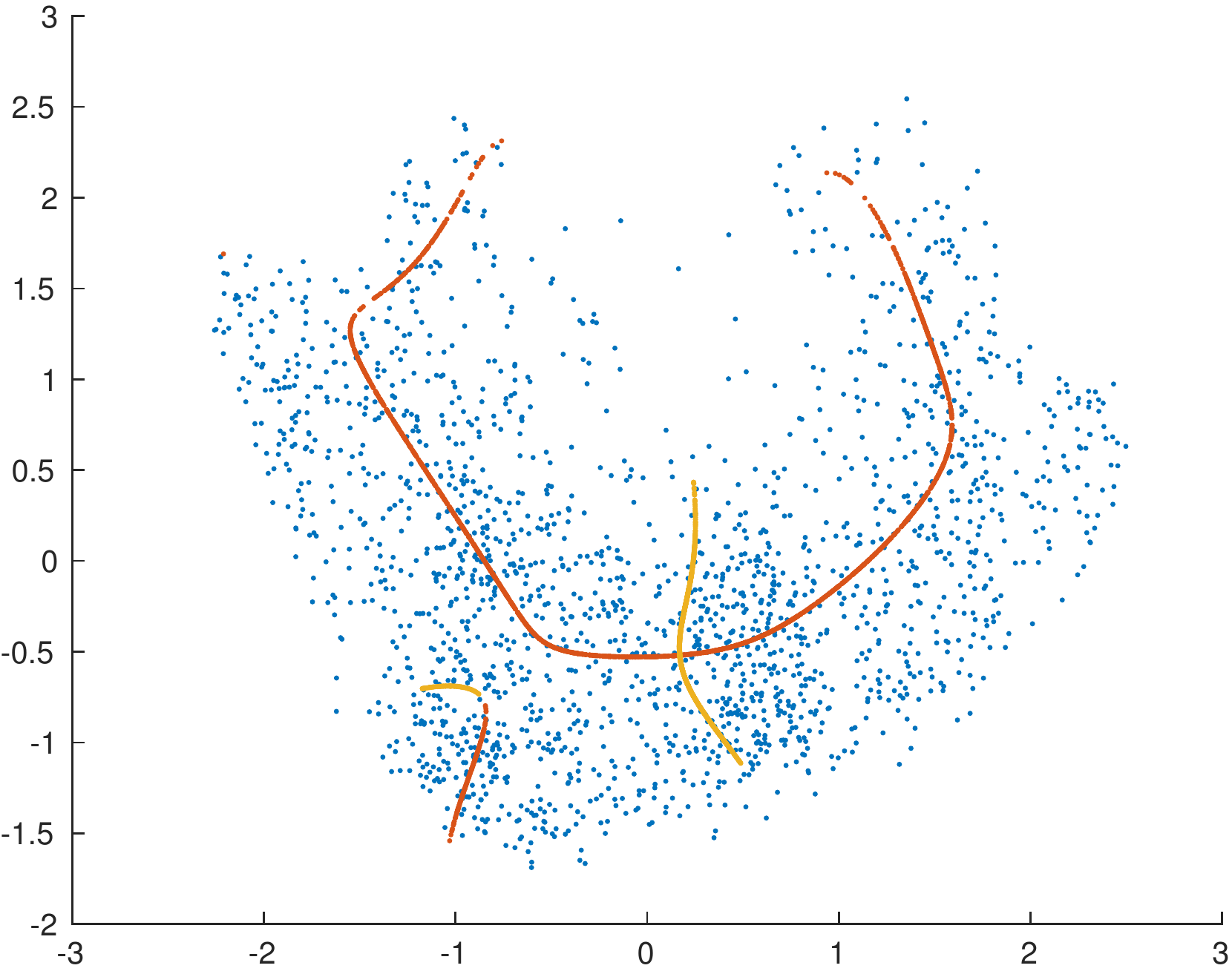}
            \caption{Frey faces image after pca with the first and second one-dimensional density
ridges as found by equation~\eqref{eq:ode_pc} shown in red and orange respectively.}
            \label{subfig:faces_and_curves}
        \end{subfigure}
    \caption{frey faces reduced to two dimensions by pca.}
    \label{fig:my_label}
\end{figure}

In this section we test the local density ridge unwrapping on the frey
faces data set. For visualization purposes we reduce the dimension of
the data set down to two by pca\@ -- shown in figure~\ref{subfig:faces_reduced}. We clearly see an underlying curved structure.
The curved structure has seemingly large orthogonal variance so we project to both of
the two available one-dimensional density ridges. The result can be seen
in figure~\ref{subfig:faces_and_curves}. The first density ridge/principal
curve, according to the ordering of the eigenvalues of the local
hessian, is shown in red. It captures well the underlying nonlinearity. The second ridge, shown in orange, fits well
with the orthogonal variation along the underlying nonlinearity.
%
We also have to note the disconnected ridges in the bottom left part of the data
set. 
This is probably due to a separate curvilinear structure in the data. We could alleviate this by choosing a larger kernel size to increase the smoothness of the ridge, but then there would be a risk of smoothing away the nonlinear curved structure of the data. We proceed with the chosen kernel size and note that this case cannot be described by a single one-dimensional manifold.

%
\begin{figure}[htbp]
\begin{center}
  \includegraphics[width=\columnwidth]{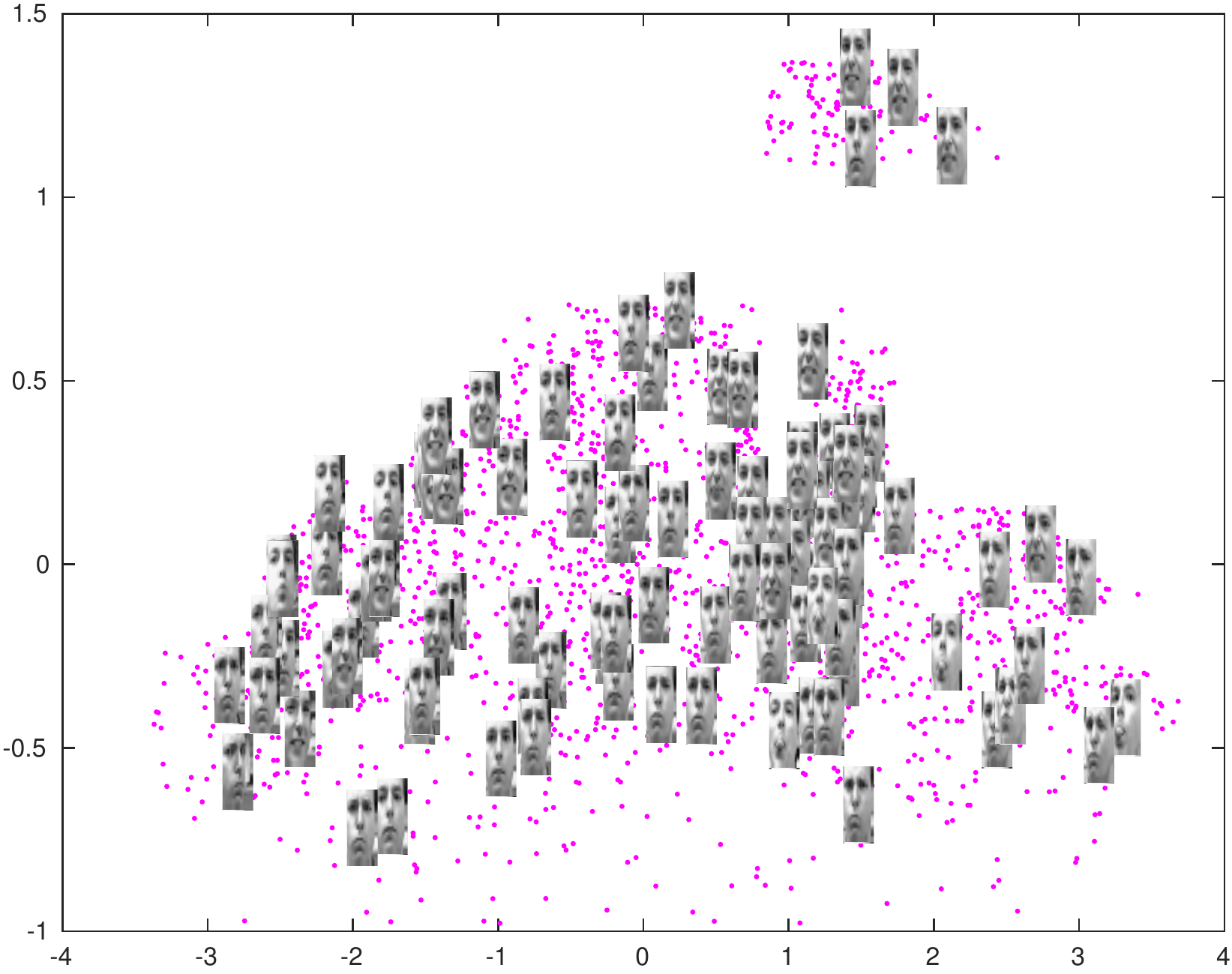}
\end{center}
\caption{Unwrapped version of the frey faces. A random selection of actual face images shown on top of the unwrapped coordinates for illustration of structure.
}
\label{fig:frey_curvilinear}
\end{figure}
\begin{figure}[htbp]
  \begin{center}
    \includegraphics[width=2.5in]{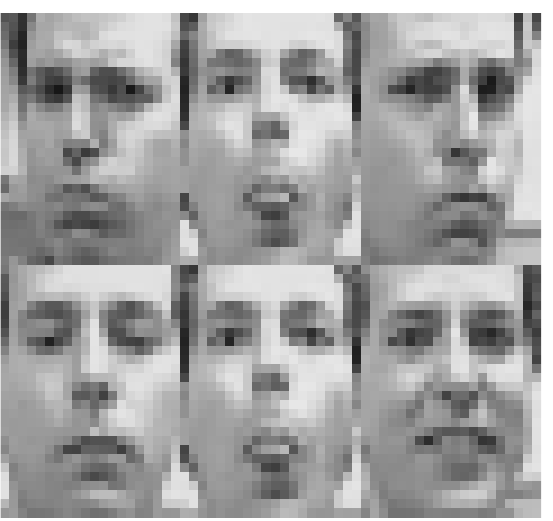}
  \end{center}
  \caption{Top row: largest, orgin and smallest first unwrapped coordinate.
  Bottom: largest, origin and smallest second unwrapped coordinate.}
  \label{fig:faces_montage}
\end{figure}

In figure~\ref{fig:frey_curvilinear} we see the results of the local
density ridge unwrapping algorithm.
A random selection of 100 of the original
input images are displayed on top of the unwrapped coordinates. We see
that the main horizontal direction, exhibiting the largest variance, captures the
left-right orientation of the faces. In figure~\ref{fig:faces_montage} each row shows the
largest negative, origin and largest positive coordinate of each of the two
coordinates of the unwrapping respectively. We again see that the first ridge - top row -
corresponds to the left-right orientation of the faces. The
second ridge is harder to interpret, but we note that the orientation of the face is looking straight ahead
in all bottom three images in figure~\ref{fig:faces_montage}, which fits well with the second coordinate being orthogonal to the
horizontal direction represented by the first coordinate.

The disconnected ridges results in the
region in the top right corner. The analysis and discussion
of these cases are left for future research.



\subsection{$d$-dimensional ridges}
\label{sub:n_dim_ridges_ex}
In this section we present some applications of our non-parametric
$d$-dimensional framework. Inspired by doll{\'a}r et
al.\, \citep{dollar2007non}, we can perform several different operations once each point that
lies on the manifold has been equipped with a tangent space - spanned by
the hessian eigenvectors. Most important in this setting are calculating geodesics as in equation~\eqref{eq:geod_opt}, and out-of-sample projections that are faster than projections by solving differential equations. 

We include two examples: the first example shows out of sample projections by projecting so point to the closest local tangent space. The second example shows smooth geodesics found by the alternating least
squares algorithm presented in
section~\ref{sub:approximating_geodesics}.

\begin{figure}[htbp]
\begin{center}
  \includegraphics[width=2.5in]{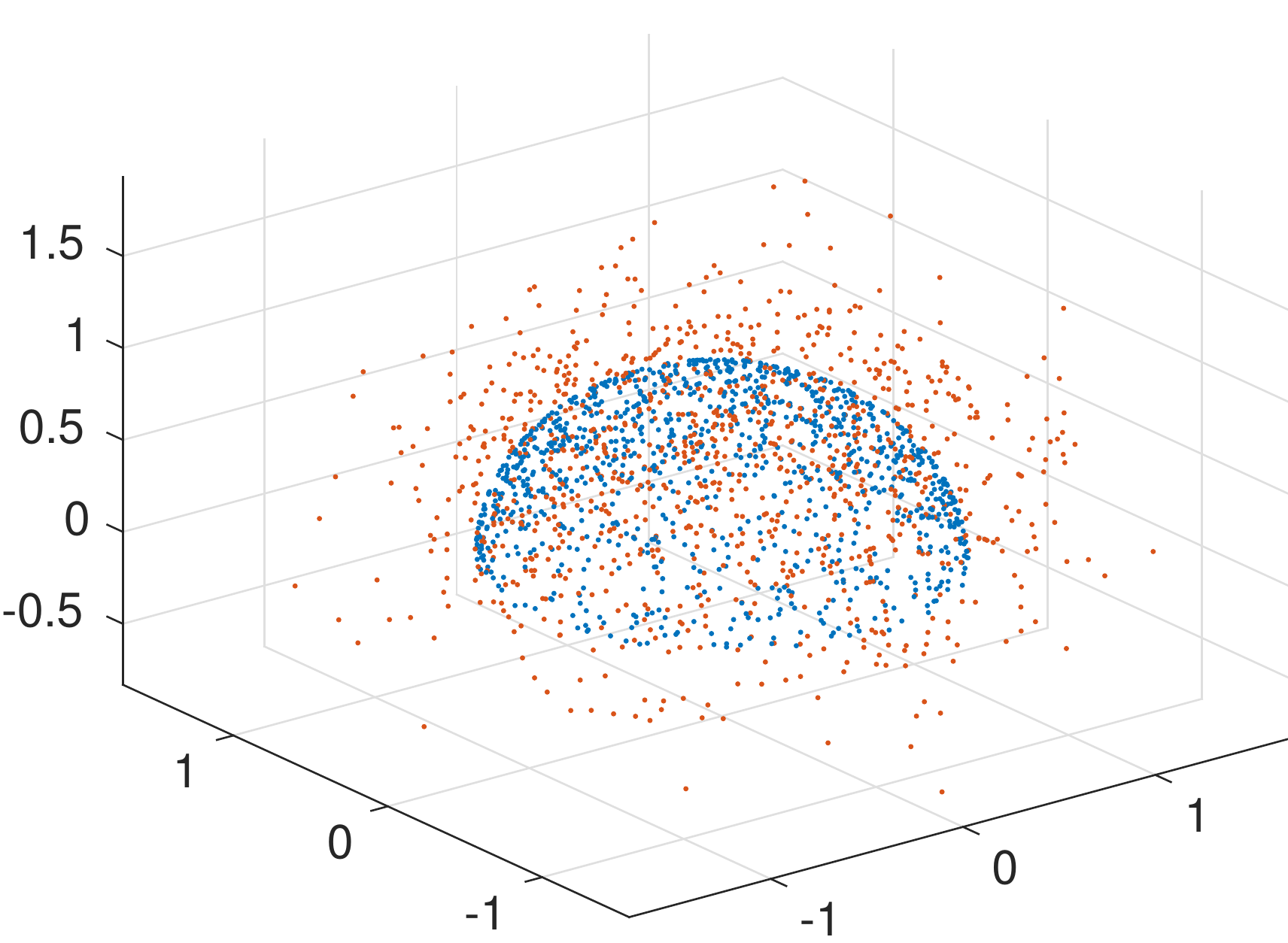}
\end{center}
\caption{Data sampled from a hemisphere with(red) and without noise(blue). The
noise level is $n(0,.3i)$.}
\label{fig:out_of_sample}
\end{figure}

\begin{figure}[htbp]
\begin{center}
  \includegraphics[width=2.5in]{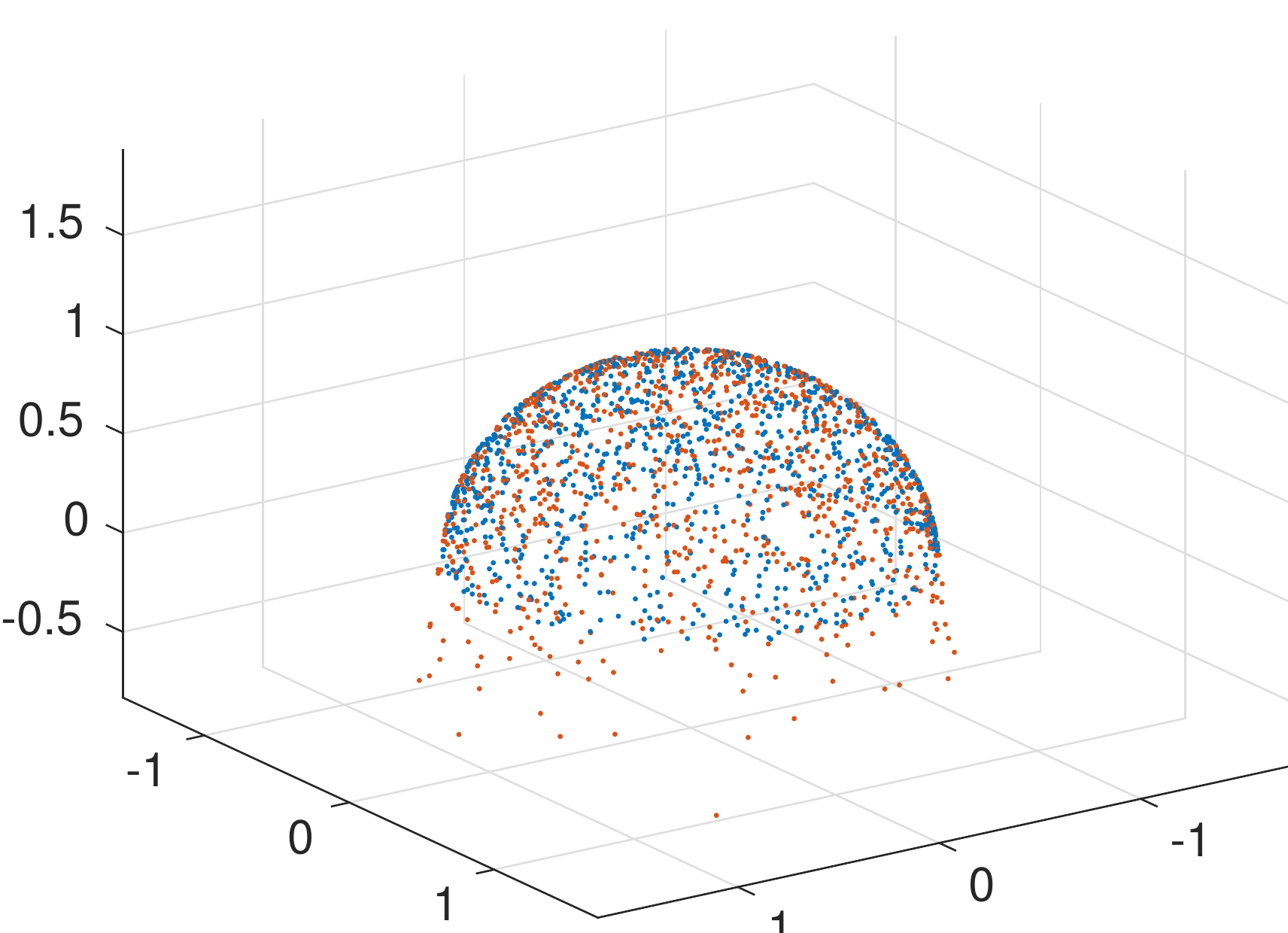}
\end{center}
\caption{The hemisphere data projected to the local tangent space using
the method from dollar}
\label{fig:out_of_sample_projected}
\end{figure}

In figure~\ref{fig:out_of_sample} we see a synthetic hemisphere sampled uniformly with $n(0,.3i)$ noise added. 
In figure~\ref{fig:out_of_sample_projected} we see the points projected to the closest tangent space of the density ridge by equation \ref{eq:out_of_sample}.
It is obvious that the out of sample projections are working as expected, but we also see that when samples are so noisy that they cannot be orthogonally projected to the manifold - in this case they are not directly above the hemisphere - they are projected to the tangent space of the boundary of the manifold, and the structure is thus lost. In this case the flat tangent spaces at the boundary gives the appearance of projecting to a cylinder below the hemisphere. Suggestions and solutions to this problem is left to future research.
\subsection{Isometric unwrapping}
\label{sub:swissroll_ex}

\begin{figure}[htbp]
\begin{center}
  \includegraphics[height=1.5in,width=3in]{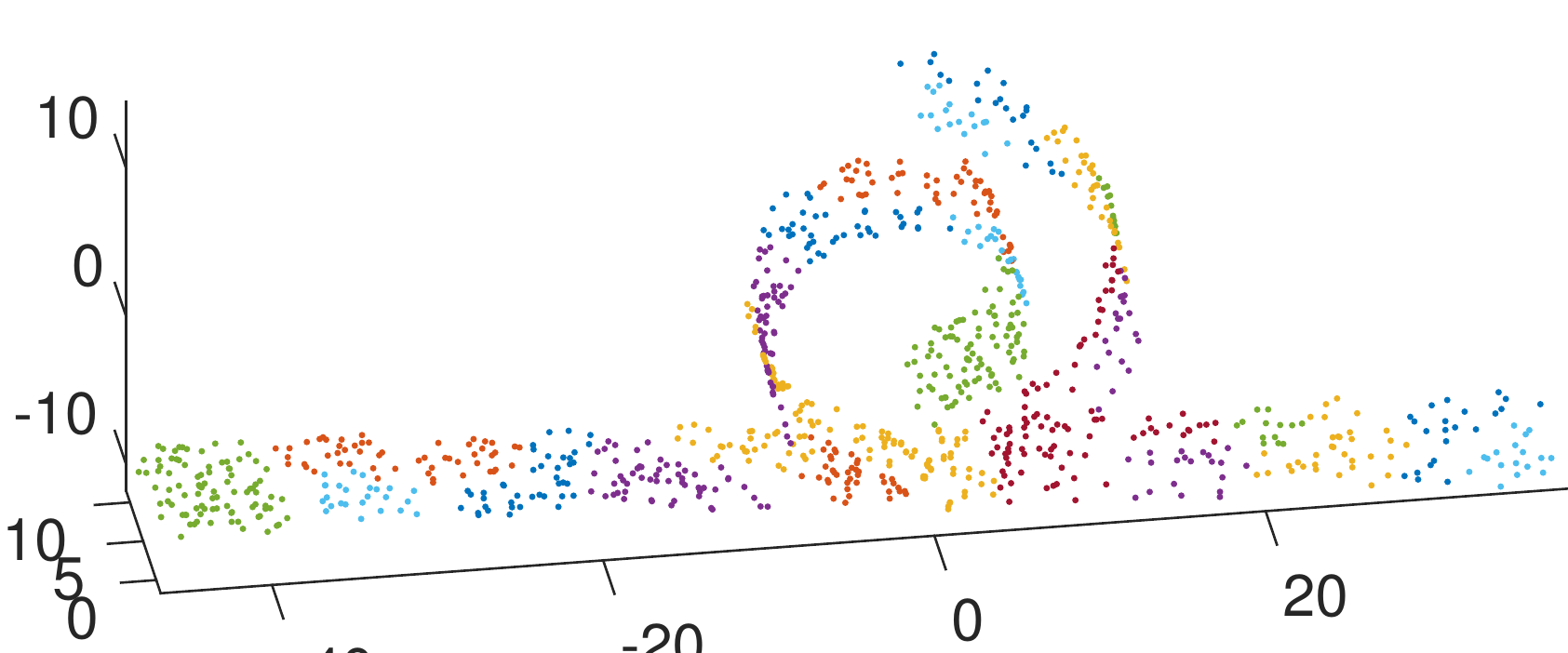}
\end{center}
\caption{Swissroll dataset with unwrapped coordinates shown in the
tangent space of the reference mode.}
\label{fig:swiss_unrolled}
\end{figure}

In the case where the intrinsic manifold has zero gaussian curvature it can be unfolded
isometrically by the algorithms from section~\ref{sec:n_dim_ridges}.
We start with the `swissroll' example, a two dimensional surface of zero
gaussian curvature distorted by a nonlinear function and $n(0,
\sigma^2)$-noise. This is a popular dataset/example to illustrate
manifold learning algorithms. It is useful because it represents a
strong non-linear structure, but at the same time it is a simple surface
without gaussian curvature such that it can in fact be represented by a
single coordinate chart. After projecting to the ridge and running the algorithms, we get the result of unwrapping shown in
figure~\ref{fig:swiss_unrolled}.
The color coding illustrates the local basins of attraction. In the
unwrapped version - shown in the tangent space of one of the modes - we
clearly see that the ordering of the local charts is preserved.

\subsection{mnist autoencoder}
We end this section with an example of a real data set where multiple $d$-dimensional charts emerge such that the isometric unwrapping algorithm, algorithms \ref{alg:dDimParallelTransport} and \ref{alg:isometricUnfolding}, has to be applied. Again we use the mnist one digits, but this time we reduce the dimension to three by using a deep autoencoder - as given in~\citep{van2009dimensionality}.

\begin{figure}[htbp]
\begin{center}
  \includegraphics[width=.5\linewidth]{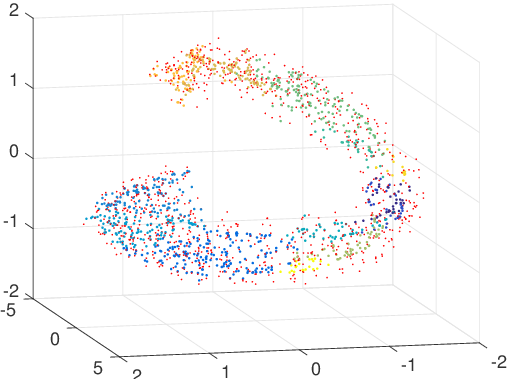}
\end{center}
\caption{Mnist one digits after reducing dimension to
three by using an autoencoder (small red dots) and the two-dimensional density ridge. The color coding of the ridge indicates the local charts/attraction basins found.}
\label{fig:mnist_autoenc}
\end{figure}

In figure~\ref{fig:mnist_autoenc} we see both the data set and the data projected to the two-dimensional density ridge. The underlying ridge fits the data well, and we see that there are several local charts involved, indicated by the color coding. 

\begin{figure}[htbp]
\begin{center}
  \includegraphics[width=\columnwidth]{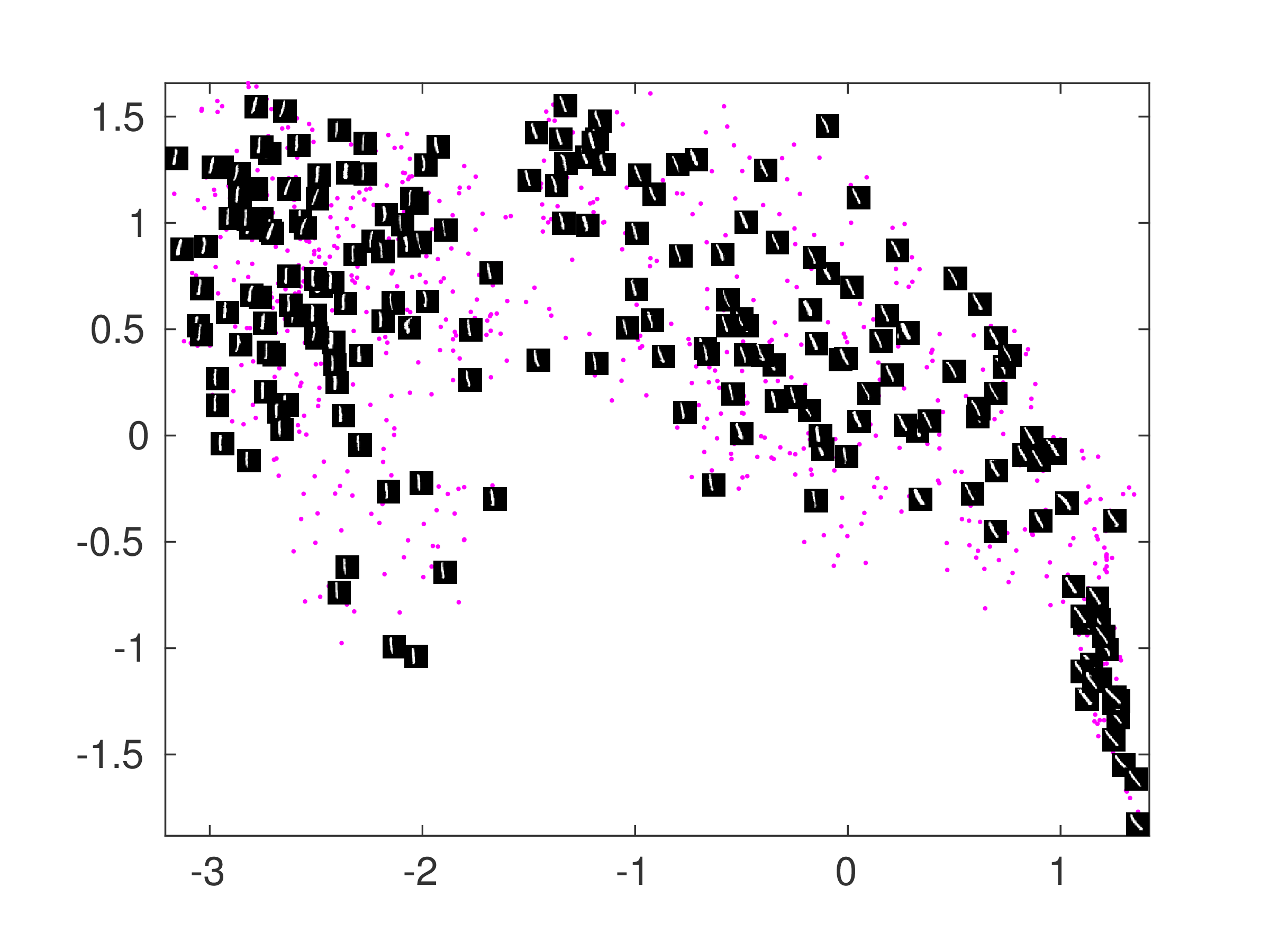}
\end{center}
\caption{Isometric unwrapping on mnist ones after reducing dimension to
three by using an autoencoder.}
\label{fig:mnist_autoenc_unwrapped}
\end{figure}

After running the isometric density ridge unwrapping scheme we get the results shown with a random selection of digits overlaid in figure~\ref{fig:mnist_autoenc_unwrapped}. We again se a clear structure in the data with the ones rotating/changing orientation along the horizontal axis. The structure in the other mnist-ones example, where the ones changed from thicker to thinner along the vertical axis is not as strong in this example. 
\subsection{Comparison with other unwrapping algorithms}
\label{sec:comparison}
Finally we compare our algorithm with isomap~\citep{tenenbaum2000global}, \emph{local linear embedding}~\citep{roweis2000nonlinear}, \emph{local tangent space alignment}~\citep{zhang2004principal} and \emph{maximum variance unfolding}~\citep{weinberger2006introduction}. We use a synthetic data set consisting of a two-dimensional surface embedded in $\mathbb R^3$ with additive/convolutional gaussian noise in addition to the mnist-ones used previously.

\begin{figure}[htbp]
    \begin{center}
        \begin{subfigure}{.4\textwidth}
            \centering
            \includegraphics[width=\columnwidth]{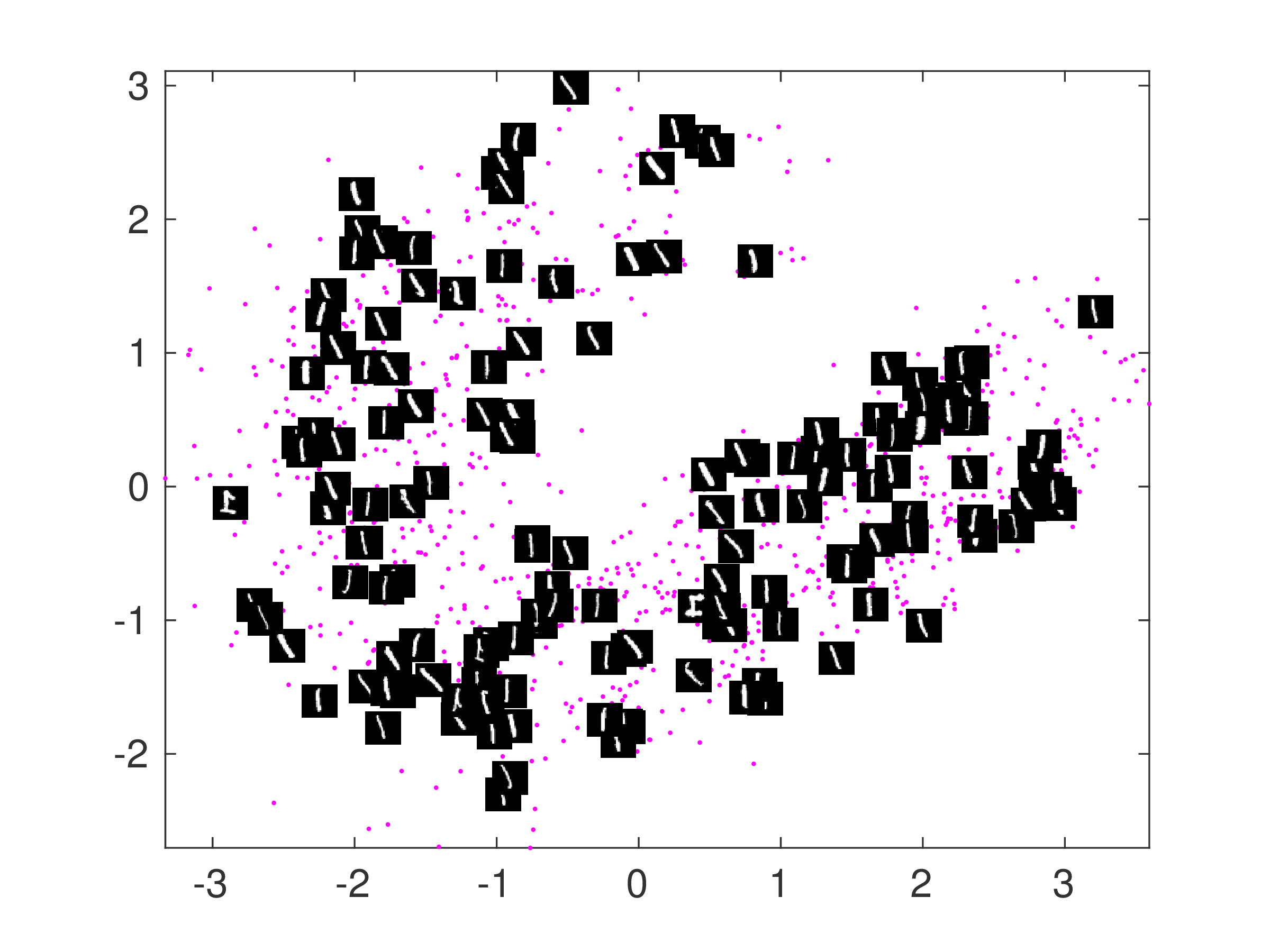} 
            \caption{Maximum variance unfolding. $k=12$}
            \label{fig:mvu_mnist}
        \end{subfigure}
        \begin{subfigure} {.4\textwidth}
            \centering
            \includegraphics[width=\columnwidth]{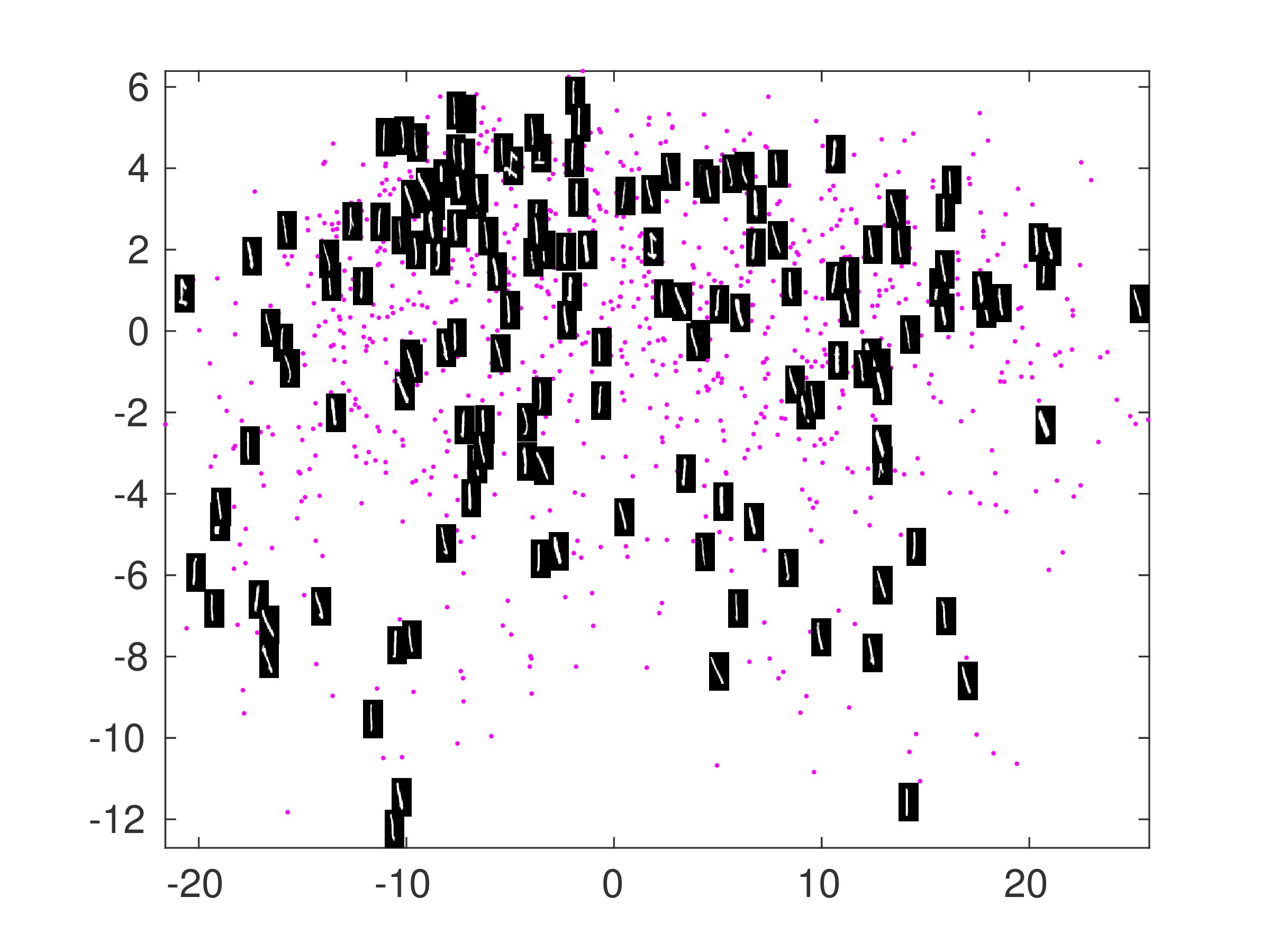}
            \caption{Isomap. $k=20$}
            \label{fig:isomap_mnist}
        \end{subfigure}
        \begin{subfigure}{.4\textwidth}
            \centering
            \includegraphics[width=\columnwidth]{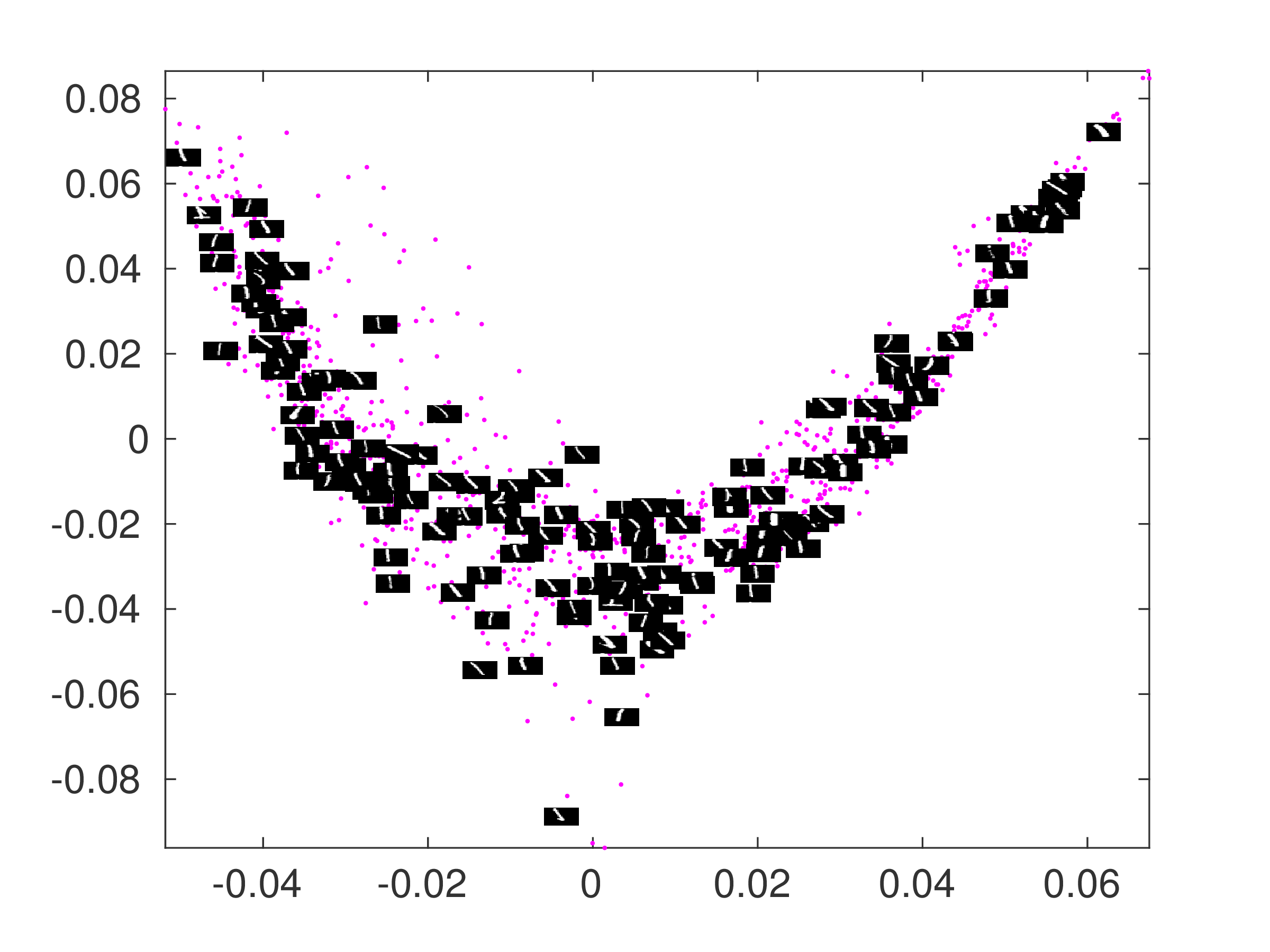}
            \caption{Local linear embedding. $k=20$.}
            \label{fig:lle_mnist}
        \end{subfigure}
        \begin{subfigure}{.4\textwidth}
            \centering
            \includegraphics[width=\columnwidth]{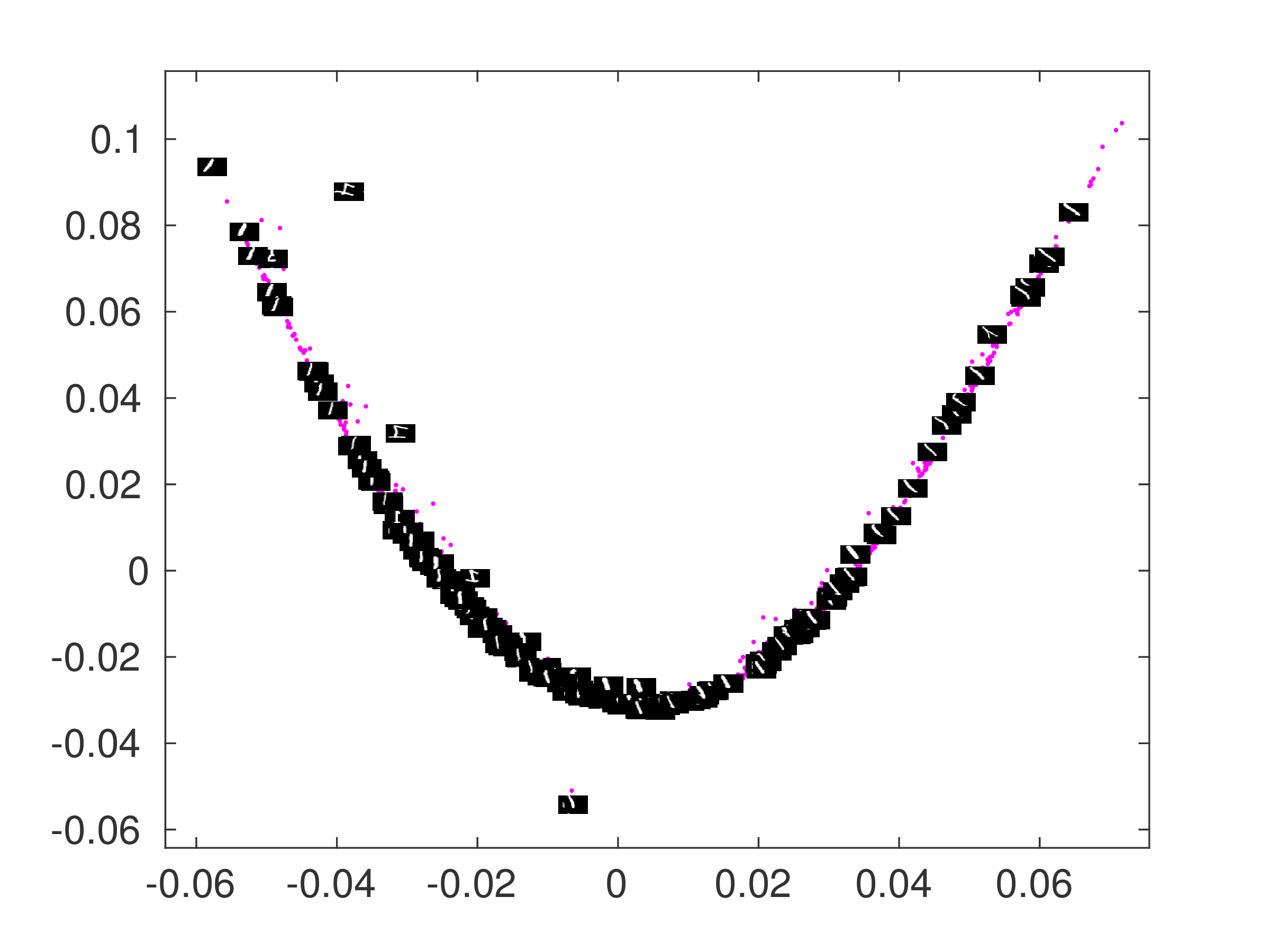}
            \caption{Local tangent space alignment. $k=150$}
            \label{fig:ltsa_mnist}
        \end{subfigure}
    \end{center}
    \caption{mnist-ones dataset tested on several benchmark manifold learning algorithms.}
    \label{fig:mnist_comp}
\end{figure}

\begin{figure}[htpb]
    \centering
    \includegraphics[width=\linewidth]{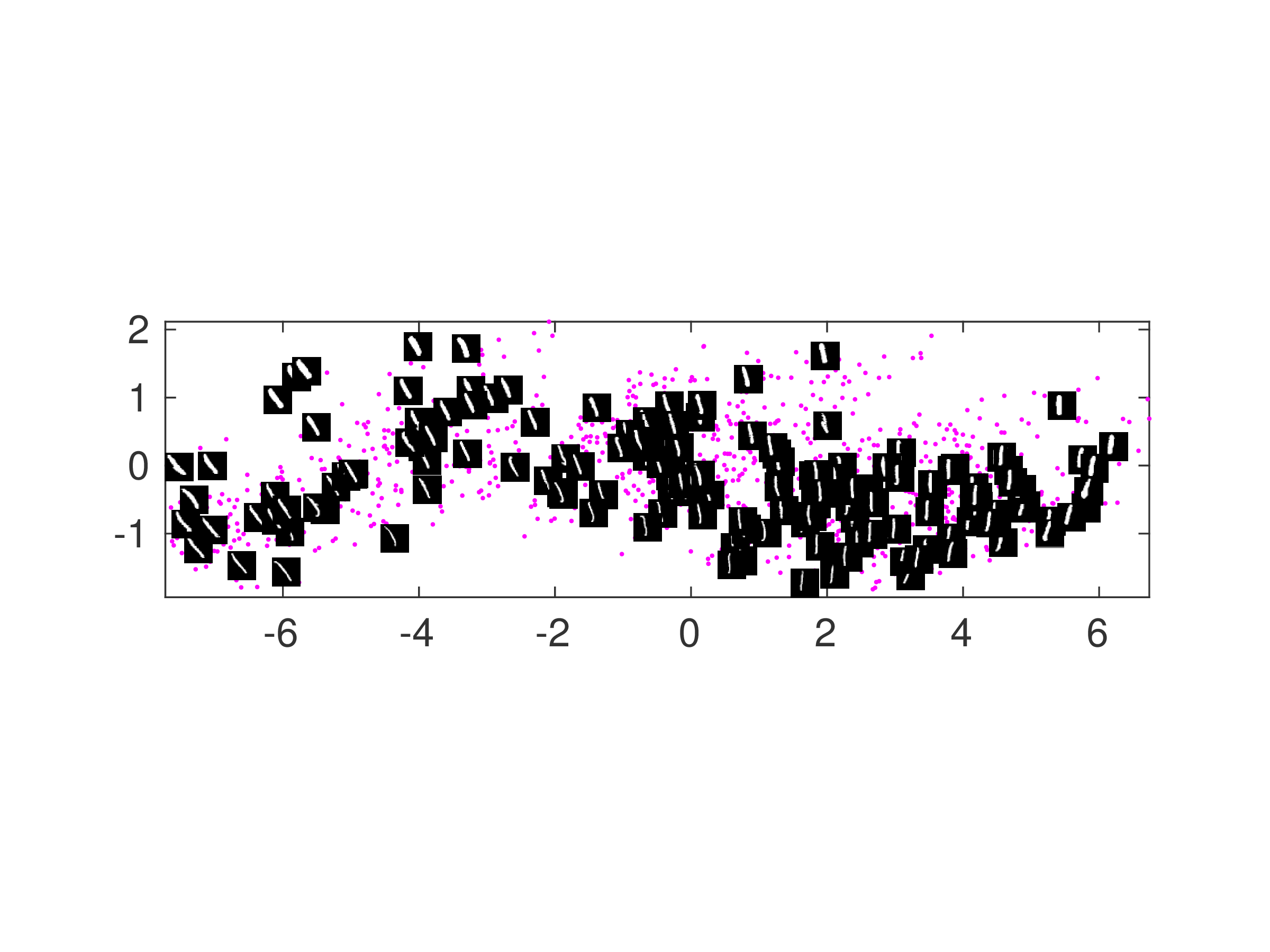}
    \caption{Density ridge unwrapping on mnist-ones.}
    \label{fig:mnist_ones_dru}
\end{figure}

For the mnist-ones, we use our isometric unwrapping with a slightly smaller kernel size than the previous example - recall that the previous example a large kernel size was used to obtain one-dimensional ridges that cover the entire manifold. This leads to several local modes, and thus several charts over the manifold such that algorithms \ref{alg:dDimParallelTransport} and \ref{alg:isometricUnfolding} have to be used. The result is shown in figure~\ref{fig:mnist_ones_dru}. The other algorithms - isomap, lle, ltsa and mvu - all have a $k$-nearest-neighbor parameter that has to be set so we run them with different values of $k=\begin{bmatrix}5, 12, 20, 50, 100\end{bmatrix}$ and select the most visually meaningful results, shown in figure~\ref{fig:mnist_comp}. We note that the ltsa had problems with finding the eigenvalue/eigenvectors such that the $k$ values had to be set much higher - $k=\begin{bmatrix}150, 300\end{bmatrix}$.
Local tangent space alignment, figure~\ref{fig:ltsa_mnist} shows a similar structure to the density ridge unwrapped result that follows the orientation of the digit, but the structure is compressed to a curve and the structure orthogonal to the curve is lost. There is also no apparent unwrapping, the manifold is still curved. The same goes for the local linear embedding result, figure~\ref{fig:lle_mnist}, no apparent unwrapping. Also, looking at the images overlaid over the data, there does not seem to be the same smooth consistent structure as in the density ridge unwrapped version. Isomap, figure~\ref{fig:isomap_mnist} has unwrapped the manifold, but the structures seen in the overlaid images does not seem to follow specific patterns. Finally, maximum variance unfolding in figure~\ref{fig:mvu_mnist} has also failed in unwrapping the noisy cylinder shape, we also note that the orientation of the numbers is not preserved properly.

\begin{figure}[htpb]
    \centering
    \includegraphics[width=.5\linewidth]{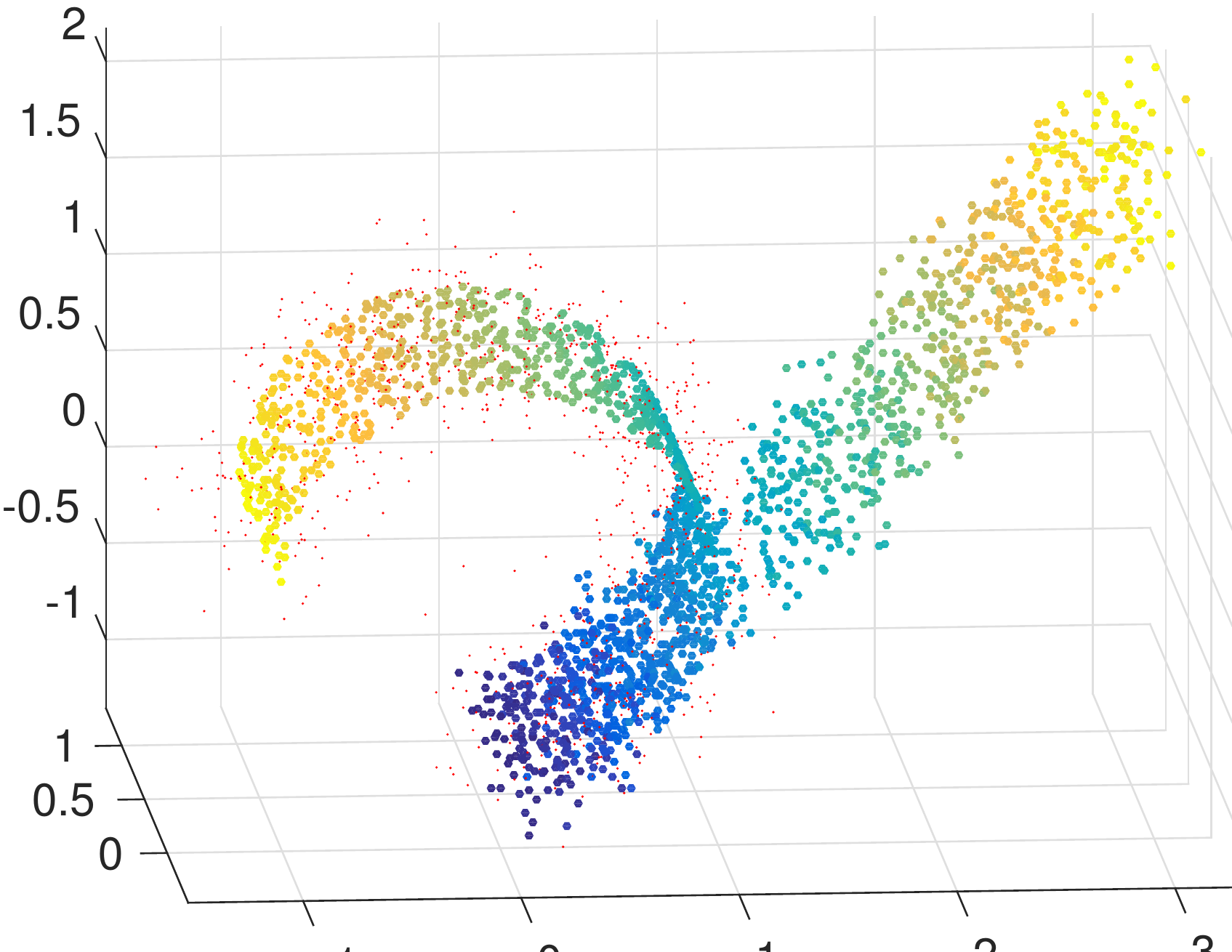}
    \caption{Dataset with(red dots) and without(colored dots) noise. The result of our algorithm projected to the tangent space of a reference mode shown in similar colors as the data without noise.}
    \label{fig:comp_semiroll}
\end{figure}

\begin{figure}[htpb]
    \centering
    \begin{subfigure}{.4\textwidth}
           \centering
            \includegraphics[width=\columnwidth]{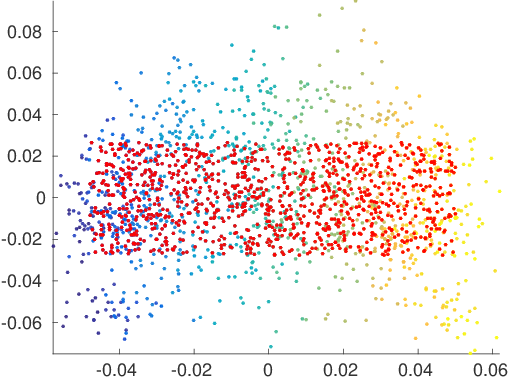} 
            \caption{Isomap $k=12$.}
            \label{fig:isomap_semiroll}
    \end{subfigure}
    \begin{subfigure}{.4\textwidth}
           \centering
            \includegraphics[width=\columnwidth]{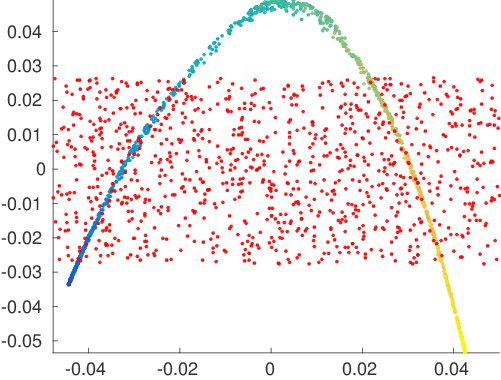} 
            \caption{Laplacian eigenmaps $k=12$.}
            \label{fig:lapeig_semiroll}
    \end{subfigure}
        \begin{subfigure}{.4\textwidth}
           \centering
            \includegraphics[width=\columnwidth]{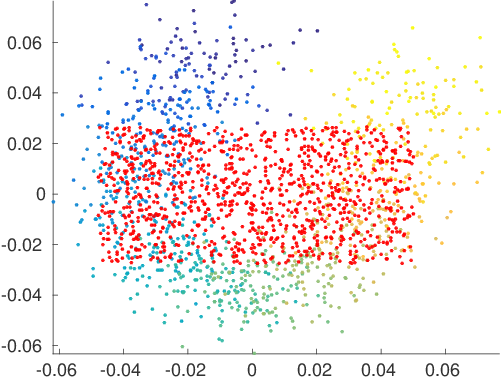} 
            \caption{Local linear embedding $k=12$.}
            \label{fig:lle_semiroll}
    \end{subfigure}
    \begin{subfigure}{.4\textwidth}
           \centering
            \includegraphics[width=\columnwidth]{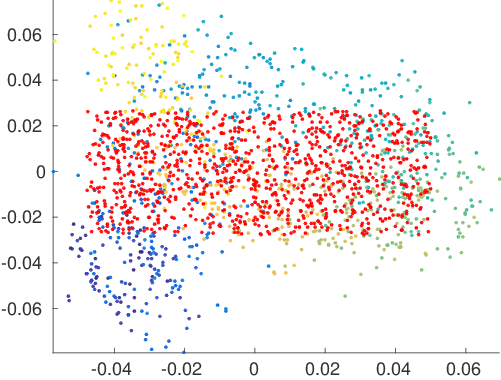} 
            \caption{Maximum variance unfolding $k=12$.}
            \label{fig:mvu_semiroll}
    \end{subfigure}
    \begin{subfigure}{.4\textwidth}
           \centering
            \includegraphics[width=\columnwidth]{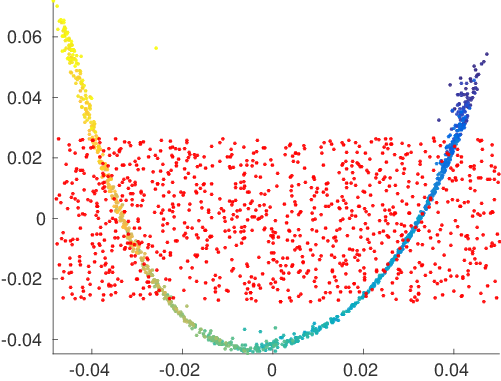} 
            \caption{Local tangent space alignment $k=12$.}
            \label{fig:ltsa_semiroll}
    \end{subfigure}
    \begin{subfigure}{.4\textwidth}
           \centering
            \includegraphics[width=\columnwidth]{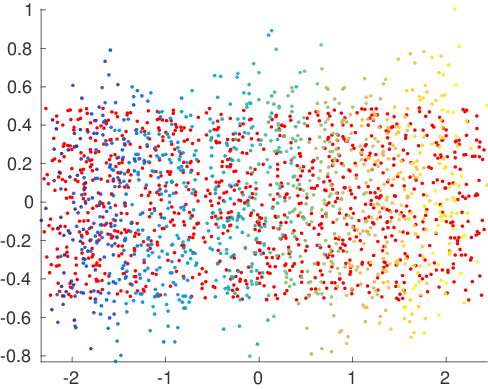} 
            \caption{Density ridge unwrapping.}
            \label{fig:density_rige_semiroll}
    \end{subfigure}
    \caption{Uncovering the underlying parameterization of figure~\ref{fig:comp_semiroll} using various manifold unwrapping algorithms.}
    \label{fig:comp_semiroll_all}
\end{figure}

Finally we do the same comparison on a synthetic data set with a high level of noise to indicate that our algorithm can handle noisy situations. The data set with and without noise and the density ridge unwrapped version projected to the tangent space of the reference mode is shown in figure~\ref{fig:comp_semiroll}. We see that even with a quite high level of noise the density ridge unwrapping gives meaningful results. In figure~\ref{fig:comp_semiroll_all} we see the data processed to two-dimensions by isomap, lle, mvu, ltsa, laplacian eigenmaps and density ridge unwrapping. The original parameterization is shown in red dots, and the unwrapped coordinates from the different algorithms are shown with the same color coding as in figure~\ref{fig:comp_semiroll}. All result were obtained with neighborhood parameter $k=12$ and the results as well as the parameterization have been normalized and centered. 

Looking at the figures we see that isomap is the closest to ours, but it is not able to capture the manifold without retaining the noise. The other four algorithms, laplacian eigenmaps, local linear embedding, maximum variance unfolding and local tangent space aligment all fail to unfold the underlying manifold and the results are accordingly hard to comment on. We tried different parameters for all algorithms, but all failed to give reasonable results, except isomap and our algorithms, that is why we chose to show only the results of the default parameter $k=12$.
\section{Conclusion and future work}
\label{sec:conlusion}
In this paper we have shown that density ridges of a kernel density estimate can be used to unwrap manifolds. 
We have presented
a geometrically intuitive algorithm for unwrapping one-dimensional
manifolds embedded in euclidean space based on
gradient flow.
We have extended this algorithm to higher dimensional manifolds, replacing coordinates calculated via integral curves to local linear projections.
promising results and illustrations have been shown on both synthetic and real data sets that match the geometric
intuition of unfolding. 

We conclude this paper with a short section on possible extensions and directions of further research.
\subsection{Future research}
The work presented in this paper contains suggestions on how to learn linear coordinate systems on non-linear data distributions that can be represented by principal curves or surfaces (density ridges). The perhaps most obvious lines of further research are related to the properties of the underlying manifold and the estimate of the manifold surrogate which relies on the kernel density estimate of the distribution of points on the manifold. 

Regarding the properties of the underlyng manifold theoretical results should be established related to upper bounds on the curvature, boundary effects and smoothness properties in higher dimension. The online appendix for the convergence proofs of the isomap algorithm could be applied here~\citep{tenenbaum2000global}.

Furthermore, the properties of the underlying manifold directly influences the density ridge estimate. Especially the choice of kernel size and how it influences the relationship between curvature, bias and smoothness of the density ridge estimate should be established theoretically. Also the properties of the kernel density estimate is known to deteriorate when the dimension increases since the distance distribution of a point set converges to the mean distance in higher dimensions. An upper bound on dimensions where the gradient flow of a kernel density estimate holds could be established and used in our setting. 

Recently chacon and duong,~\citet{chacon2013data} suggested kernel size estimators that yield different sizes for kernel density its gradient and its hessian. These could be implemented in our setting, perhaps increasing accuracy in density ridge estimations.




\section*{Aknowledgements}
This work was partially funded by the research council of norway over fripro grant no.\ 239844 on developing \emph{the next generation learning machines} (rj).

\appendix

\section{Approximate parallel transport}
\label{app:parallelTransport}
In this section we present the algorithm for approximate parallel transport in the $d$-dimensional setting. 
Let $\hat{R}_i$ be the points on the ridge estimate that lies in the basin of attraction of mode $\mathbf m_i$ and $C_i$ the local coordinates of $\hat{R}_i$,  $C_i = Q_{||}(\mathbf m_i)Q_{||}(\mathbf m_i)^T \hat{R}_i$. 
Let $\boldsymbol\gamma$ be the vector containing the points of the approximate geodesic from $\gamma_0 = \mathbf m_i$ to $\gamma_n = \mathbf m_j$ as found by \eqref{eq:geod_opt}.
The approximate parallel transport is performed by translating the local coordinates $C_i$ along the finite difference tangent vectors of the approximate geodesic. At each step of the translation the points are projected/rotated to the local tangent space by $Q_{\parallel}(\gamma_t)Q_{\parallel}(\gamma_t)^T$. This is to ensure that the local coordinates stay in the tangent space all the way along the geodesic towards the target mode $\mathbf m_j$.
The algorithm for approximate parallel transport is summarized in Algorithm~\ref{alg:dDimParallelTransport}.
\begin{algorithm}[htbp]
  \caption{Approximate parallel transport from $\mathbf m_i$ to $\mathbf m_j$}
  \begin{algorithmic}[1]
    \REQUIRE $C_i$ and $\boldsymbol\gamma=\left [ \gamma_t \right]_{t=1}^n$.
    \STATE Calculate $Q_{||}(\gamma_t)$ for $t=[1, \cdots, n]$.
    \STATE Initialize $C_i^{'} = C_i$.
  \FOR{$t = [2, \cdots, n]$}
\STATE \begin{equation}
 C_i^{''} = Q_{||}(\gamma_t) Q_{||}(\gamma_t)^T\left(C_i^{'} + (\gamma_t - \gamma_{t-1})\right ) 
\end{equation}
\STATE $C_i^{'} = C_i^{''}$
  \ENDFOR
    \ENSURE Local coordinates $C_i^{'}$ transported from $\mathbf m_i$ to $\mathbf m_j$ along the manifold.
  \end{algorithmic}
  \label{alg:dDimParallelTransport}
\end{algorithm}

\section{Algorithm for unwrapping transported coordinates}
\label{app:unfolding}
Given a framework for approximate parallel transport, we can unwrap isometric manifolds. This is done by first transporting all local linear attraction basin approximations (charts) to a reference mode. Then each chart is iteratively translated back along the geodesic it came from whilst at the same time projecting each step down to the dimension of the tangent space along the same geodesic. This will give an unfolded version of the manifold where all charts lie in the dimension of the tangent space of the manifold centered on the reference mode.


Let $\boldsymbol\nu$ be the vector $\boldsymbol\gamma$ in reversed order. Since $\boldsymbol\nu$ lies approximately on the manifold the finite difference tangent vectors $\delta\nu = \nu_{t+1} - \nu_t$ will also lie on the manifold. 
So by pushing the coordinates along the geodesic and at the same time projecting to the intrinsic dimension of the manifold, we get a direct unfolding. The complete algorithm is presented in Algorithm~\ref{alg:isometricUnfolding}.

Consider standing on the north pole of the earth and taking a step in some direction along the surface of the earth. Following the algorithm, one would take a step in the chosen direction and project the vector pointing from the north pole to the destination of the step to the tangent space of the north pole. This would yield a two-dimensional step in the tangent space of the north pole. Then the procedure is repeated from the next position; take a step and then project to the tangent space of the origin. The tangent space will still be two-dimensional, but the orientation might have changed slightly, so that we need to compensate for the change of basis from tangent space to tangent space.

In this manner one will get a two-dimensional representation of the path walked, e. g.\ walking from the north pole to the south pole along a great circle would yield a straight line in the unfolded coordinates. Note the similarity to stereographic projection coordinates. Of course this example is in effect not relevant since the earth is close to a sphere which cannot be unfolded isometrically, but the analogies to the algorithm still holds. 
\begin{algorithm}[htbp]
  \caption{Isometric unfolding}
  \begin{algorithmic}[1]
    \REQUIRE Local coordinates transported to a reference mode $C = \{C_i^{'}\}_{i=1}^{m-1}$ and the reversed geodesic from reference mode to all other modes, $\boldsymbol\nu_i$.
    \STATE Choose a basis $E_i$ for the tangent space of a point on the reversed geodesic $\nu_t$, $T_{\nu_t}M$:
    \begin{equation*}
    E_i = 
    \begin{bmatrix}
    \Delta\nu_t & \text{null}\left(\begin{bmatrix}\Delta\nu_t & Q_{\perp}\left(\nu_t\right)     \end{bmatrix}\right)\end{bmatrix}^T.
    \end{equation*}
    \FORALL{Translated charts $C_i^{'}$}
    \FOR{$t = [2, \cdots, n]$, $n$ is the number of steps in $\boldsymbol\nu_i$}
    \STATE Choose the translation direction as being along $\nu_t$ such that the step along the geodesic will be $\xi = [\|\nu'_t\|, \; \mathbf 0]^T$, $\xi\in\mathbb R^D$, $\mathbf 0 \in \mathbb R^{D-1}$. 
    \STATE Make sure that the basis stays consistent: $B = BE_t^TE_{t+1}$, initialized as $B = E_1$.
    \STATE Project the translation to the tangent space, $U_i = Q_{\parallel}(\nu_t)Q_{\parallel}(\nu_t)^T\left(C^{'}_i + B\xi_t\right)$
    \ENDFOR
    \ENDFOR
    \ENSURE Unwrapped manifold consisting of all unwrapped charts $\widehat M = \bigcup_{i=1}^{m}U_i$. 
  \end{algorithmic}
  \label{alg:isometricUnfolding}
\end{algorithm}

\bibliographystyle{IEEEtran}
\bibliography{refs.bib}

\end{document}